\documentclass[sigconf]{acmart} %
\usepackage{amsmath,amssymb,bbm,bm}
\usepackage[utf8]{inputenc} 
\usepackage[T1]{fontenc}    
\usepackage{hyperref}       
\usepackage{url}            
\usepackage{booktabs}       
\usepackage{amsfonts}       
\usepackage{nicefrac}       
\usepackage{microtype}      
\usepackage{graphicx}
\usepackage[font=footnotesize]{subcaption}
\usepackage{lipsum}
\usepackage{multirow}
\usepackage{comment}
\usepackage{color}
\usepackage{etoolbox}
\usepackage{booktabs}
\PassOptionsToPackage{table,xcdraw}{xcolor}

\newtoggle{comments}
\settoggle{comments}{true} 
\iftoggle{comments}{
  \newcommand {\alberto}[1]{{\color{orange}{~Alberto: #1}\normalfont}}
  \newcommand {\bjin}[1]{{\color{blue}{~Baihong: #1}\normalfont}}
  \newcommand {\yuxin}[1]{{\color{violet}{~Yuxin: #1}\normalfont}}
  \newcommand {\poolla}[1]{{\color{green}{~Kameshwar: #1}\normalfont}}
  \newcommand {\red}[1]{{\color{red}{#1}\normalfont}}
}{
  \newcommand {\alberto}[1]{{}}
  \newcommand {\bjin}[1]{{}}
  \newcommand {\yuxin}[1]{{}}
  \newcommand {\poolla}[1]{{}}
  \newcommand {\red}[1]{{}}
}

\newcommand{\expctover}[2]{\mathbb{E}_{#1}\!\left[#2\right]}

\usepackage{acronym}
\acrodef{OOD}{out-of-distribution}
\acrodef{FDD}{Fault Detection and Diagnosis}
\acrodef{LDA}{Linear Discriminant Analysis}
\acrodef{ML}{Machine Learning}
\acrodef{CFAR}{Constant False Alarm Rate}
\acrodef{RF}{Random Forest}
\acrodef{LR}{Logistic Regression}
\acrodef{NN}{Neural Network}
\acrodef{DT}{Decision Tree}
\acrodef{OC-SVM}{One-Class Support Vector Machine}
\acrodef{AE}{Autoencoder}

\acrodef{FPR}{False Positive Rate}
\acrodef{FNR}{False Negative Rate}
\acrodef{SL}{Severity Level}
\acrodef{ROC}{Receiver Operating Characteristic}
\acrodef{CI}{confidence interval}
\acrodef{IS}{Intermediate-Severity}
\acrodef{RBF}{Radial Basis Function}
\acrodef{PCA}{Principle Component Analysis}
\acrodef{LSTM}{Long Short-Term Memory}

\AtBeginDocument{%
  \providecommand\BibTeX{{%
    \normalfont B\kern-0.5em{\scshape i\kern-0.25em b}\kern-0.8em\TeX}}}

\setcopyright{acmcopyright}
\copyrightyear{2018}
\acmYear{2018}
\acmDOI{10.1145/1122445.1122456}

\acmConference[]{}{}{}
\acmBooktitle{}
\acmPrice{}
\acmISBN{}

\begin{document}

\title{Are Ensemble Classifiers Powerful Enough for the Detection and Diagnosis of Intermediate-Severity Faults?}

\author{Baihong Jin}
\affiliation{%
  \institution{University of California, Berkeley}
}
\email{bjin@eecs.berkeley.edu}

\author{Yingshui Tan}
\affiliation{%
  \institution{University of California, Berkeley}
}
\email{tys@eecs.berkeley.edu}

\author{Yuxin Chen}
\affiliation{%
  \institution{University of Chicago}
}
\email{chenyuxin@uchicago.edu}

\author{Kameshwar Poolla}
\affiliation{University of California, Berkeley}
\email{poolla@eecs.berkeley.edu}

\author{Alberto Sangiovanni~Vincentelli}
\affiliation{\institution{University of California, Berkeley}}
\email{alberto@eecs.berkeley.edu}

\renewcommand{\shortauthors}{B.~Jin, et al.}

\begin{abstract}
\ac{IS} faults present milder symptoms compared to severe faults, and are more difficult to detect and diagnose due to their close resemblance to normal operating conditions. The lack of \ac{IS} fault examples in the training data can pose severe risks to \ac{FDD} methods that are built upon \ac{ML} techniques, because these faults can be easily mistaken as normal operating conditions. Ensemble models are widely applied in \ac{ML} and are considered promising methods for detecting \ac{OOD} data. We identify common pitfalls in these models through extensive experiments with several popular ensemble models on two real-world datasets. Then, we discuss how to design more effective ensemble models for detecting and diagnosing \ac{IS} faults.
\end{abstract}




\maketitle

\section{Introduction}\label{sec:intro}



\begin{figure}[b]
    \centering
    \vspace{-7mm}
    \includegraphics[height=3.5cm]{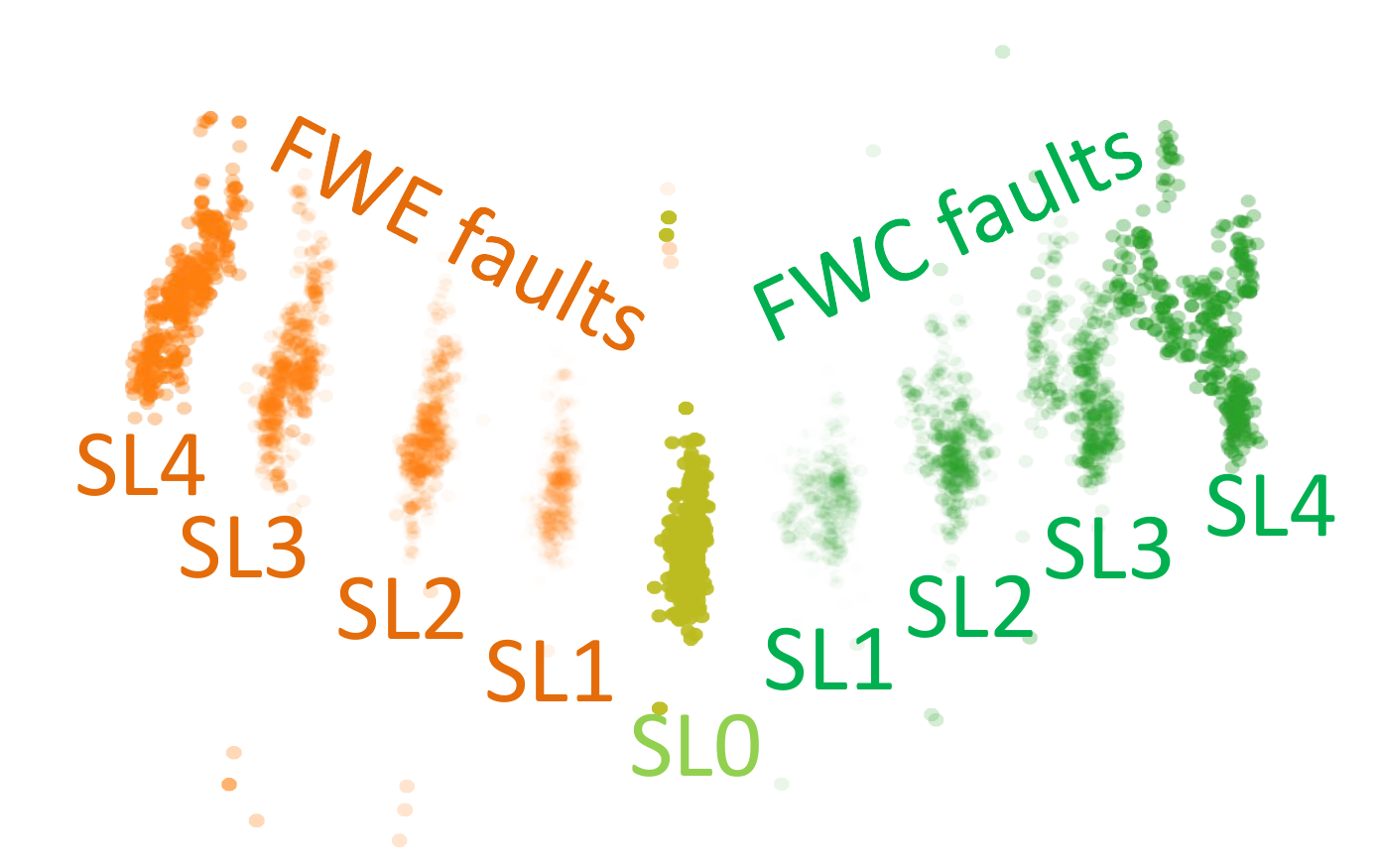}
    \vspace{-3mm}
    \caption{Visualization of part of the RP-1043 data~\cite{comstock1999development}. The normal condition and two fault conditions (each with four \acp{SL}) are shown.}
    \label{fig:visualization-chiller}
\end{figure}

In \acf{FDD} applications\footnote{In this paper, a ``fault'' can mean either a machine fault in industrial applications or a human disease in health applications.}, it is common to encounter fault data examples whose symptoms correspond to different \acp{SL}. Figure~\ref{fig:visualization-chiller} shows a real-world example where faults are categorized into four different \acp{SL}, from SL1 (slightest) to SL4 (most severe). The ability of accurately assessing the severity of faults/diseases is important for \ac{FDD} applications, yet also very difficult especially on low-severity examples; SL1 data clusters are much closer to the normal cluster than to their corresponding SL4 clusters in Figure~\ref{fig:visualization-chiller}. An \ac{FDD} system needs to be very sensitive to identify the low-severity faults; at the same time, it should not have too many false positives, which makes the design of such decision systems a challenging task.

\begin{figure}[tb]
    \centering
    \includegraphics[height=4cm]{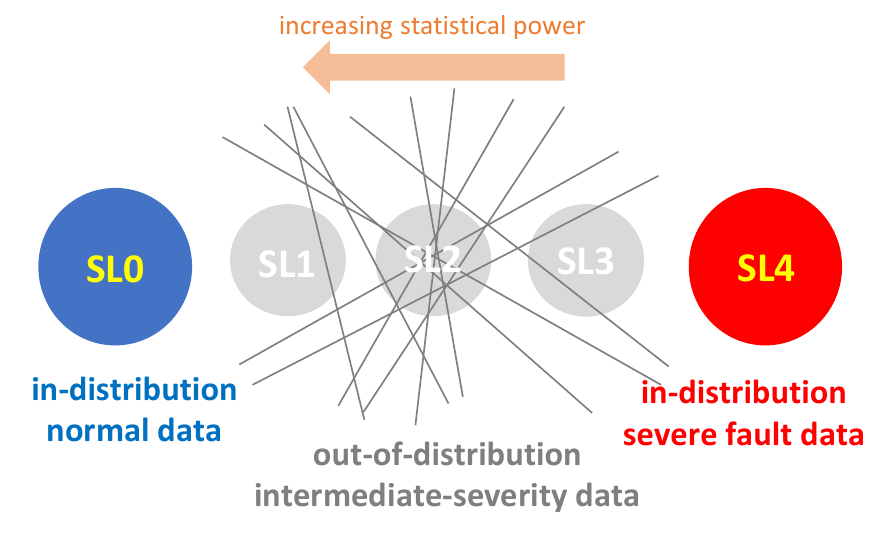}
    \caption{Illustration showing how an ensemble classifier can conceptually help detect \ac{IS} faults. The gray lines represent the decision boundaries of base learners in the ensemble.}
    \label{fig:ensemble-illustration}
\end{figure}

If labeled data from different \acp{SL} are available, then regular regression or classification approaches are suitable, as already exemplified by previous research~\cite{krause2018grader,li2016fault}.
However, these fine-grained labeled datasets can take much effort to prepare and we may not always have \textit{a priori} access to a full spectrum of fault \acp{SL}. In an extreme case, as illustrated in Figure~\ref{fig:ensemble-illustration}, suppose we only have access to the two ends (i.e. SL0 \& SL4) of the fault \ac{SL} spectrum, and the \acf{IS} fault instances are not available to us. If we train a classification system only using the available SL0 and SL4 data, the resulting classifier may have great performance on in-distribution data (SL0 \& SL4). However, it may fail badly on identifying the \ac{IS} data. For example, most SL1 faults may be mistakenly recognized as normal by any of the decision boundaries shown in Figure~\ref{fig:ensemble-illustration}. More generally, classical supervised learning approaches designed for achieving maximal separation between labeled classes (e.g. margin-based classifiers, (discriminative) neural networks, etc), are no longer effective in detecting IS fault instances.

In the absence of labeled data for certain categories of fault instances, common practices are to develop generative models, such as Gaussian mixture model~\cite{ZongSMCLCC18}, \ac{PCA}~\cite{huang2007network,zhang2017improved}, \ac{LSTM} network~\cite{du2017deeplog} and autoencoder \cite{sakurada2014anomaly,jin2019encoder}. A potential problem for these models is that they may not always generalize well---that is, a single-trained model, when applied to an unseen IS fault instance during test time, can be classified as normal (i.e.,~false positive)~\cite{du2019lifelong}. 
%
The solution we propose in this paper is based on ensemble learning~\cite{zhou2012ensemble}, i.e., on the process of training multiple classifiers and leveraging their joint decisions to recognize \ac{IS} faults. In literature, a variety of ensemble methods have been proposed on the estimation of decision uncertainties~\cite{leibig2017leveraging,gal2016uncertainty,lakshminarayanan2017simple}. Figure~\ref{fig:ensemble-illustration} shows that the individual classifiers have much disagreement on the SL2 data. The amount of disagreement can be used to measure the decision uncertainties, and is therefore useful for indicating \ac{IS} faults. However, for SL1 data that are much closer to the normal cluster, the above approach will be less effective. We find this is a common phenomenon in our empirical studies. A remedy to the problem is to increase the \textit{statistical power} of the base learners by moving the decision boundaries towards the normal cluster. In this paper, we focus on the special considerations required for designing effective ensemble models for \ac{FDD} applications.


We believe our methods are a useful complement to the literature on multi-grade \ac{FDD} systems~\cite{krause2018grader, li2016fault, li2016data}, specifically under cases where the available fault data for training are insufficient and cannot cover the entire severity spectrum. 
In this paper, we give some caveats and an easy-to-use recipe for \ac{ML} practitioners to develop ensemble \ac{FDD} models that can more effectively recognize \ac{IS} examples. We summarize our contributions in this paper as follows:
\begin{itemize}
    \item We show by experiments that \ac{IS} examples, when missing from the training distribution, can pose severe risks for several popular supervised and unsupervised \ac{ML}-based \ac{FDD} models.
    \item Further experiments show that the appropriate setting of the statistical power of ensemble classifiers has a vital impact on the \ac{FDD} performance. We propose using an adaptive approach, \textsc{cfar} (i.e. Constant False Alarm Rate) for setting decision thresholds that improve the statistical power, while at the same time keeping the \ac{FPR} under control. 
    \item In order to aggregate the probabilistic outputs from ensemble classifiers, we propose a novel statistic \textsc{mean-std}. \textsc{mean-std} leads to similar detection performance as the simple average method (\textsc{mean}) for aggregating probabilities; however, experiments show that \textsc{mean-std} yields significant improvement on the diagnosis performance of ensemble models.
\end{itemize}
The rest of this paper is organized as follows. We formulate the fault detection and diagnosis problems in Section~\ref{sec:problem-formulation}. Next, in Section~\ref{sec:methodology}, we  describe in details our methodology. The two datasets used in our empirical study are briefly described in Section~\ref{sec:dataset-descriptions}, and in Section~\ref{sec:experiment}  experimental results are presented. In Section~\ref{sec:related-works}, we review related research topics found in the literature. We summarize the findings in this paper and discuss future work in Section~\ref{sec:conclusion}.

\section{Problem Formulation}\label{sec:problem-formulation}

Suppose an \ac{FDD} system aims at differentiating the normal condition from $n-1$ fault conditions by monitoring the system state $\bm{x}$. 
A classifier that takes $\bm{x}$ as input can be used for that purpose. Let ${z}\in\{0,1,2,\ldots,n-1\}$ be the ground-truth label of state $\bm{x} \in\mathbb{R}^d$. An ``ordinary'' classifier is some rule, or function, that assigns a class label $\hat{z}\in\{0,1,2,\ldots,n-1\}$ to an input example $x$. 
Without loss of generality, we will use index $0$ for the normal condition, and the fault conditions will be denoted by $1,2,\ldots,n-1$, respectively. 

\paragraph{Fault Detection}

Let $\mathcal{X}$ be the set of data points, and $\mathcal{M}$ be a model class. Each $M\in\mathcal{M}$ defines an \textit{anomaly score} function $s^M:\mathcal{X}\rightarrow\mathbb{R}$ that characterizes how likely a data point corresponds to a fault state; a larger $s^M(x)$ implies a higher chance of a data point $\bm{x}$ being a fault. For a given threshold value $\tilde{s}$, we can define the \acf{FNR} and \acf{FPR} of the model $M$ on the test data distribution as follows:
\begin{align}
  &\text{FNR}(s^M,\tilde{s}) = \expctover{}{\mathbbm{1}\{s^M(x) \leq \tilde{s}\} \mid x \text{~is a fault}}.\notag\\
  &\text{FPR}(s^M,\tilde{s}) = \expctover{}{\mathbbm{1}\{s^M(x) > \tilde{s}\} \mid x \text{~is normal}}.\notag
\end{align}

Let $\mathcal{X}^\text{train}$ be a subset of labeled data points for training, each labeled as one of the $n$ healthy states. Ideally, the goal is to learn an anomaly score function $s^\ast$ by minimizing its classification error on $\mathcal{X}^\text{train}$, and then decide a corresponding threshold $\tilde{s}$, such that $(s^\ast,\tilde{s})$ can optimize the \ac{FNR} and the \ac{FPR} on unseen test data. However, in practice, one has to make a trade-off between the two metrics. In this paper, we aim to learn $(s^\ast,\tilde{s})$ that minimizes the \ac{FNR} (i.e., missed detections of faults), while ensuring that the \ac{FPR} is below a certain level.


\paragraph{Fault Diagnosis}
A fault diagnosis model not only needs to detect the existence of faults, but should also be able to differentiate between faults of different classes, and come up with a diagnosis decision. In this paper, we view a fault diagnosis task as \textit{ranking} the $n$ health states by the likelihood of the system to be in that state.

The actual form of diagnosis decisions is based on the context. For evaluating the diagnosis performance, we can generally use the top-$k$ metric that is widely applied in multiclass classification tasks. The top-$k$ metric measures how accurately a model can rank the true class label into the top $k$ labels. Given $k$, our goal is to train a classifier that can maximize the top-$k$ accuracy on unseen test data which may include \ac{OOD} \ac{IS} faults.


\section{Methodology}\label{sec:methodology}

In this paper, we restrict our attention to classifiers that produce probabilistic outputs $p\in\mathbb{R}^n$. Each output $p_i$ can be interpreted as the likelihood of input $\bm{x}$ belonging to class $i$. Typical classification schemes falling under this category include \acf{NN} and \acf{DT} models. In the sequel, we  describe our methodology of using ensemble learning for \ac{FDD}. In particular, we adopt a statistical hypothesis testing view when designing the \ac{FDD} algorithms. The null hypothesis will be ``$\bm{x}$ is in normal condition'', and we will decide whether to reject the null hypothesis based on the outputs from the trained classification models.

\subsection{Fault Detection Tasks}\label{sec:detection}

Suppose we have $m$ data points for testing, and they are organized into a design matrix $\bm{X}\in\mathbb{R}^{m\times d}$. The output probabilities can be accordingly written as a $m\times n$ matrix $\hat{\bm{Y}}$. Now consider an ensemble case with $r$ members (learners) in the ensemble. The output probabilities from the ensemble model form an $r\times m\times n$ tensor. Using the superscripts to differentiate ensemble members, we can denote the prediction results from learner $k$ ($k=0,1,2,\ldots,r-1$) by
\begin{align}
\hat{\bm{Y}}^{(k)} =\begin{pmatrix}
\hat{y}_{0,0}^{(k)} & \hat{y}_{0,1}^{(k)} & \ldots & \hat{y}_{0,n-1}^{(k)}\\
\hat{y}_{1,0}^{(k)} & \hat{y}_{1,1}^{(k)} & \ldots & \hat{y}_{1,n-1}^{(k)}\\ 
\vdots & \vdots & \ddots & \vdots\\
\hat{y}_{m-1,0}^{(k)} & \hat{y}_{m-1,1}^{(k)} & \ldots & \hat{y}_{m-1,n-1}^{(k)}
\end{pmatrix}
\end{align}\label{eqn:prob-matrix}
The two subscripts $i,j$ indicate the index of data points ($0,1,\ldots,m-1$), and the index of classes ($0,1,\ldots,n-1$). 

To come up with a decision threshold for each input $\bm{x}_i$, we need to first find a way to aggregate the $r$ probability values from each learner into test statistics. A simple and widely used way, a.k.a.~soft voting~\cite{zhou2012ensemble}, is to take their arithmetic average (hereafter referred to as \textsc{mean}).
\begin{align}
    \mu_{i,j} \triangleq \frac{1}{r}\sum_{k=0}^{r-1} y_{i,j}^{(k)}.
\end{align}
\textsc{mean} 
captures the \textit{consensus} among ensemble members, but does not reflect the disagreement well.
To better capture the disagreement among ensemble members, the standard deviation of the ensemble probabilities $\sigma_{i,j}$ is a better choice.
  
Intuitively, the higher the standard deviation, the greater the amount of ``disagreement'' among ensemble members. We propose to combine the first-order and second-order moment information using the formula below (hereafter referred to as \textsc{mean-std}),
\begin{align}\label{eqn:mean-var-score}
    \theta_{i,j} \triangleq 
\begin{cases}
    \mu_{i,j} - \lambda\sigma_{i,j},~~j=0,\\ 
    \mu_{i,j} + \lambda\sigma_{i,j},~~j=1,2,\ldots,n-1.
\end{cases}
\end{align}
We would like $\theta_{i,j}$ to be a statistic that indicates how likely $\bm{x}_i$ belongs to class $j$. Recall that $\hat{y}_{i,j}$ is an estimation of $\Pr(z_i=j\,\vert\,\bm{x}_i)$. For $j=0$ (the normal class), a large standard deviation $\sigma_{i,j}$ reflects a larger risk in believing $\bm{x}_i$ is in normal condition, which is the reason why we introduce a discount of $\lambda\sigma_{i,j}$ in the above definition. For similar reasons, we add an extra term to $\theta_{i,j}$ for cases where $j \neq 0$.    

By comparing the two definitions \textsc{mean} and \textsc{mean-std}, we can immediately see that \textsc{mean} is a special case ($\lambda=0$) of \textsc{mean-std}. The setting of $\lambda$ is itself a topic that is worth further investigation; in this study for \textsc{mean-std} we will fix $\lambda=1$ for simplicity.


\subsubsection{\textsc{argmax} detection}\label{sec:argmax-detection}

The \textsc{argmax} rule picks the class with the largest entry (probability etc.) as the prediction, which is 
a widely applied decision method for \ac{ML}-based classification schemes. For detection, we can simply check whether the predicted class is normal (index 0) or not. 
Considering \textsc{mean} as a special case of \textsc{mean-std} and use $\theta$ as the symbol for test statistics in general, the decision rule for detection can be written as,
\begin{align}\label{eqn:detection-rule-argmax}
    \bm{x}_i\text{ is}
    \begin{cases}
        \text{normal,} &\text{if }\arg\max_j \theta_j = 0,\\
        \text{fault,} &\text{otherwise}. 
    \end{cases}
\end{align}

In a binary classification scenario, the \textsc{argmax} detection rule is equivalent to using $0.5$ as the decision threshold.


\subsubsection{\textsc{cfar} detection}\label{sec:cfar-detection}


Target detection is an important task in radar systems. In real-world applications, the background noise is often unknown and difficult to characterize, making it challenging to decide the detection threshold that maximizes the chance of correct detection while keeping the false alarm rate low. The \acf{CFAR} scheme~\cite{richards2005fundamentals} is an adaptive approach for varying the detection thresholds as a function of the sensed environment, such that the false alarm rate is always controlled to be below a certain level.

The challenge that we encounter in \ac{ML}-based \ac{FDD} applications is similar: a trained model $M$ will assign (non-zero) anomaly scores to both normal and fault examples, and we need to decide an optimal detection threshold to divide them. Due to the costs associated with false alarms as explained earlier in Section~\ref{sec:problem-formulation}, we want to control the \ac{FPR} to be below a preset level of $\alpha$. 
Recall that $\theta_{i,0}$ can be used as a statistic for evaluating how likely $\bm{x}_i$ belongs to the normal class.
To achieve this goal in practice, we choose a detection threshold $\zeta$ such that the \ac{FPR} on the training distribution is $\alpha$. 
\begin{align}
     \frac{1}{\left\vert{\mathcal{X}^\text{train}_\text{normal}}\right\vert}\sum_{i\in\mathcal{X}^\text{train}_\text{normal}} \mathbb{I}(\theta_{i,0} < \xi) = \alpha,\label{eq:cfar}
\end{align}
where $\mathcal{X}^\text{train}_\text{normal}$ represent the set of normal data points in the training set.
In other words, $\zeta$ is the $\alpha$-percentile of all $\theta_{i,0}$ values on the training set normal data. The detection rule can now be written as,
\begin{align}\label{eqn:detection-rule-cfar}
    \bm{x}_i\text{ is}
    \begin{cases}
        \text{normal,} &\text{if }\theta_{i,0} > \xi\\
        \text{fault,} &\text{otherwise.} 
    \end{cases}
\end{align}
Due to the similarity of Equation~\eqref{eq:cfar} to the \ac{CFAR} scheme in radar systems, we will hereafter refer to our decision criterion~\eqref{eqn:detection-rule-cfar} as \textsc{cfar}.

In addition to supervised classification tasks, the \textsc{cfar} scheme can also be applied to autoencoder models for deciding a threshold on the reconstruction error for detecting anomalies. 





\subsection{Fault Diagnosis Tasks}\label{sec:diagnosis}

When there is more than one type of fault, a subsequent step to detection tasks are \textit{diagnosis} tasks; we want to find out potential root causes of anomalous behaviors. Unsupervised learning approaches are typically not suitable for diagnosis tasks, because they by nature lack the ability to discriminate between different types of faults.

To use a supervised classifier for fault diagnosis, an easy and commonly used approach is to use the \textsc{argmax} decision rule---pick the class with the highest (top-1) score. The approach is sound and reasonable; however, it can be further improved by taking into account the detection decision and the class with the \textit{second} highest score. Suppose that an alarm has been raised by some detection system for a given input $\bm{x}_i$. Even if the top-1 probability still points to the normal class, we already know from detection results that there is probably some anomaly happening, so the highest-score class (the normal class in this case) will not matter anyway. If now the class with the second highest probability is the actual fault class, we will still obtain the correct diagnosis. Therefore, the top-2 accuracy is a reasonable metric for evaluating the fault diagnosis performance of multi-class classifiers, in addition to the top-1 accuracy.

\subsection{Creating Diversity among Ensembles}

Diversity is recognized as one of the key factors that contribute to the success of ensemble approaches~\cite{brown2005diversity}. As illustrated in Fig.~\ref{fig:ensemble-illustration}, the diversity among ensemble members is crucial for improved detection performance on \ac{OOD} data instances. In our empirical study to be described later, we will employ bagging and dropout techniques for creating diversity among ensemble members. The two techniques will be briefly reviewed next. 

\subsubsection{Bagging}

Bagging~\cite{breiman1996bagging} (or bootstrap aggregation) is a classical approach for creating diversity among ensemble members. The core idea is to construct models from different training datasets using \textit{randomization}. In the original bagging approach~\cite{breiman1996bagging}, a random subset of the training samples is selected for training each member classifier. A later variant, the so-called ``feature bagging'' selects a random subset of the features for training. One famous application of bagging in \ac{ML} is the \ac{RF} model. In our empirical study, we will use both ``sample bagging'' and ``feature bagging'' to induce diversity among ensemble classifiers.

In this study, only \textit{homogeneous} base learners, i.e. models of the same type, are used to construct ensembles. The case of heterogeneous ensembles is an interesting setting and we leave it for future investigation.

\subsubsection{Monte Carlo Dropout (MC-Dropout)}

Dropout~\cite{srivastava2014dropout} is a popular and powerful regularization technique to prevent overfitting neural network parameters. The key idea is to randomly drop units along with their connections from the network during training. Each individual hidden node is dropped at a probability of $p$ (i.e.~the dropout rate). Recently, Gal and Ghahramani proposed using MC-dropout~\cite{gal2016uncertainty} to estimate a network's prediction uncertainty by using dropout also at test time. By sampling a dropout model $\mathcal{M}$ using the same input for $T$ times, we can obtain an ensemble of prediction results with $T$ individual probability vectors. In effect, the dropout technique provides an inexpensive approximation to training and evaluating an ensemble of exponentially many similar yet different neural networks. The MC-dropout technique was also applied to the estimation of decision uncertainty in diabetic retinopathy diagnosis~\cite{leibig2017leveraging} and also the detection of low-severity faults in chiller systems~\cite{jin2019detecting}.

\section{Dataset Descriptions}\label{sec:dataset-descriptions}

\subsection{RP-1043 Chiller Dataset}\label{sec:rp-1043}

We used the ASHRAE~RP-1043 Dataset~\cite{comstock1999development} (hereafter referred to as the ``chiller dataset'') to test out the proposed ensemble approach for detecting and diagnosing \ac{IS} faults. In RP-1043 dataset, sensor measurements of a typical cooling system---a 90-ton centrifugal water-cooled chiller---were recorded under both fault-free and various fault conditions. In this study, we include the six faults (FWE, FWC, RO, RL, CF, NC) as used in Jin~et~al's previous study~\cite{jin2019detecting}; each fault was introduced at four levels of severity (SL1--SL4, from slightest to severest). For feature selection, we also follow previous study~\cite{jin2019detecting} and use the sixteen key features therein for training our models.

To give the readers an intuitive view about the distribution of RP-1043 data, we employ the \ac{LDA} algorithm to reduce the data into two dimensions, and visualized part of the reduced-dimension data as shown in Figure~\ref{fig:visualization-chiller} that was described earlier. We can also see a general trend that, data points will deviate further away from the normal cluster when the corresponding fault develops into a higher \ac{SL}.

\begin{figure}[tb]
    \centering
    \includegraphics[width=0.9\linewidth]{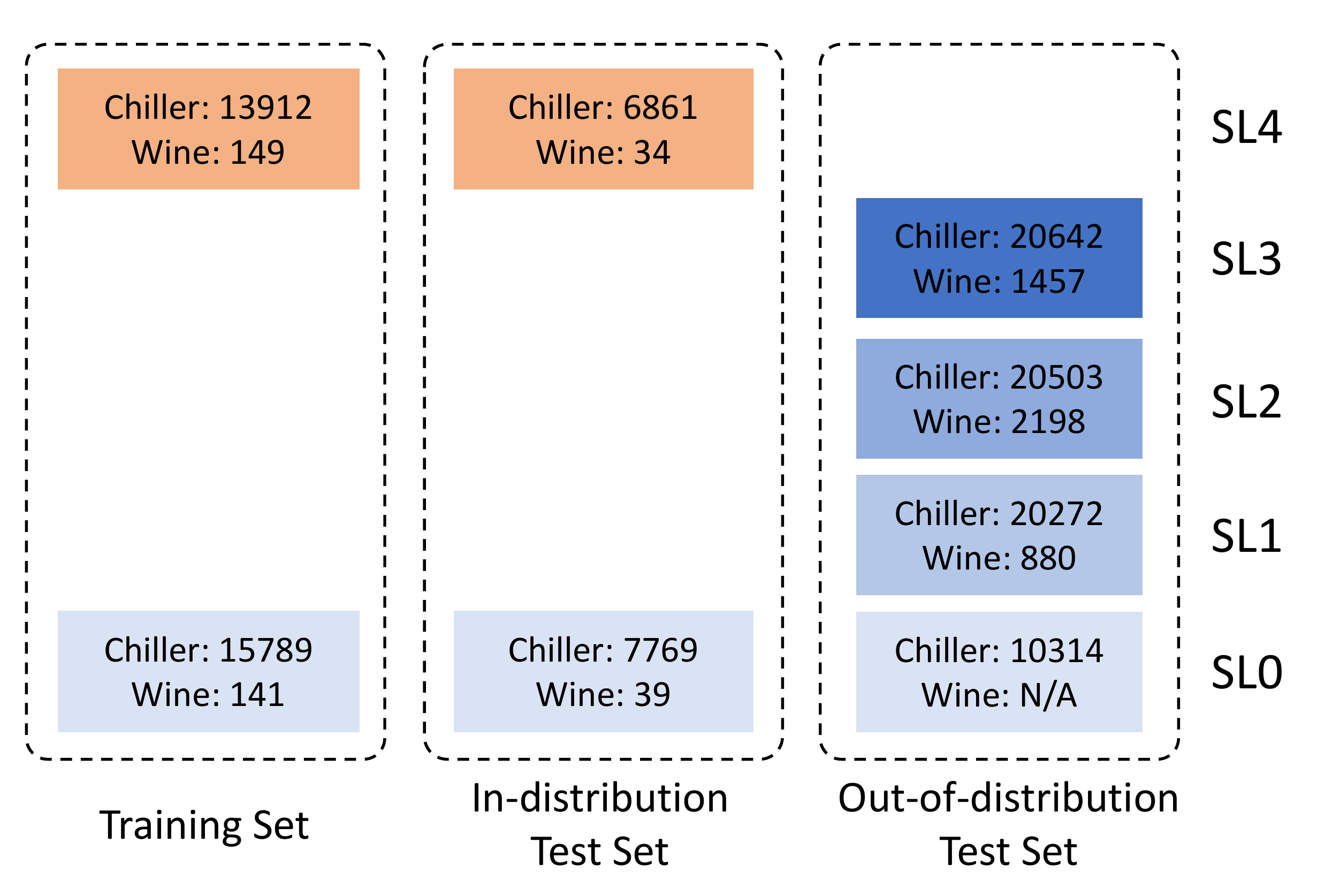}
    \caption{Layout of the training sets and test sets used in this study.}
    \label{fig:dataset-layout}
\end{figure}

Figure~\ref{fig:dataset-layout} illustrates the layout of the training and test sets generated from the chiller dataset that will later by used in our case study. For training our classification models, only part of the SL0 and the SL4 data are used; the rest are reserved for testing purposes. We have two test sets in this study, the first test set is comprised of only SL0 and SL4 data (referred to as the \textit{in-distribution test set}). Another test set (i.e. the \ac{OOD} test set) includes SL1--SL3 data (\ac{IS} faults). Also included in the \ac{OOD} test set is a different set of SL0 data (normal data collected under different circumstances than in the training distribution). The number of data points in each subset can also be found in Figure~\ref{fig:dataset-layout} in the appendix. More details about our preprocessing steps of the chiller dataset will be given in Section~\ref{sec:chiller-dataset-appendix} in the appendix.

\subsection{Wine Quality Dataset}\label{sec:wine}

The wine quality dataset (hereafter referred to as the ``wine dataset'') contains the physicochemical properties and the associated ratings that characterize samples of the Portuguese ``Vinho Verde'' wine~\cite{cortez2009modeling}. We pick the white wine part of the dataset for our case study. Each data point is comprised of 11 sensor measurements and the associated numeric rating (ranging from 1 to 10) of a wine sample. For the ease of exposition, we will use the same notions of \acp{SL} (SL0 - SL4) as in the chiller dataset, instead of the original ratings, to describe the wine quality levels. Wines that receive a rating of 8 or above were considered normal (SL0). Ratings 7, 6, 5 are mapped to SL1, SL2, and SL3, respectively. Wines with a rating of 4 or below are considered SL4. As illustrated in Figure~\ref{fig:dataset-layout}, the training and test sets generated from the wine dataset follow a similar layout as with the chiller dataset.

There are two major differences between the chiller dataset and the wine dataset: 1) the normal subset of the wine data has far fewer data points than the chiller dataset, and 2) there is not an ``\ac{OOD} SL0'' subset in the wine dataset. More details about our preprocessing steps of the wine dataset will be given in Section~\ref{sec:wine-dataset-appendix} in the appendix.

\section{Experimental Results}\label{sec:experiment}

Real-world \ac{ML} practitioners perform extensive \textit{model selection} to search for models that are likely to perform well on training and test sets. In our empirical study, we will employ a similar workflow. For each type of model under study, we will conduct an extensive sweeping of hyperparameter settings to assess whether or not our proposed method can deliver consistent performance improvement. It is worthy to note that, our model selection process is purely based on available training data (SL0 \& SL4), and does not have access to the model's performance on \ac{IS} faults.
More details about our experimental setup and implementations will be given in Section~\ref{sec:app-experimental-setup} in the appendix.

Since we care about the performance improvement that is achievable by using model selection procedures with available data, the performance metrics to be shown next will be the distributions of the top $K$ selected models\footnote{If there are fewer than $K$ models before selection, we will use all of them.}. We will use $K = 100$ for all the results to be reported next. For visualizing the distributions of the selected models, we will plot the $95\%$ \ac{CI} of the performance indices. The \ac{CI} is computed with the bootstrap method by repeatedly sampling for $5000$ times.

Since there is naturally a tradeoff between \ac{FPR} and \ac{FNR}, two classification models are not directly comparable unless one dominates the other in both aspects. To get a clearer understanding of the performance distributions of the generated classifier instances, we first group them into different bins by their \ac{FPR} on the training data, and then plot the performance indices of the $K$ models with lowest \ac{FNR} (highest sensitivity). The reason for binning classifiers first by their \ac{FPR} is because \ac{FDD} systems usually prioritize high specificity (as few false alarms as possible) over sensitivity. The detection rates (proportion of data points detected as faults) of the three types of supervised \ac{FDD} models (\textsc{mean} is used as the aggregation function) on the two datasets are plotted in Figure~\ref{fig:FPR-sweep}. Also, In Figure~\ref{fig:FPR-hist} in the appendix, the number of models falling under each \ac{FPR} interval is given.

We can see from the left panels of Figure~\ref{fig:FPR-sweep} that all three supervised models, especially those using \textsc{cfar}, can achieve high sensitivity and specificity on in-distribution data, across a wide range of \ac{FPR} levels. In comparison, models using \textsc{cfar} deliver superior performance than those using \textsc{argmax} on in-distribution (SL4) and \ac{OOD} faults (SL1--SL3), which is probably due to the increased statistical power (sensitivity) in models with \textsc{cfar}. For the upcoming plots and analyses on the chiller dataset, we will use $1\%$ as the \ac{FPR} threshold for model selection. 
Similar plots for the wine dataset are shown in the right panels of Figure~\ref{fig:FPR-sweep}. It can be seen that it is more difficult to achieve high accuracy for the wine dataset. For the wine dataset, we will use $10\%$ as the \ac{FPR} threshold for model selection. 

\begin{figure}[tb]
  \centering
  \begin{subfigure}[t]{0.49\linewidth}
    \centering
    \includegraphics[height=3.3cm]{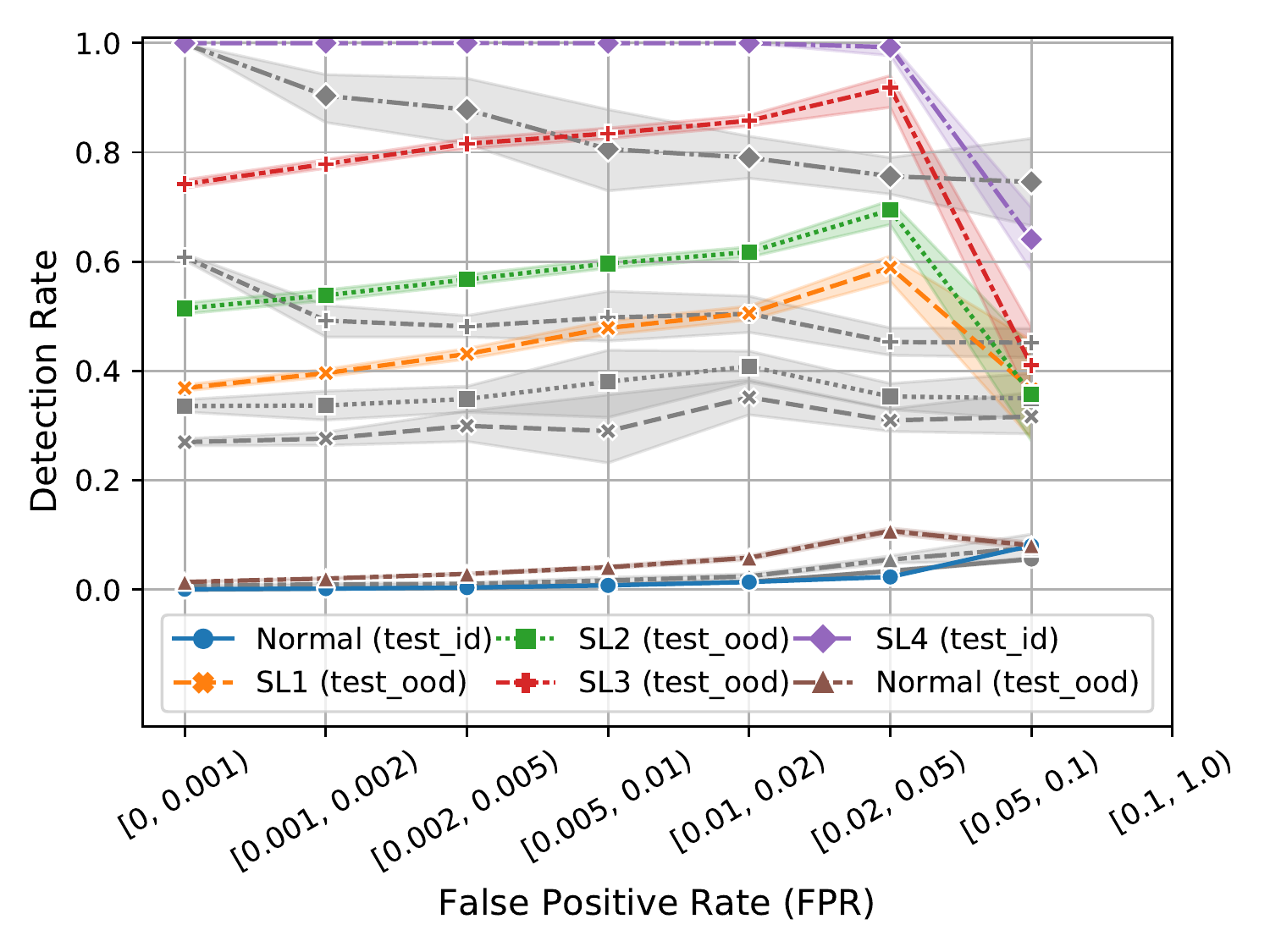}
    \caption{\ac{RF}: chiller}
  \end{subfigure}
  \begin{subfigure}[t]{0.49\linewidth}
    \centering
    \includegraphics[height=3.3cm]{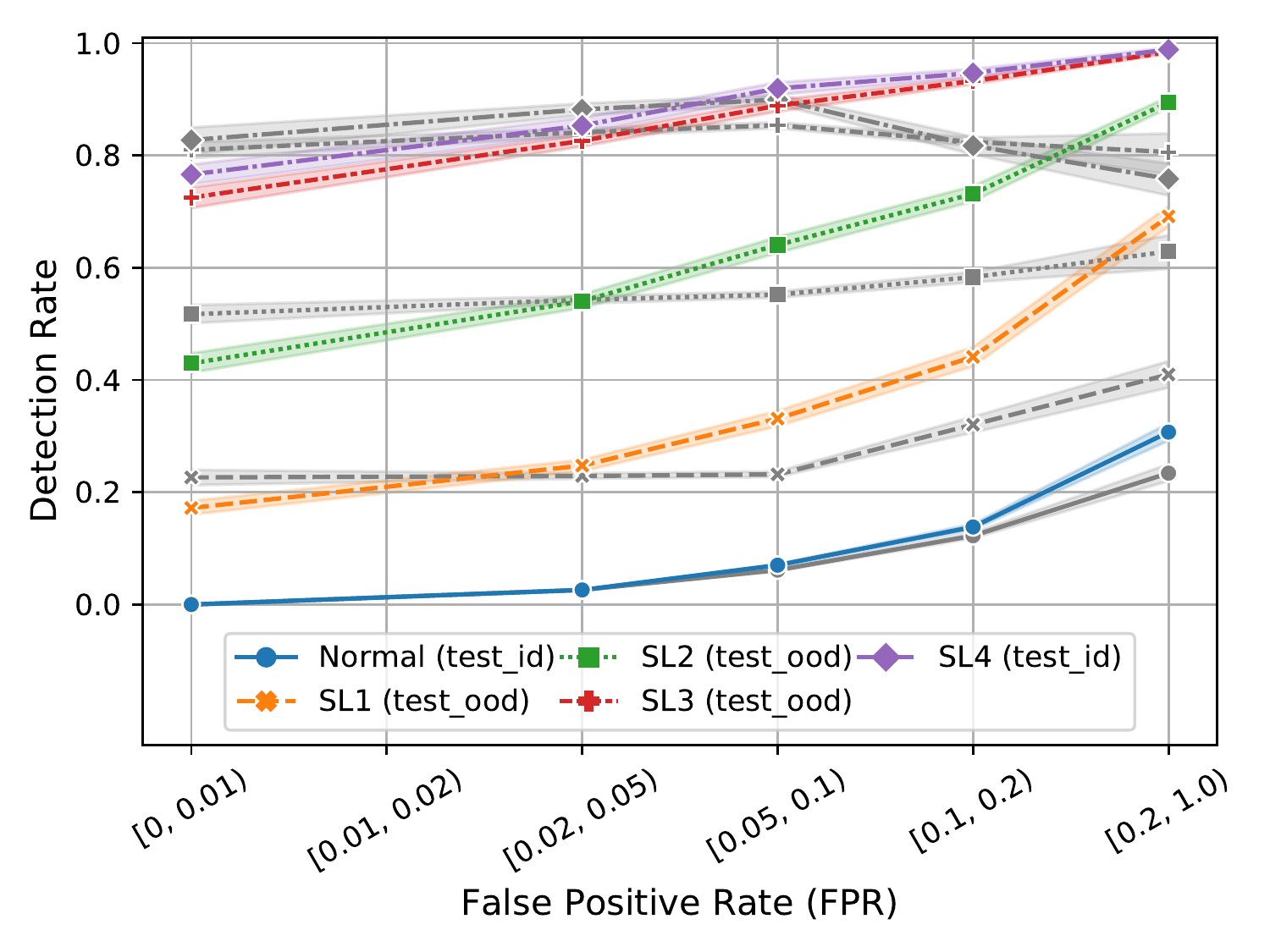}
    \caption{\ac{RF}: wine}
  \end{subfigure}
  
  \begin{subfigure}[t]{0.49\linewidth}
    \centering
    \includegraphics[height=3.3cm]{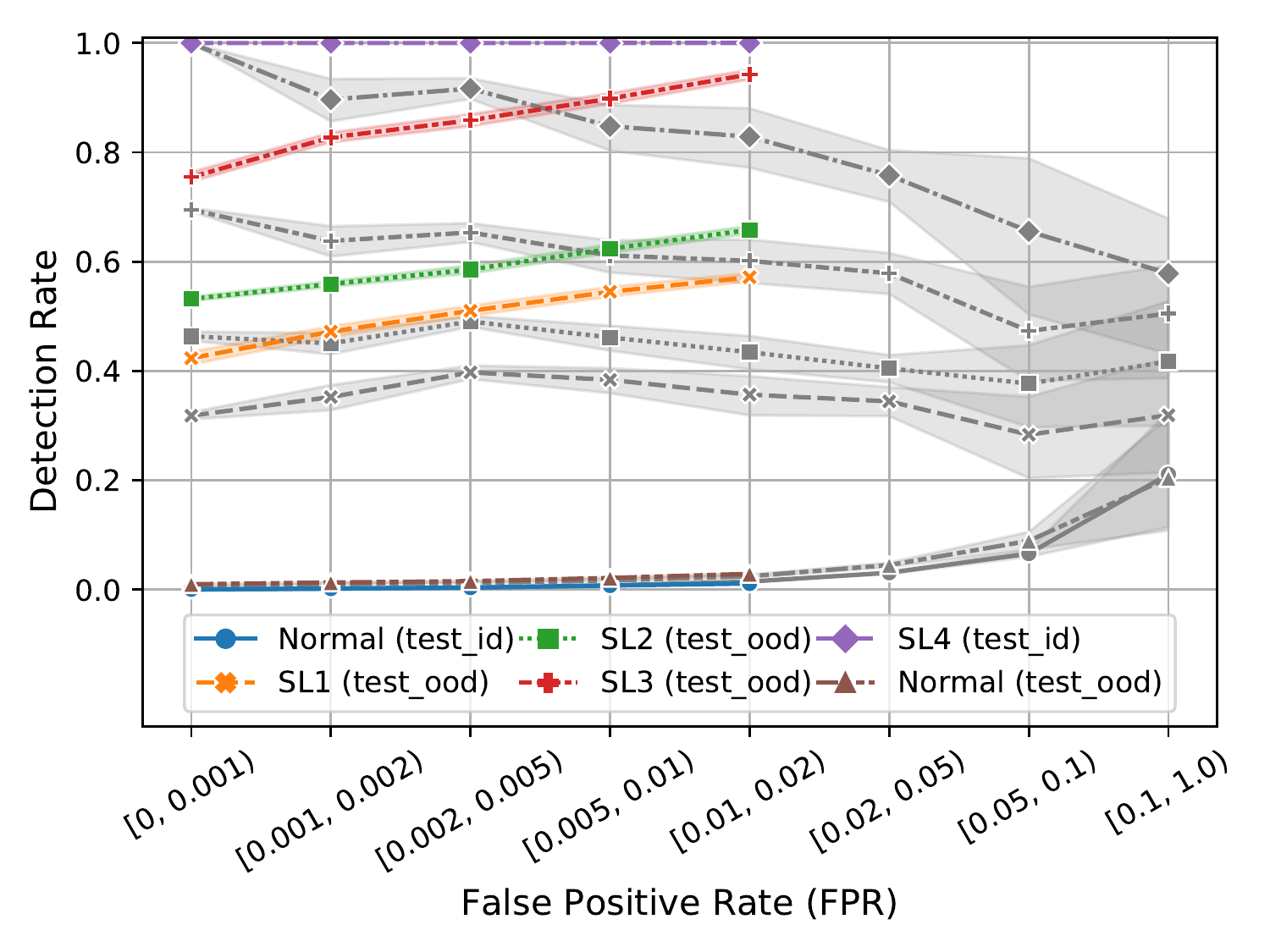}
    \caption{\ac{NN}: chiller}
  \end{subfigure}
  \begin{subfigure}[t]{0.49\linewidth}
    \centering
    \includegraphics[height=3.3cm]{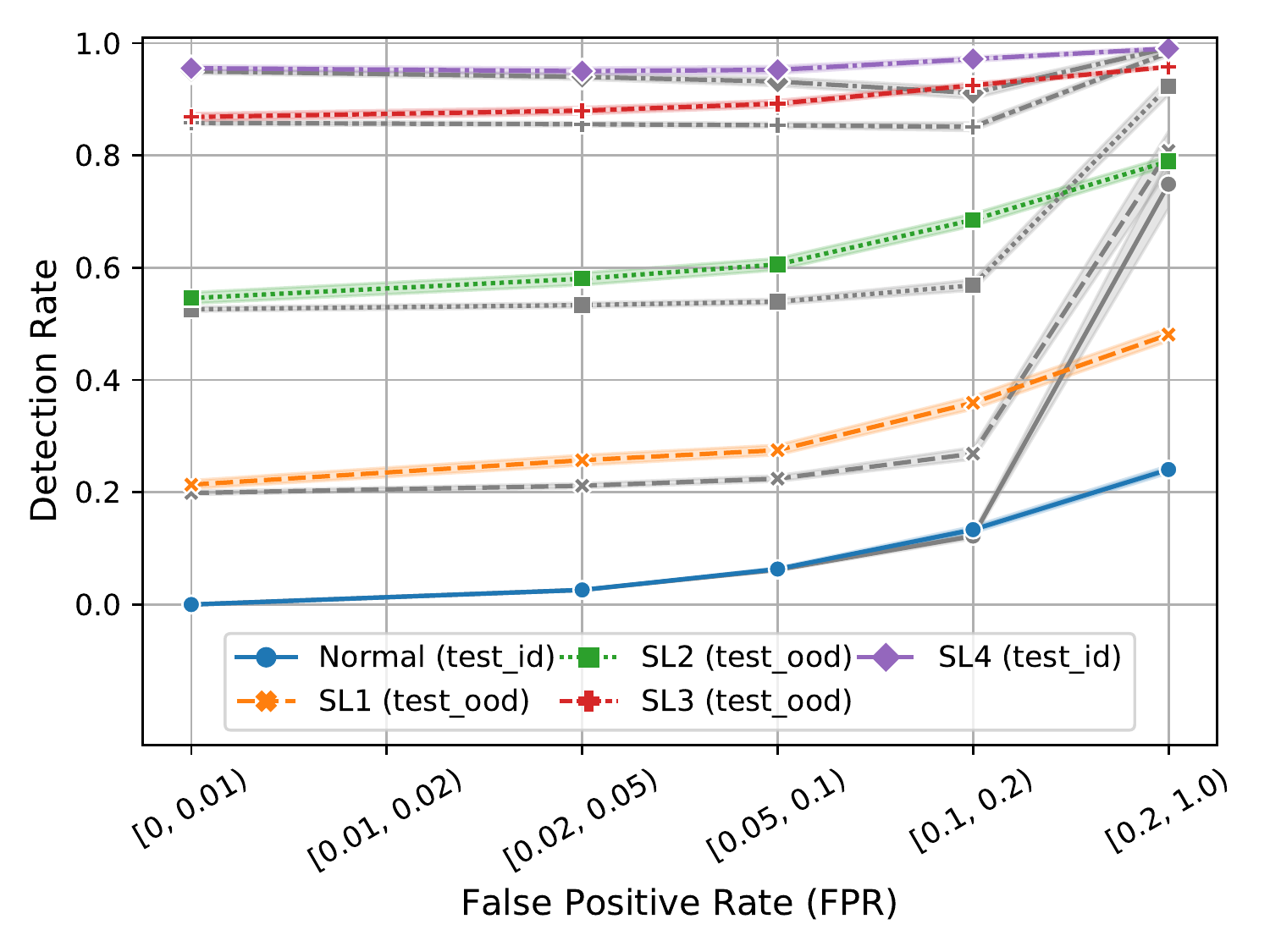}
    \caption{\ac{NN}: wine}
  \end{subfigure}
  
  \begin{subfigure}[t]{0.49\linewidth}
    \centering
    \includegraphics[height=3.3cm]{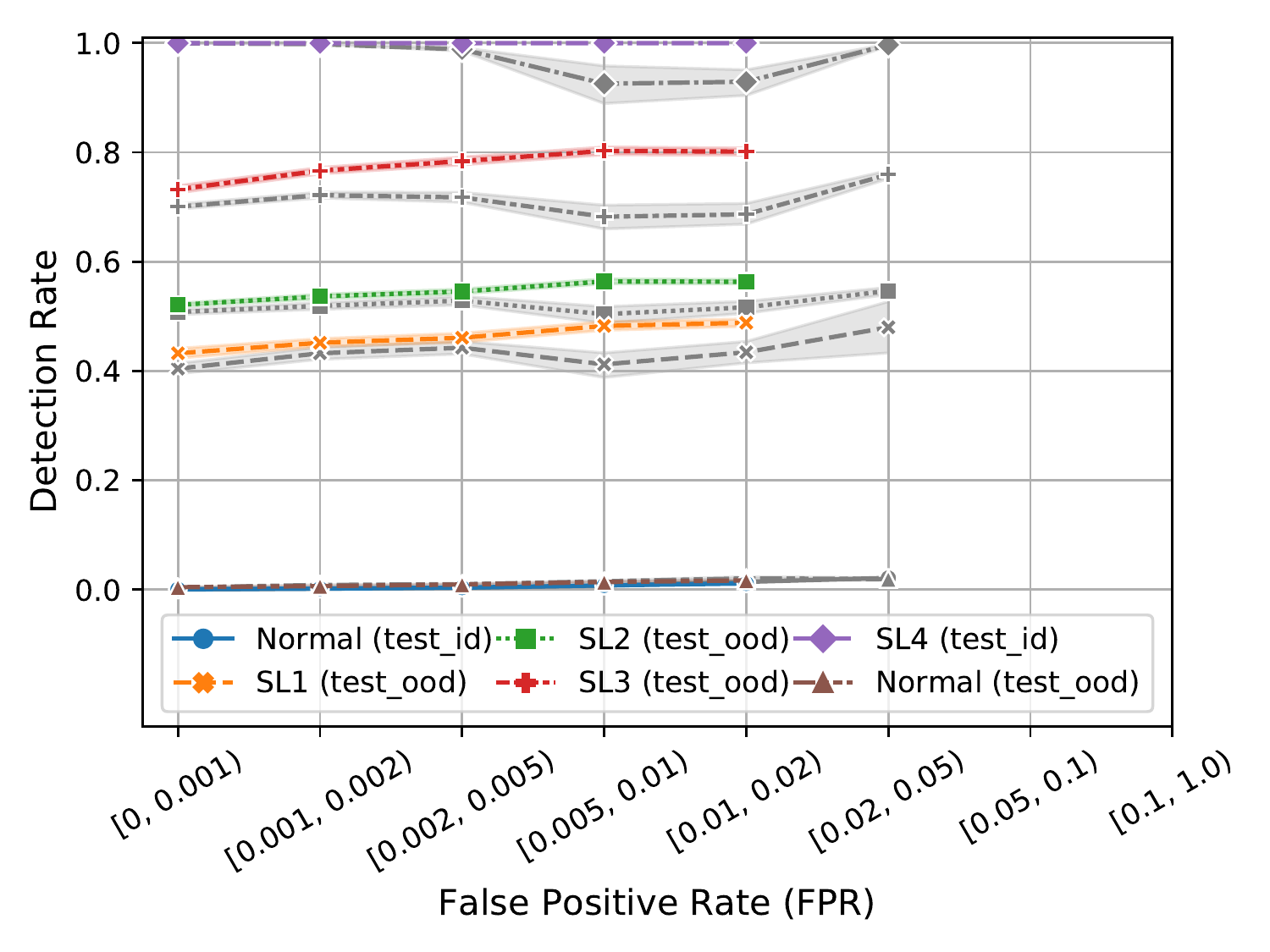}
    \caption{Dropout \ac{NN}: chiller}
  \end{subfigure}
  \begin{subfigure}[t]{0.49\linewidth}
    \centering
    \includegraphics[height=3.3cm]{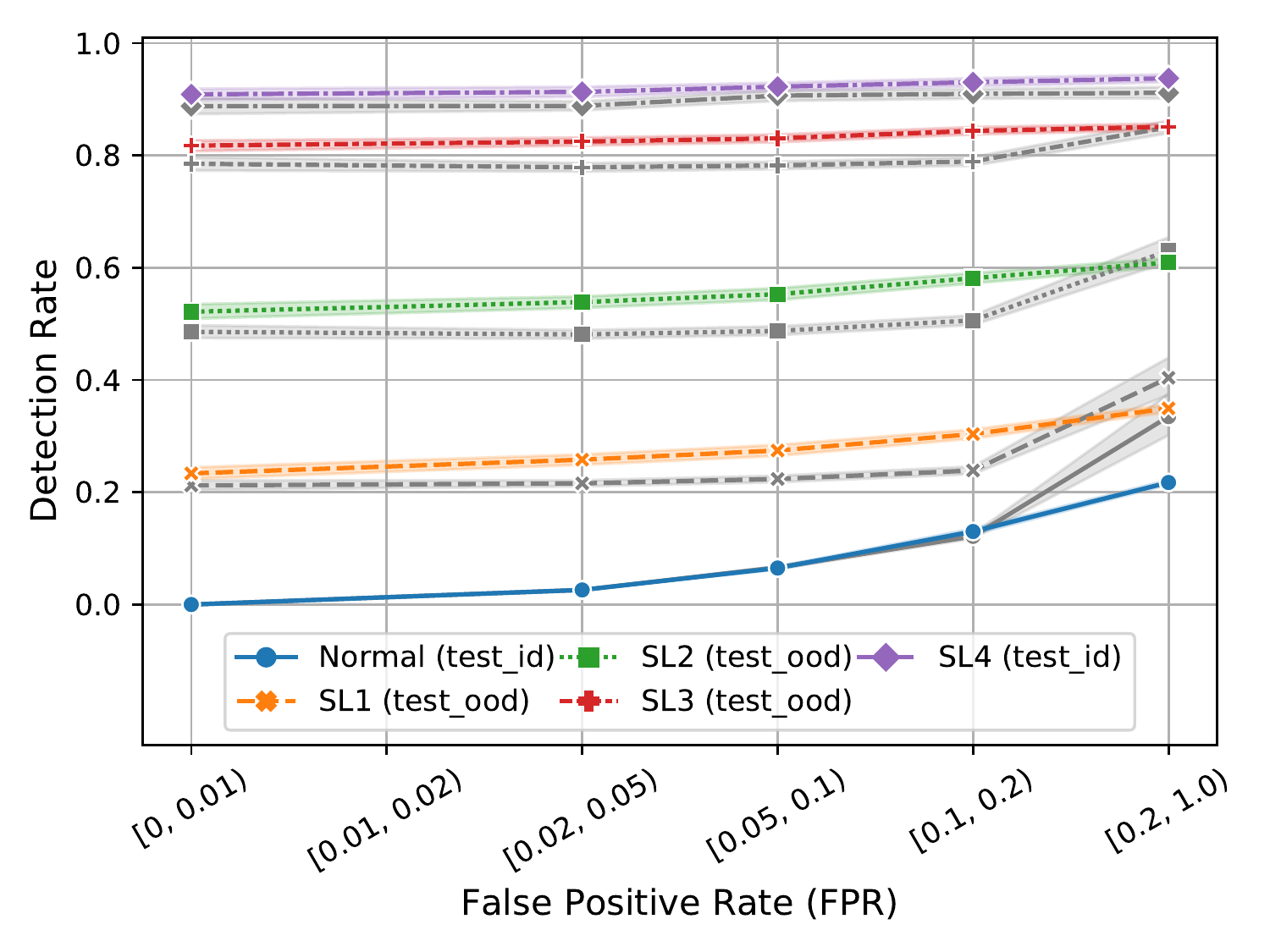}
    \caption{Dropout \ac{NN}: wine}
  \end{subfigure}
  
  \caption{Performances of the selected supervised classifiers under different \ac{FPR} intervals. The performance indices for models using \textsc{cfar} are shown in colors, and indices for models using \textsc{argmax} are shown in gray.}
  \label{fig:FPR-sweep}
\end{figure}
\begin{figure*}[tb]
  \centering
  \begin{subfigure}[t]{0.245\linewidth}
    \centering
    \includegraphics[height=3.4cm]{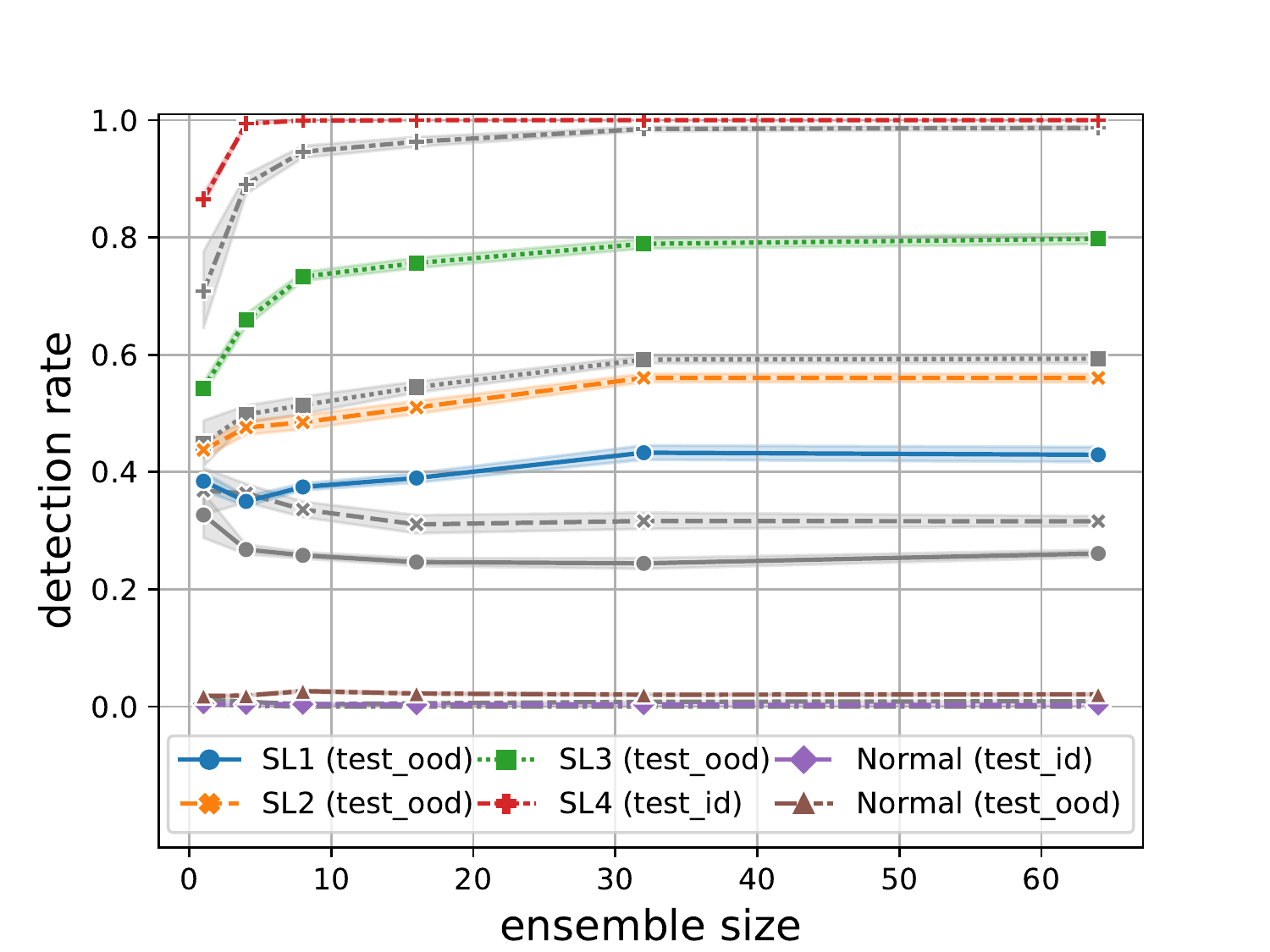}
    \caption{\ac{RF}: \textsc{mean}}
    \label{fig:chiller_DT_detect_mean}
  \end{subfigure}  
  \begin{subfigure}[t]{0.245\linewidth}
    \centering
    \includegraphics[height=3.4cm]{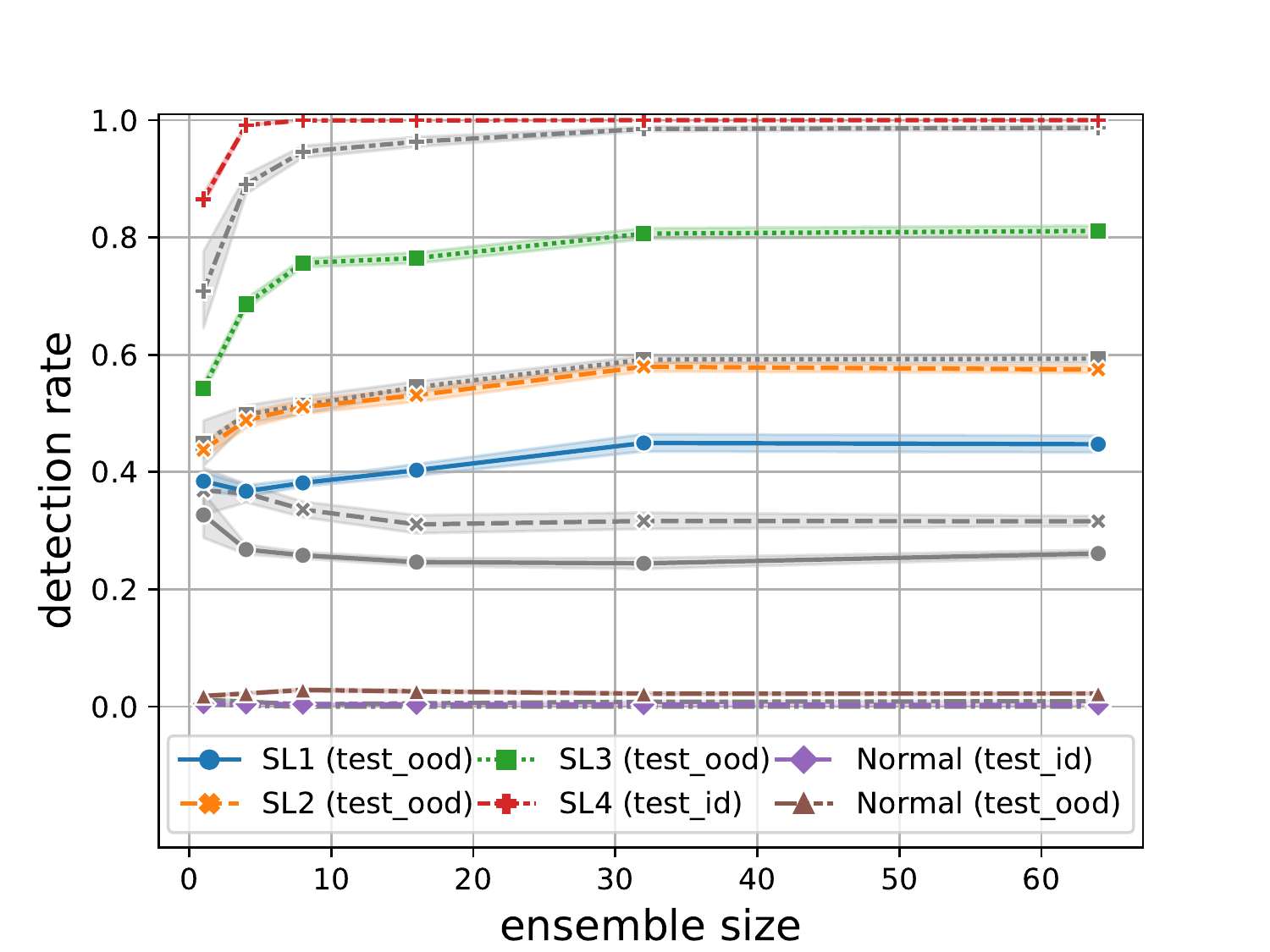}
    \caption{\ac{RF}: \textsc{mean-std}}
    \label{fig:chiller_DT_detect_meanvar}
  \end{subfigure}
  \begin{subfigure}[t]{0.245\linewidth}
    \centering
    \includegraphics[height=3.4cm]{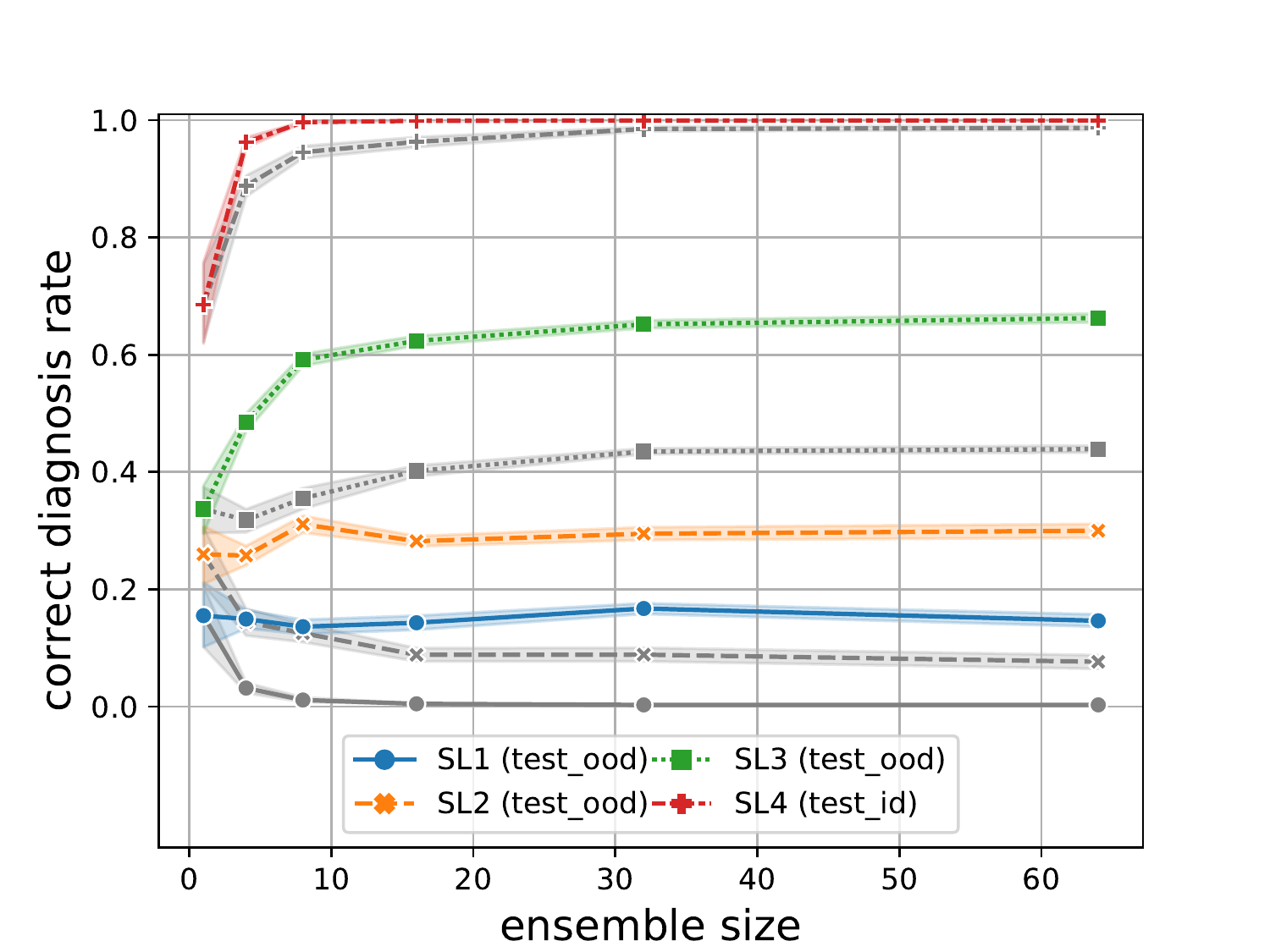}
    \caption{RF: top-1 accuracy}
  \end{subfigure}
  \begin{subfigure}[t]{0.245\linewidth}
    \centering
    \includegraphics[height=3.4cm]{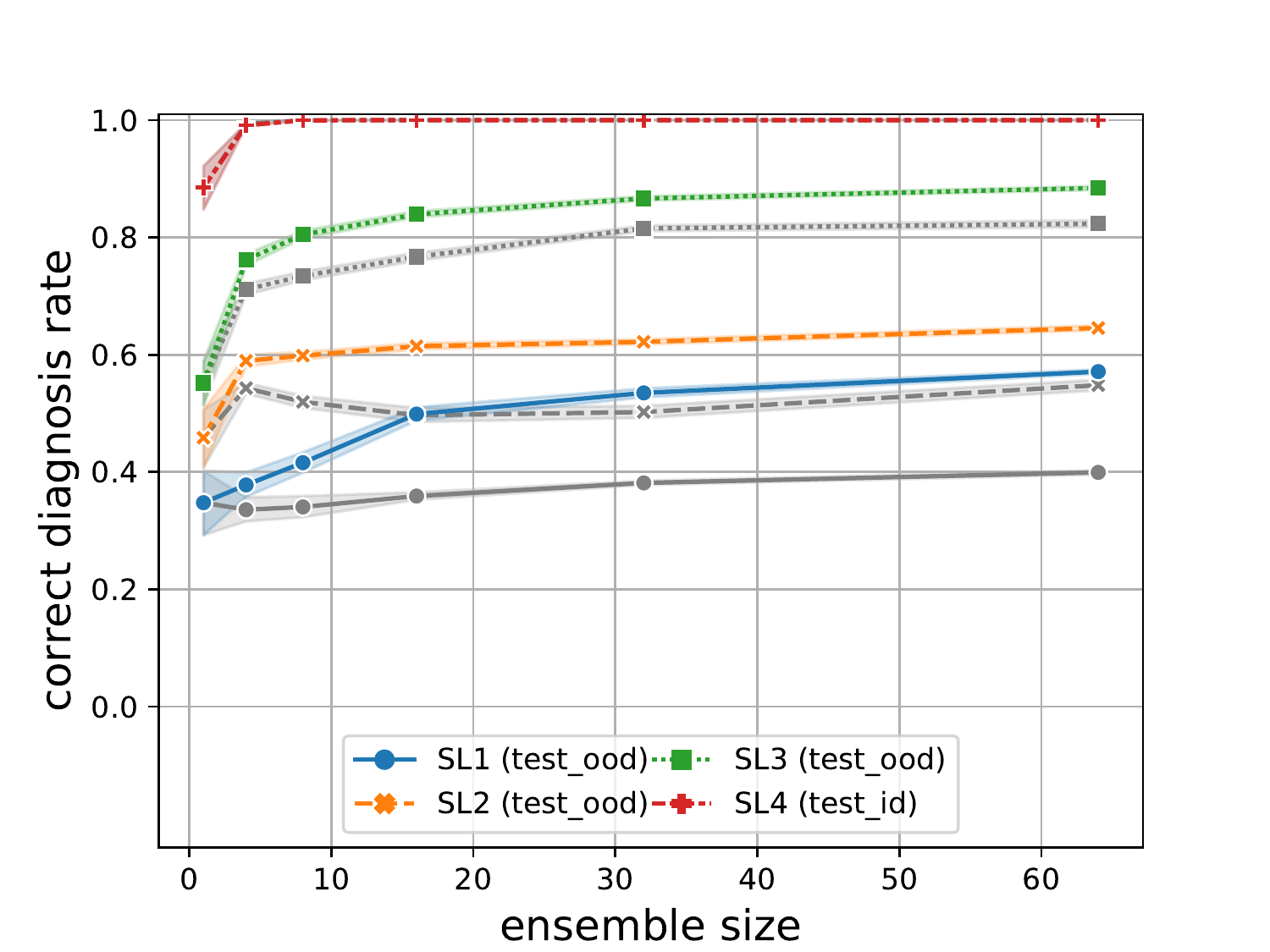}
    \caption{\ac{RF}: top-2 accuracy}
  \end{subfigure}
  
  \begin{subfigure}[t]{0.245\linewidth}
    \centering
    \includegraphics[height=3.4cm]{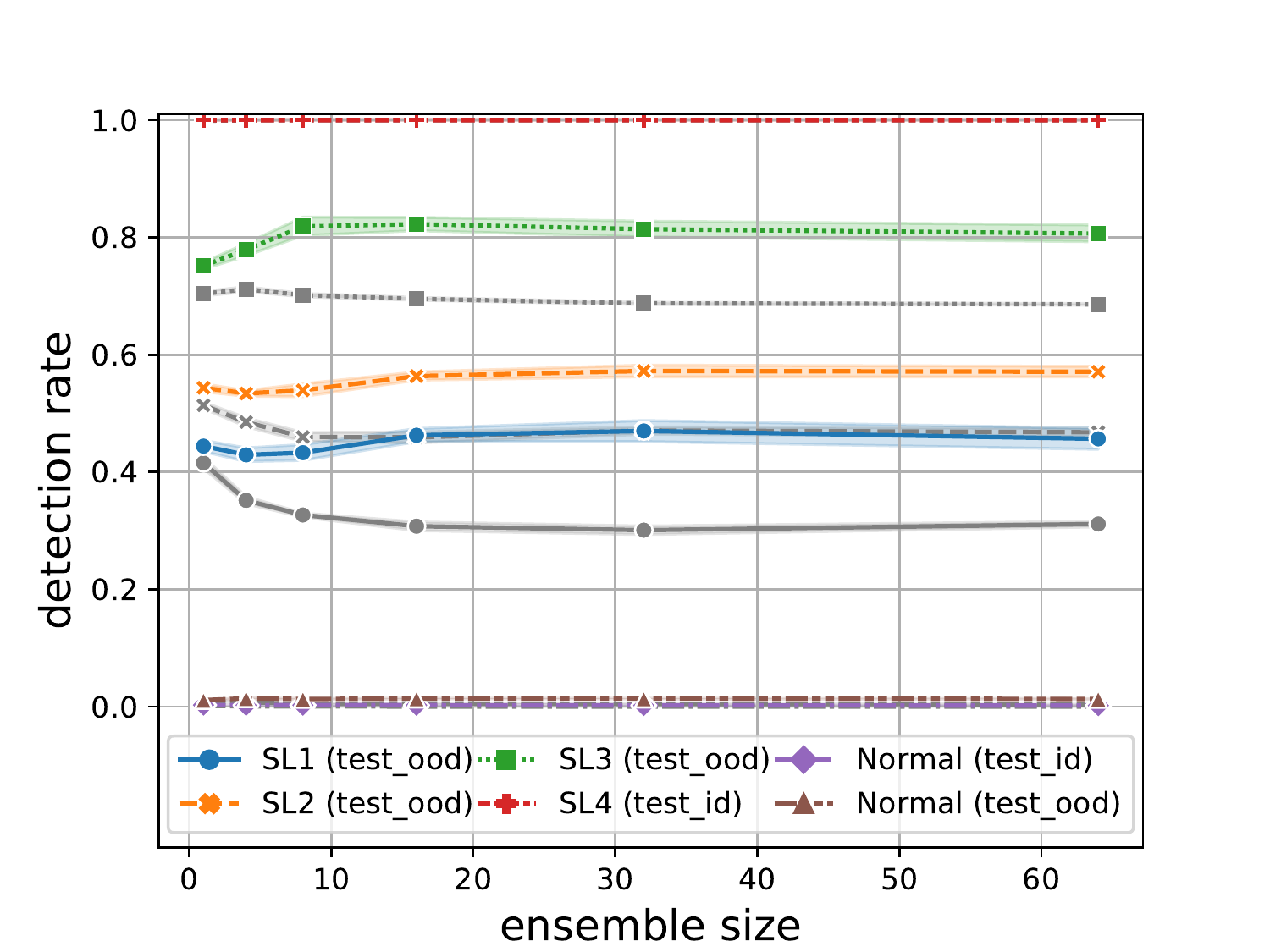}
    \caption{\ac{NN}: \textsc{mean}}
    \label{fig:chiller_NN_detect_mean}
  \end{subfigure}
  \begin{subfigure}[t]{0.245\linewidth}
    \centering
    \includegraphics[height=3.4cm]{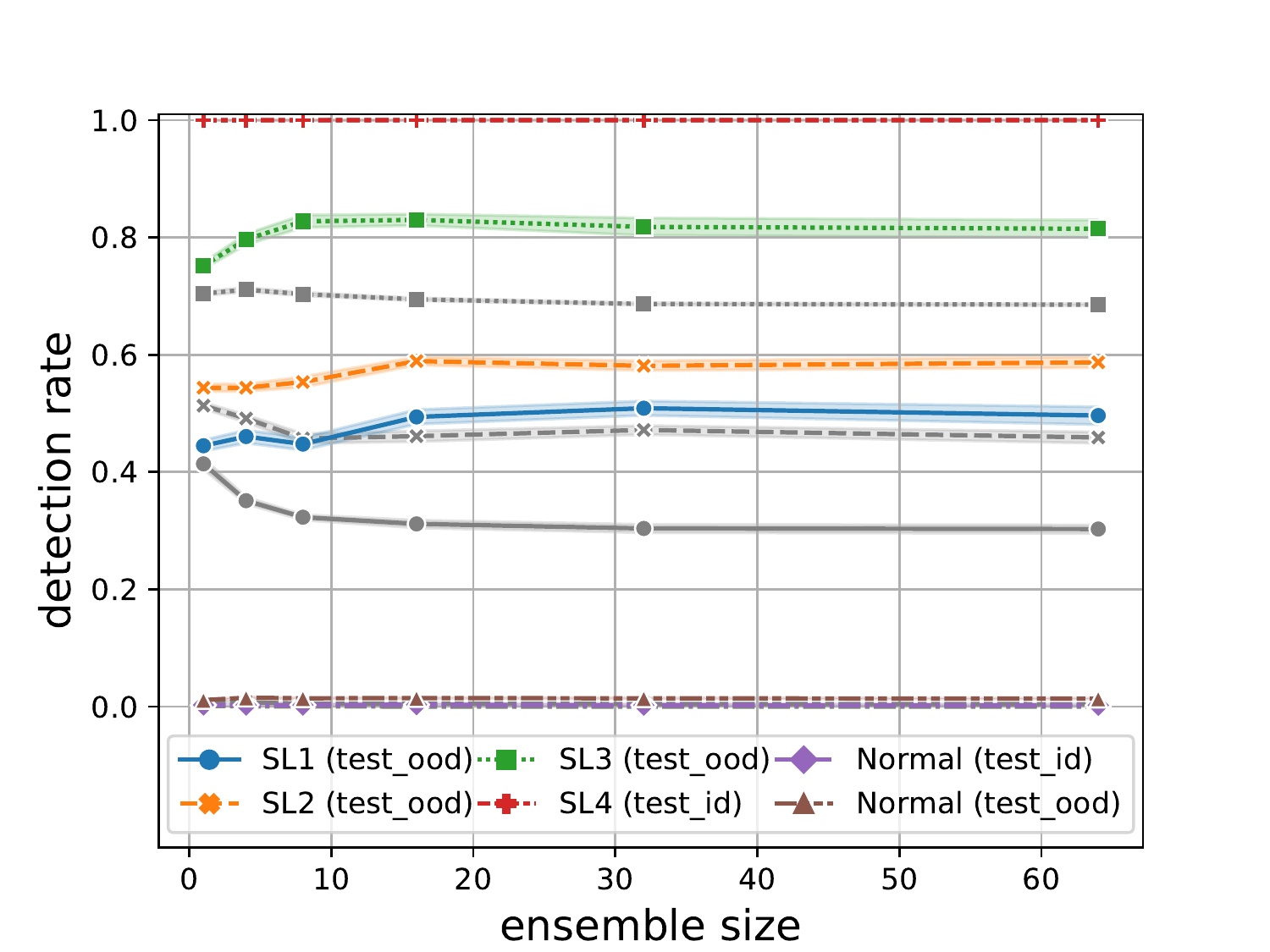}
    \caption{\ac{NN}: \textsc{mean-std}}
    \label{fig:chiller_NN_detect_meanvar}
  \end{subfigure}
  \begin{subfigure}[t]{0.245\linewidth}
    \centering
    \includegraphics[height=3.4cm]{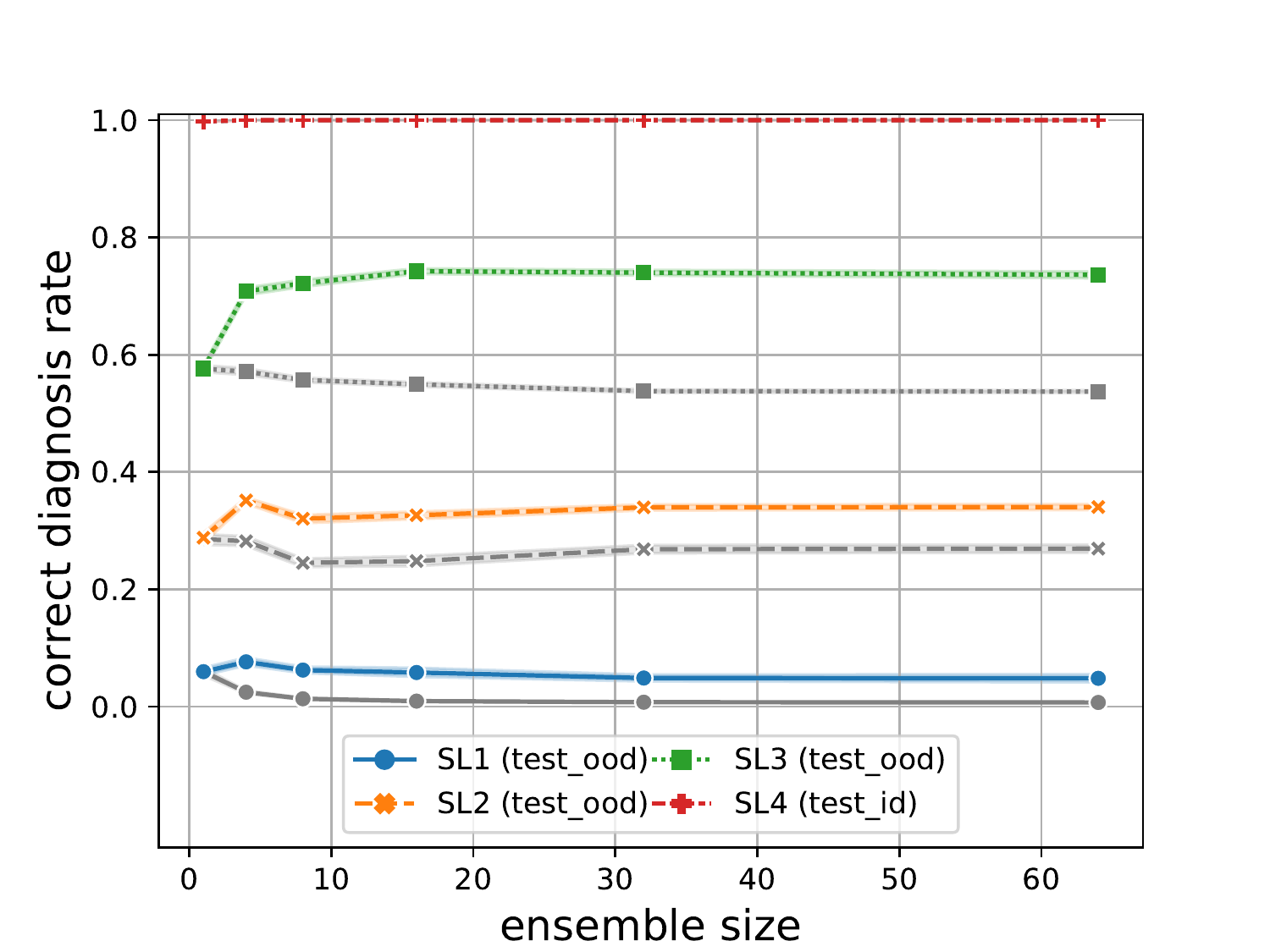}
    \caption{NN: top-1 accuracy}
  \end{subfigure}
  \begin{subfigure}[t]{0.245\linewidth}
    \centering
    \includegraphics[height=3.4cm]{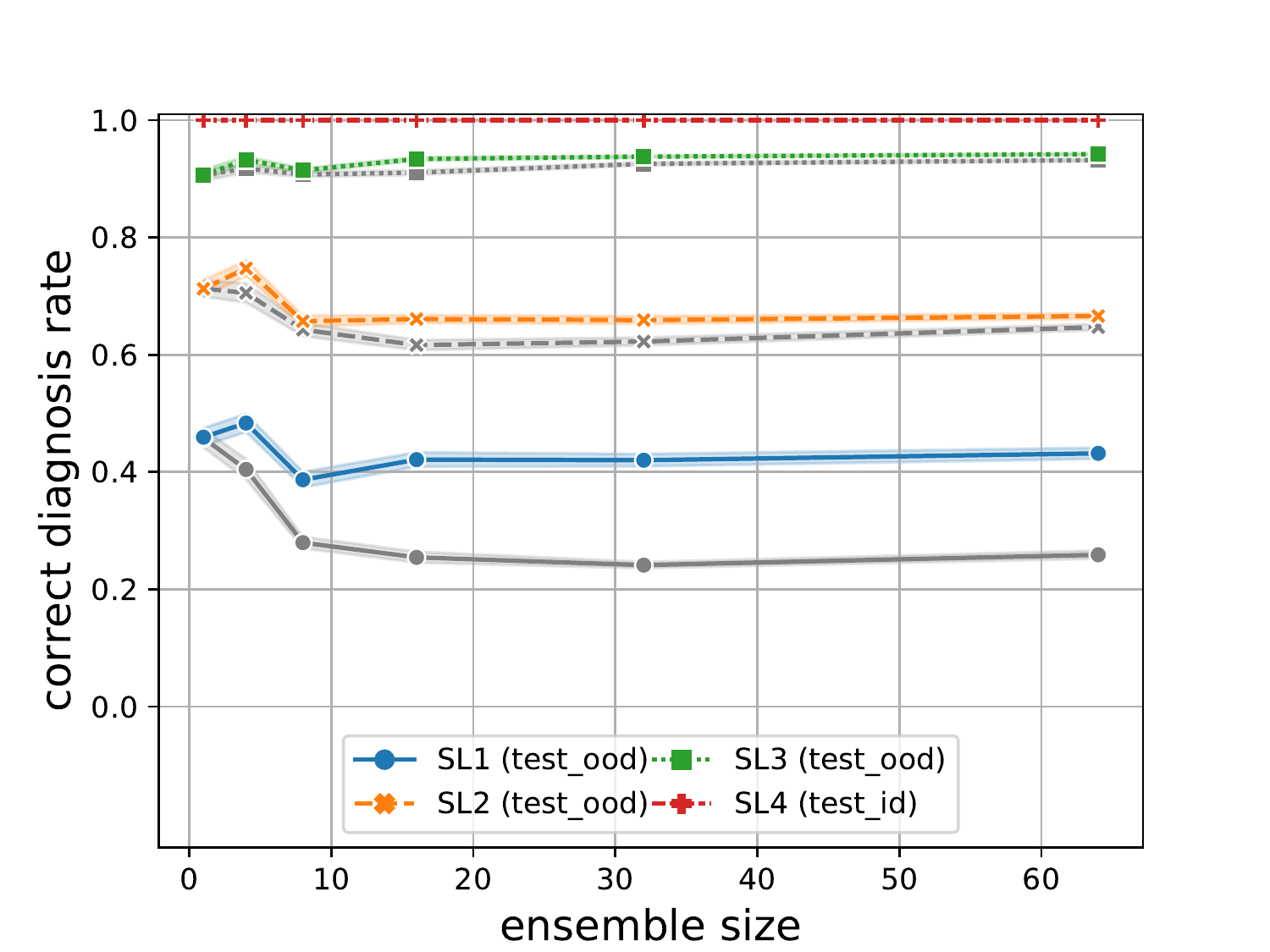}
    \caption{\ac{NN}: top-2 accuracy}
  \end{subfigure}
  
  \begin{subfigure}[t]{0.245\linewidth}
    \centering
    \includegraphics[height=3.4cm]{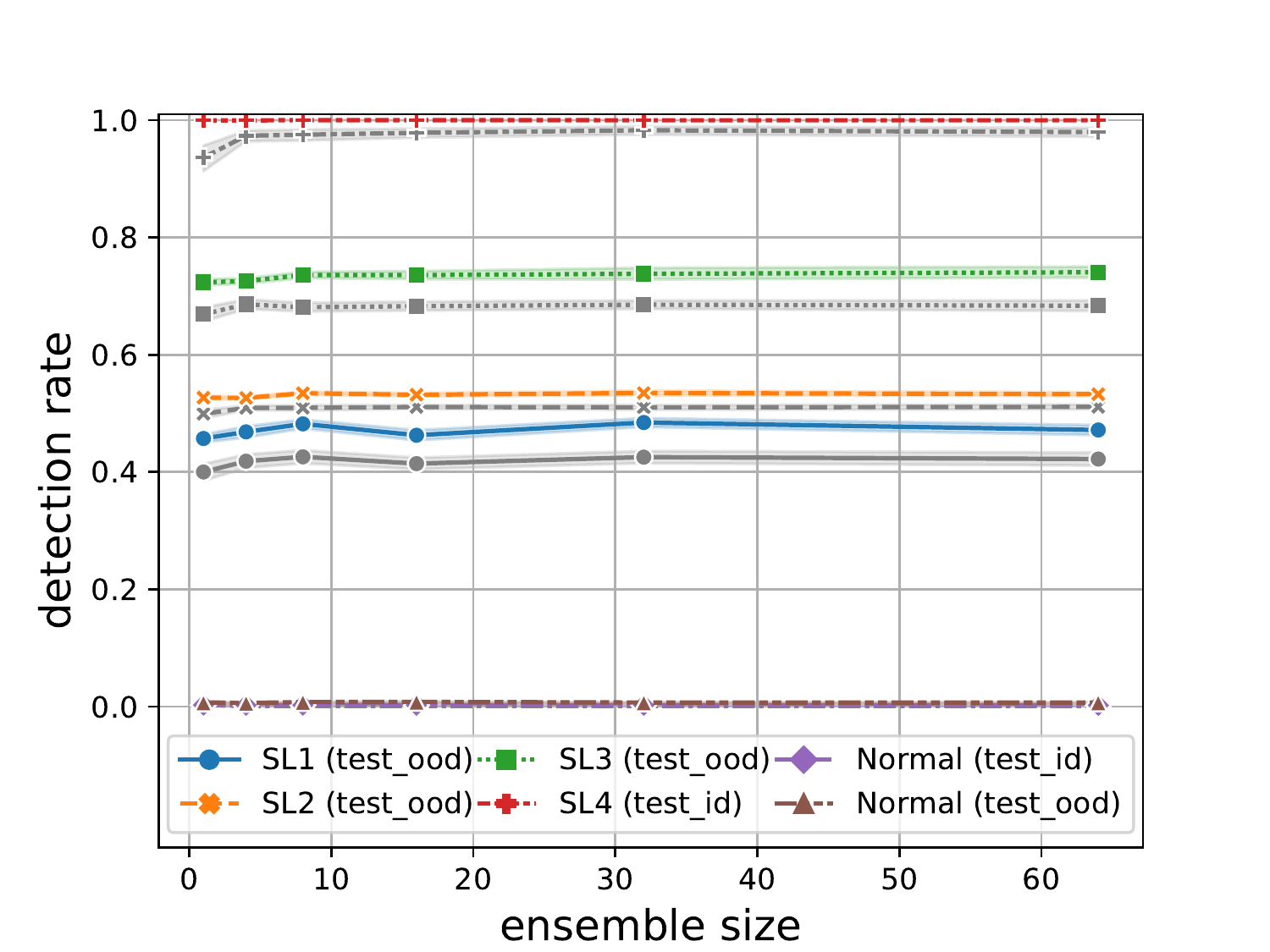}
    \caption{Dropout \ac{NN}: \textsc{mean}}
    \label{fig:chiller_DPNN_detect_mean}
  \end{subfigure}
  \begin{subfigure}[t]{0.245\linewidth}
    \centering
    \includegraphics[height=3.4cm]{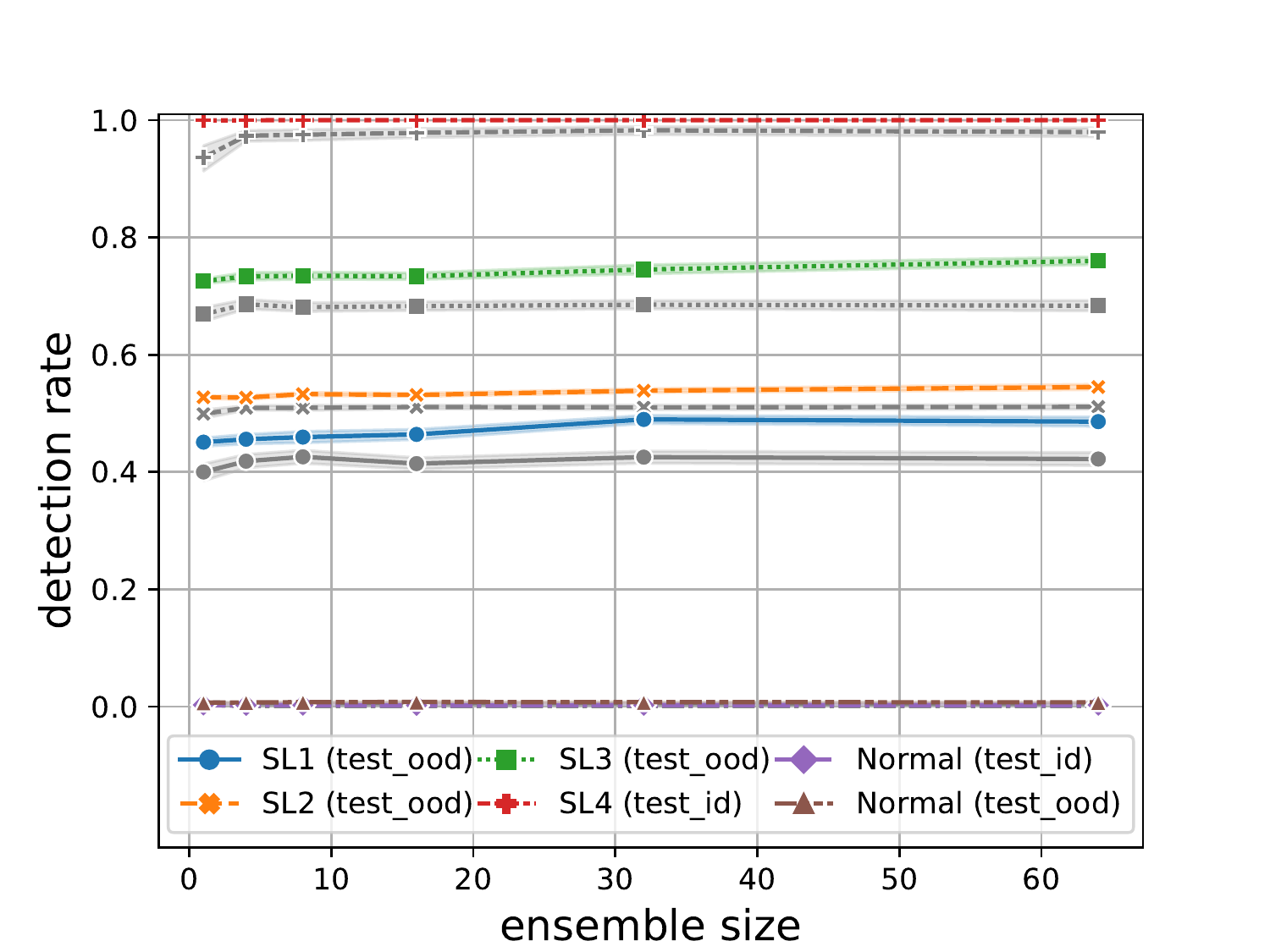}
    \caption{Dropout \ac{NN}: \textsc{mean-std}}
    \label{fig:chiller_DPNN_detect_meanvar}
  \end{subfigure}
  \begin{subfigure}[t]{0.245\linewidth}
    \centering
    \includegraphics[height=3.4cm]{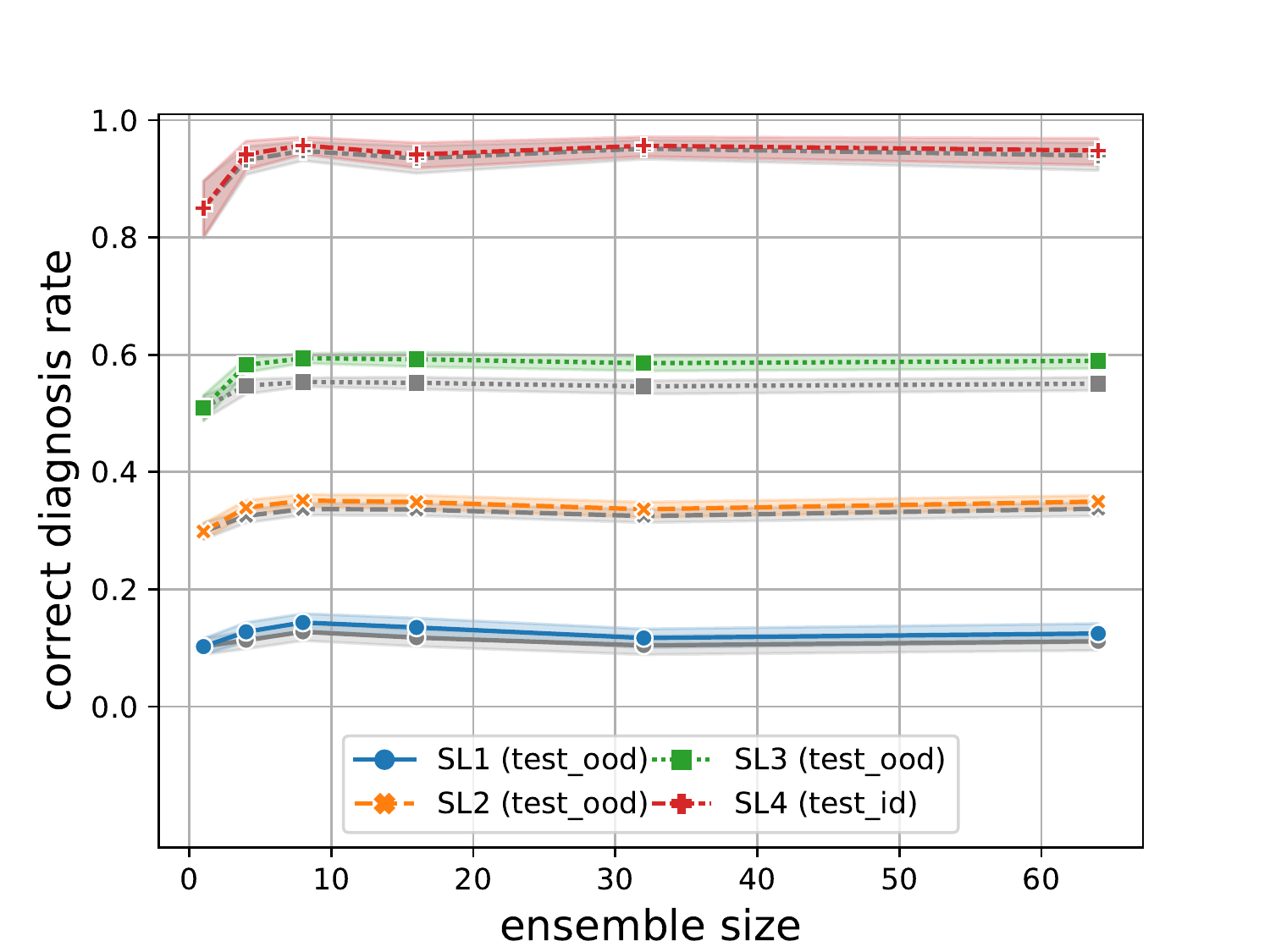}
    \caption{Dropout NN: top-1 accuracy}
    \label{fig:DPNN-top1-diagnosis}
  \end{subfigure}
  \begin{subfigure}[t]{0.245\linewidth}
    \centering
    \includegraphics[height=3.4cm]{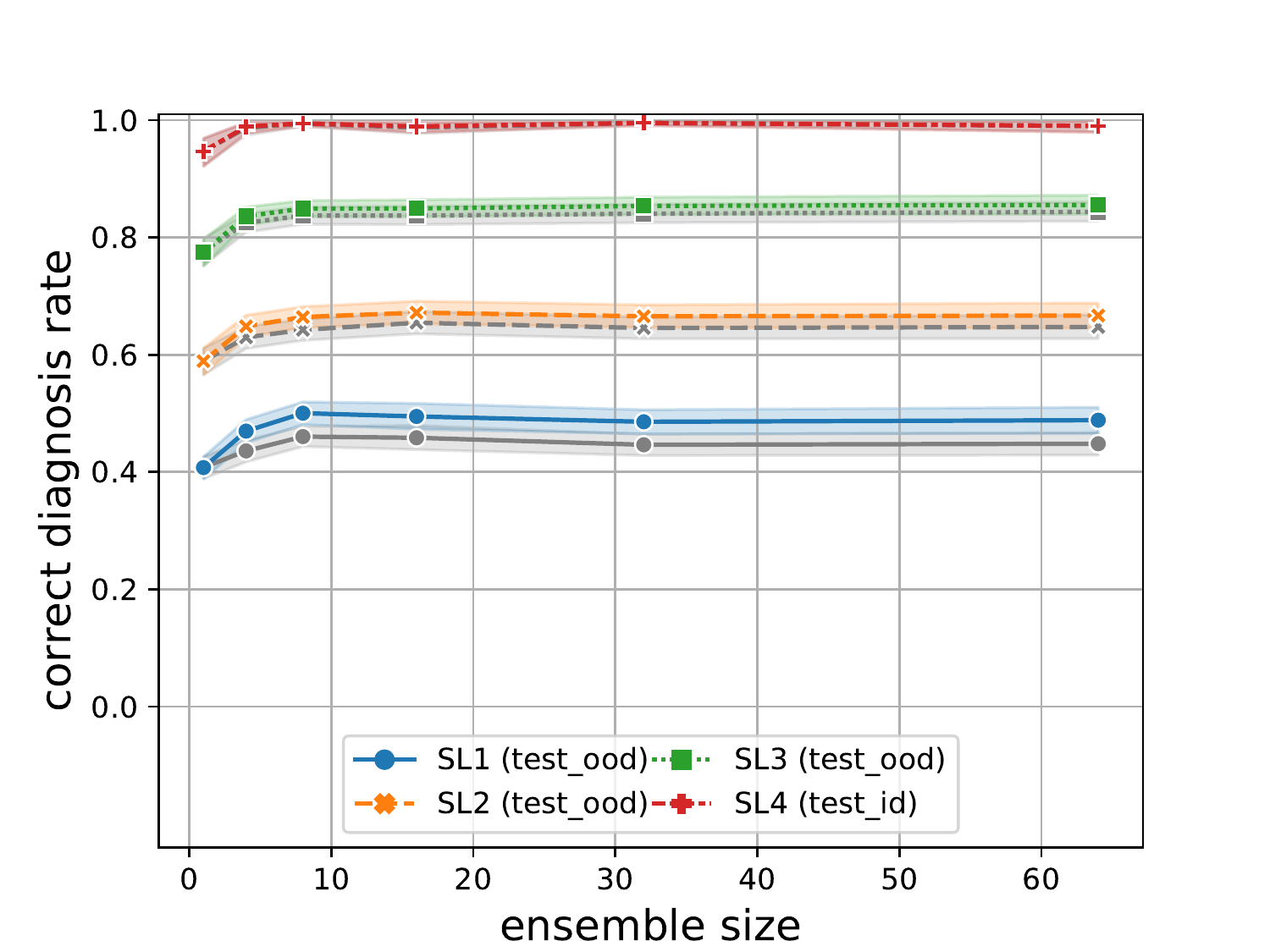}
    \caption{Dropout NN: top-2 accuracy}
    \label{fig:DPNN-top2-diagnosis}
  \end{subfigure}
  \caption{\ac{FDD} performance of ensemble models on the chiller dataset. In plots for detection performance, \textsc{cfar} (colored) and \textsc{argmax} (gray) are compared. In plots for diagnosis performance, \textsc{mean-std} (colored) and \textsc{mean} (gray) are compared.}
  \label{fig:detection-performance-chiller}
\end{figure*}
\begin{figure}[tb]
  \centering
  \begin{subfigure}[t]{0.49\linewidth}
    \centering
    \includegraphics[height=3.3cm]{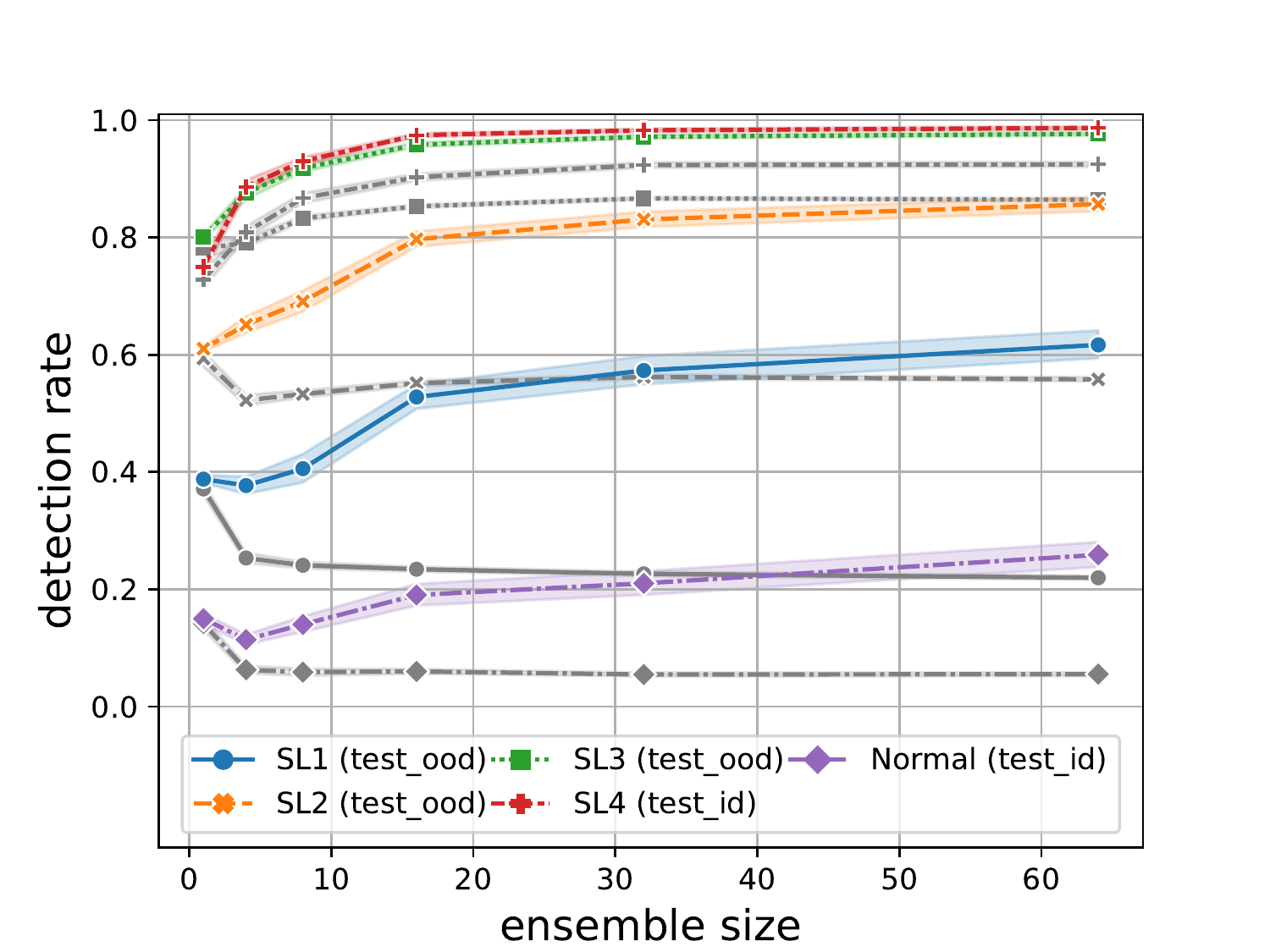}
    \caption{\ac{RF}: \textsc{mean}}
    \label{fig:wine_DT_detect_mean}
  \end{subfigure}  
  \begin{subfigure}[t]{0.49\linewidth}
    \centering
    \includegraphics[height=3.3cm]{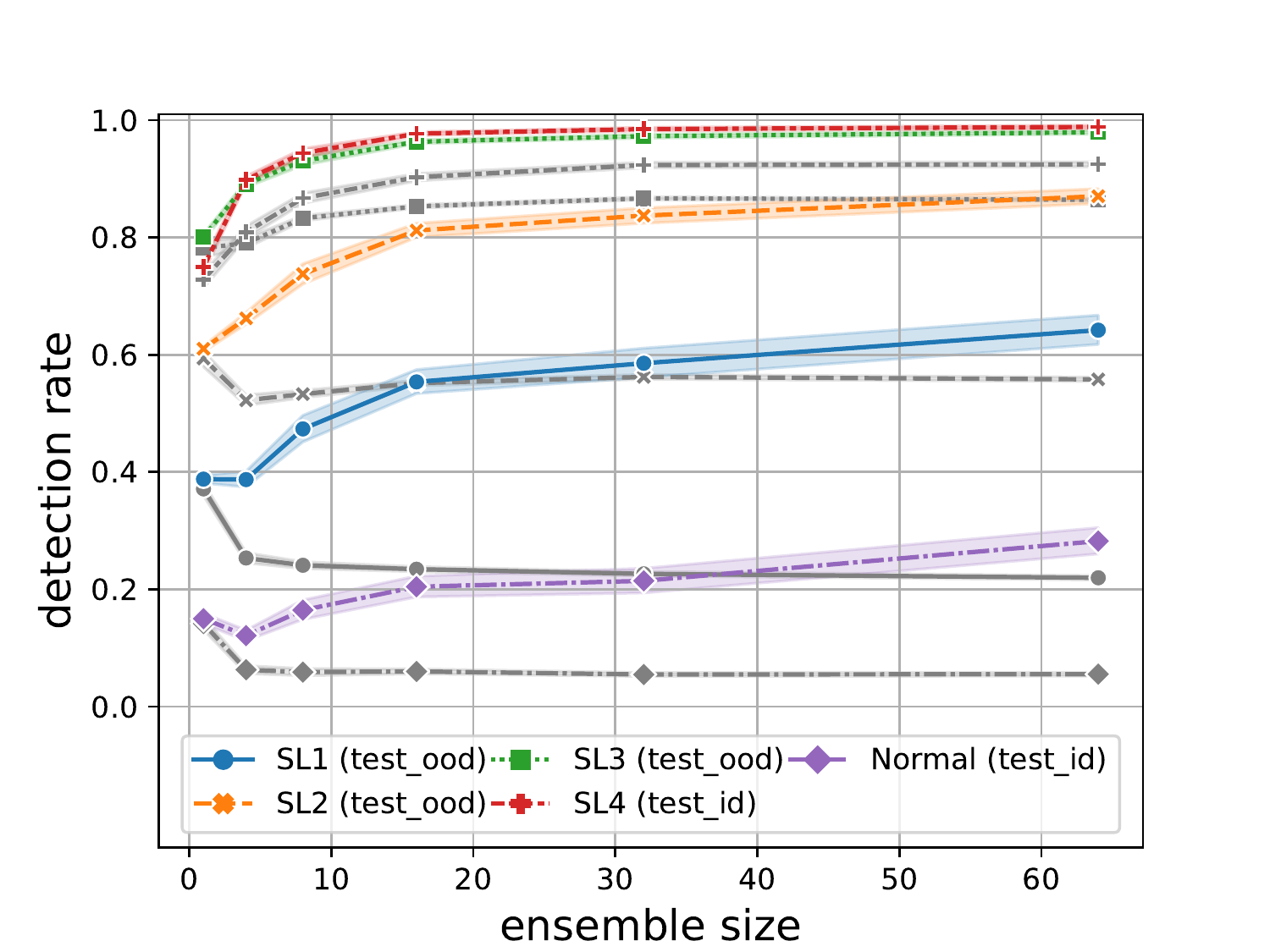}
    \caption{\ac{RF}: \textsc{mean-std}}
    \label{fig:wine_DT_detect_meanvar}
  \end{subfigure}
  
  \begin{subfigure}[t]{0.49\linewidth}
    \centering
    \includegraphics[height=3.3cm]{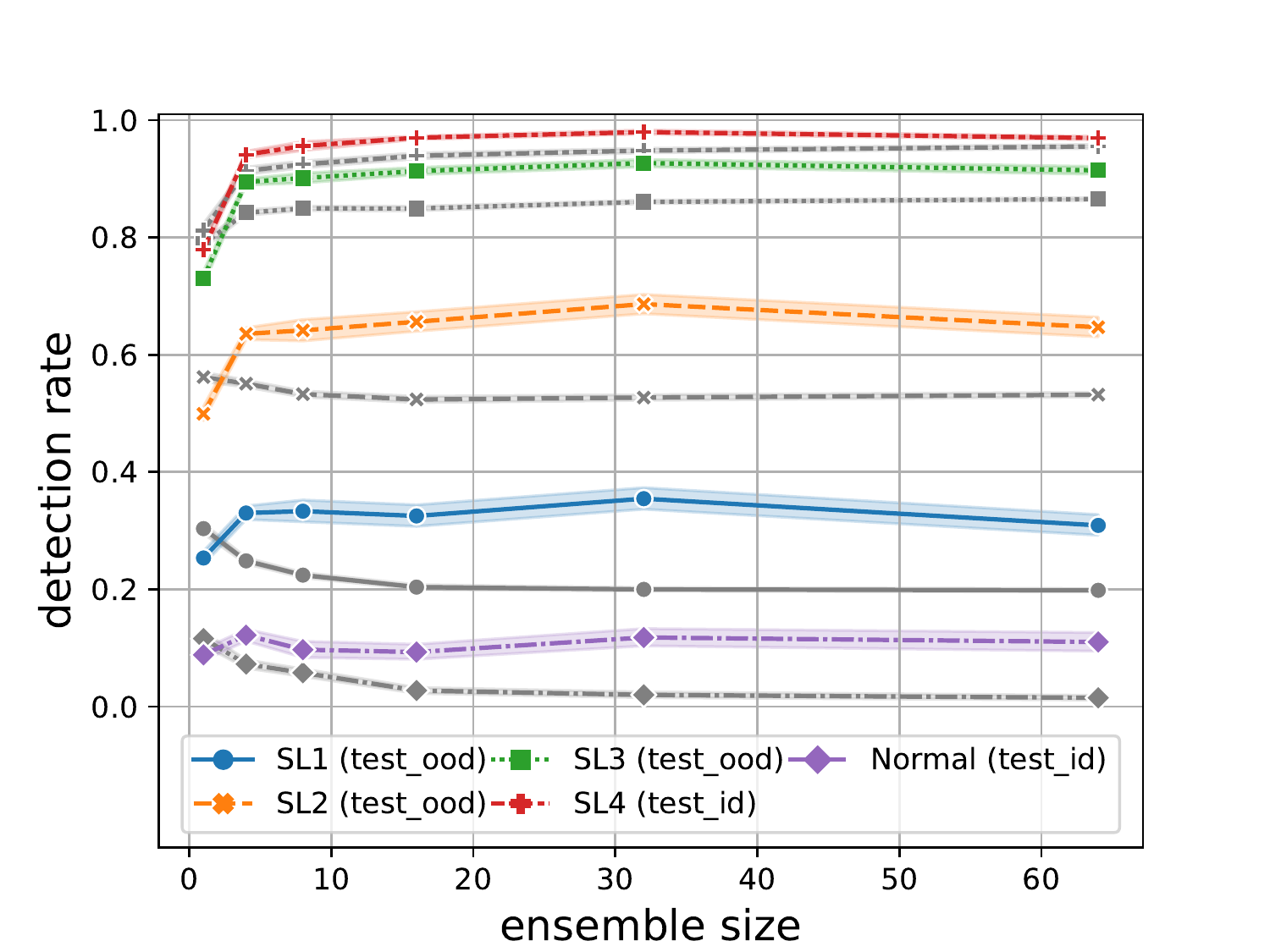}
    \caption{\ac{NN}: \textsc{mean}}
    \label{fig:wine_NN_detect_mean}
  \end{subfigure}
  \begin{subfigure}[t]{0.49\linewidth}
    \centering
    \includegraphics[height=3.3cm]{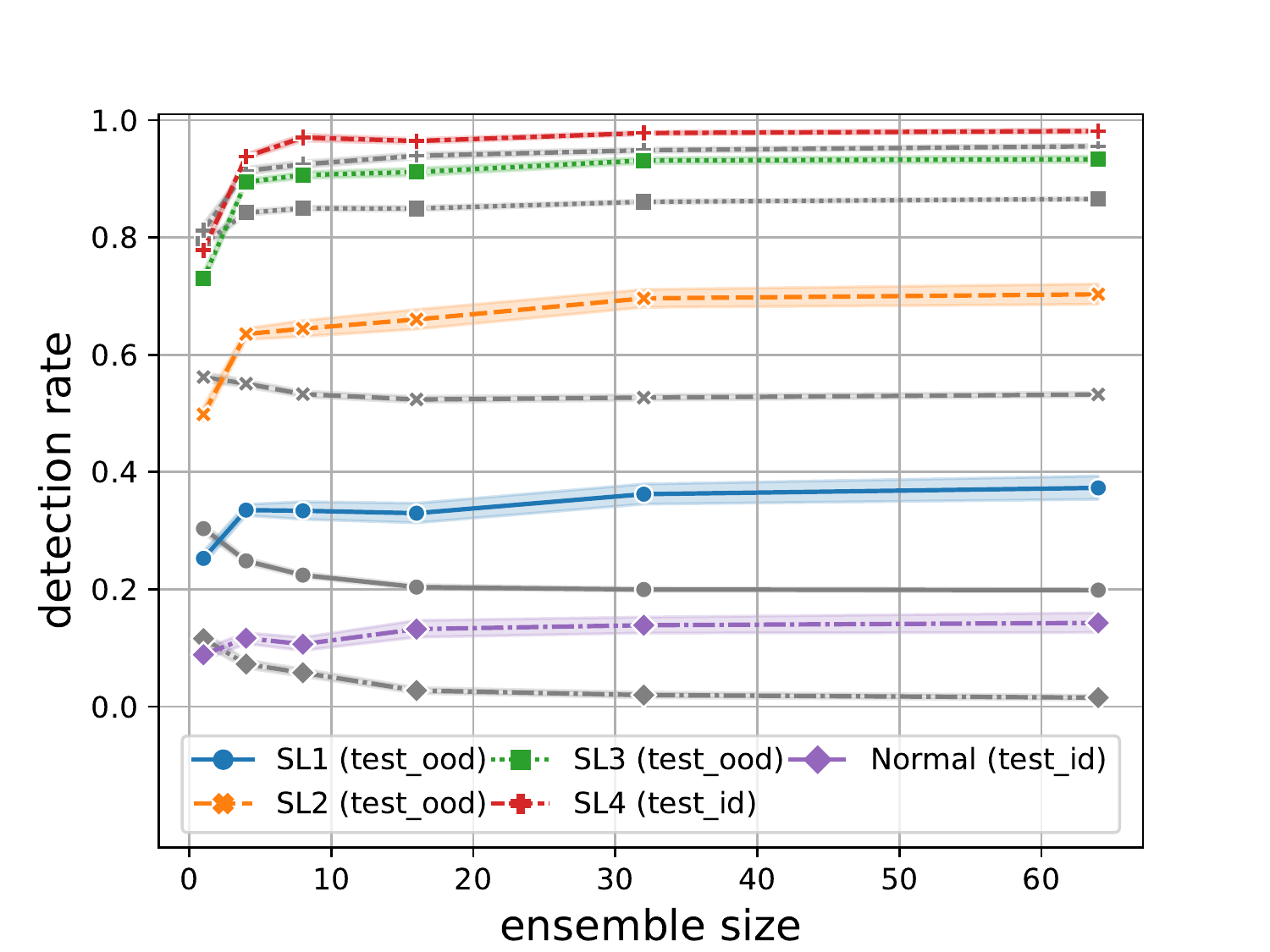}
    \caption{\ac{NN}: \textsc{mean-std}}
    \label{fig:wine_NN_detect_meanvar}
  \end{subfigure}
  
  \begin{subfigure}[t]{0.49\linewidth}
    \centering
    \includegraphics[height=3.3cm]{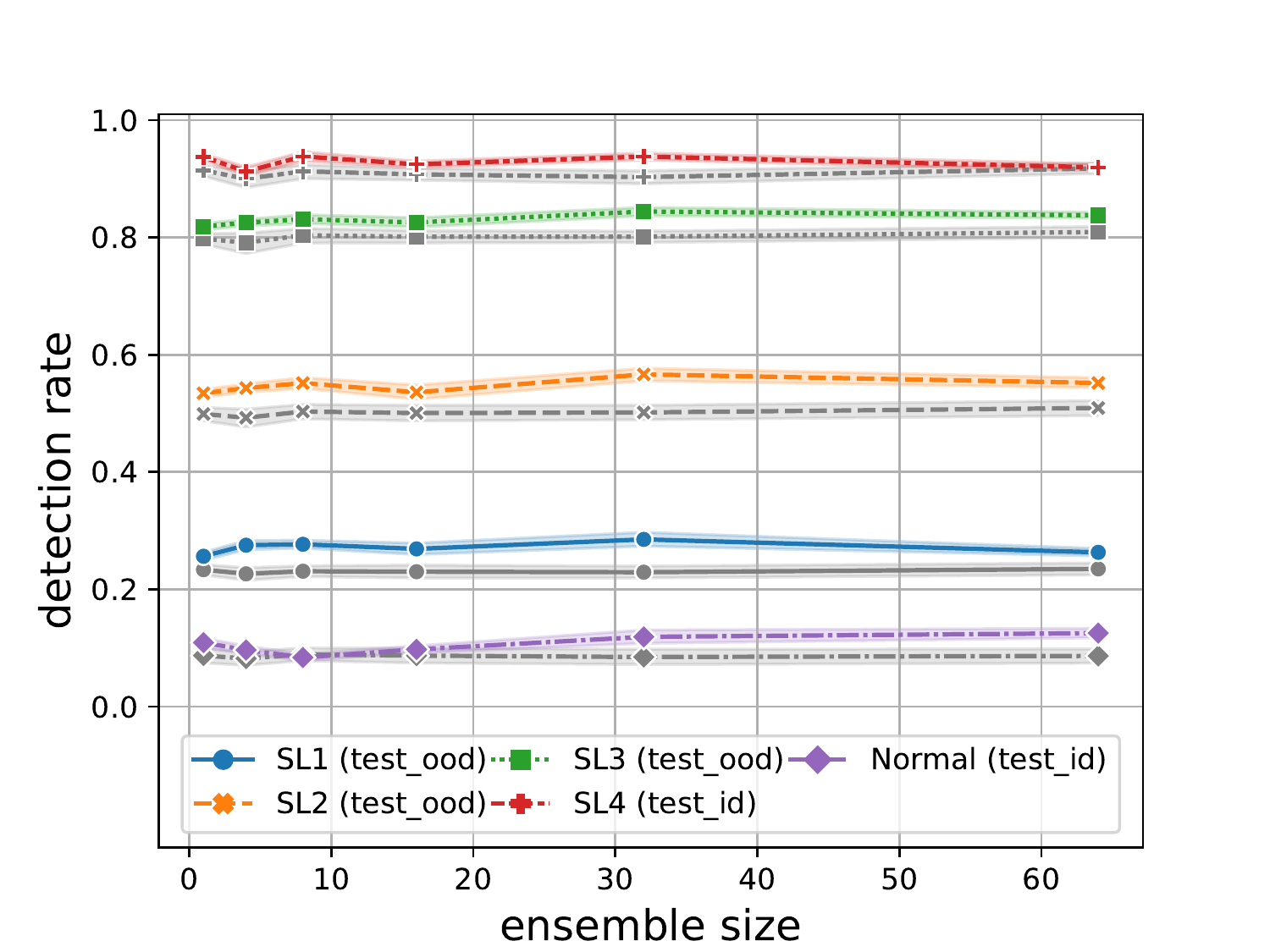}
    \caption{Dropout \ac{NN}: \textsc{mean}}
    \label{fig:wine_DPNN_detect_mean}
  \end{subfigure}  
  \begin{subfigure}[t]{0.49\linewidth}
    \centering
    \includegraphics[height=3.3cm]{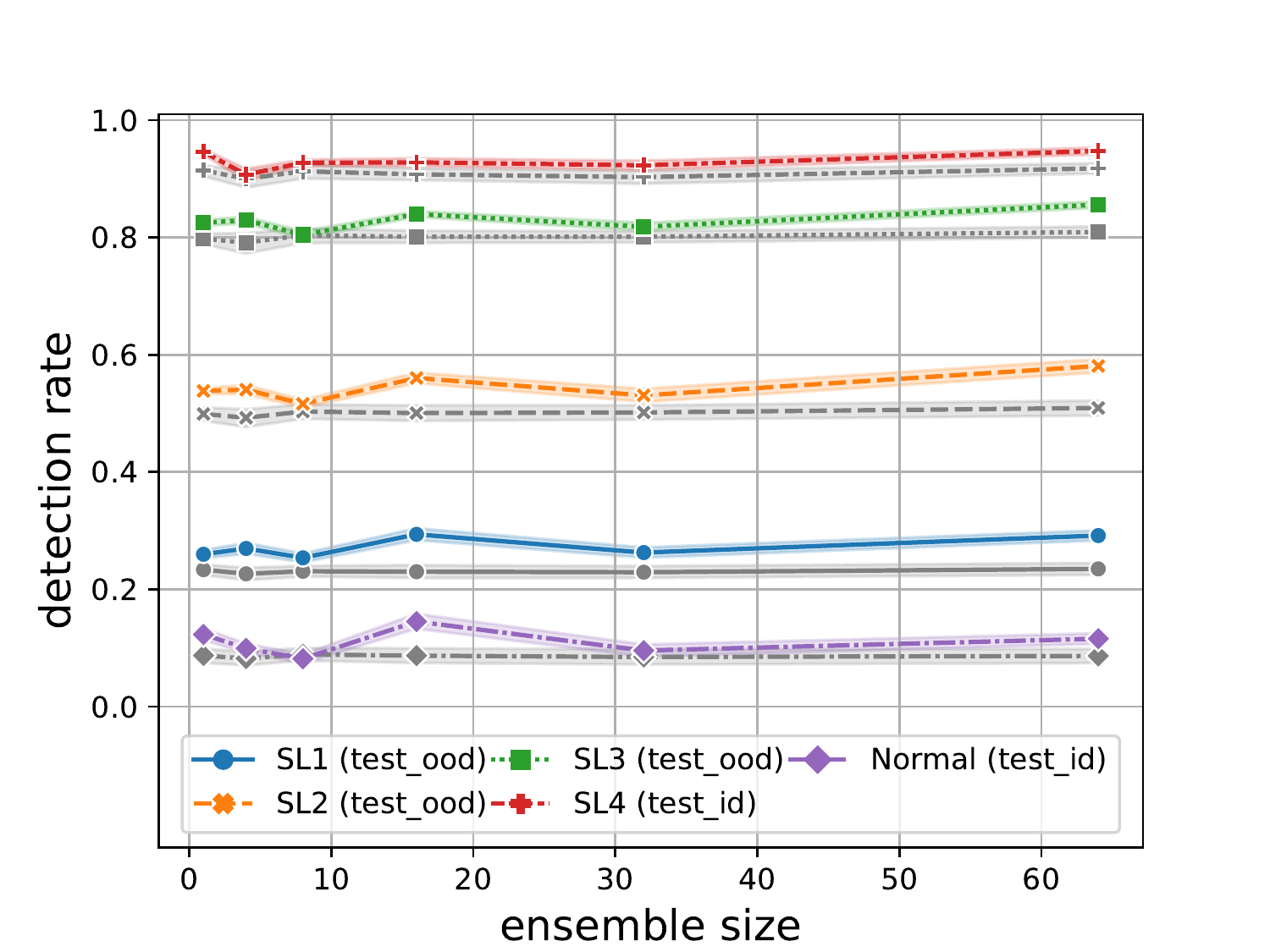}
    \caption{Dropout \ac{NN}: \textsc{mean-std}}
    \label{fig:wine_DPNN_detect_meanvar}
  \end{subfigure}
  
  \caption{Detection performances of ensemble models on the wine dataset. \textsc{cfar} (colored) and \textsc{argmax} (gray) are compared.}
  \label{fig:detection-performance-wine}
\end{figure} 

\subsection{Bagging Ensembles}

In our empirical study, we evaluated bagging ensembles made up of \ac{DT} or \ac{NN} base learners. Four different combinations of aggregation functions (\textsc{mean} \& \textsc{mean-std}) and detection rules (\textsc{argmax} \& \textsc{cfar}) were examined. For each model, an extensive grid search was performed over different combinations of hyperparameters, such as ensemble size and training epochs. The details of our experimental setup can be find in Section~\ref{sec:app-experimental-setup} in the appendix. 

The detection performances of selected classifiers (with \acp{FPR} under the set thresholds) are plotted in Figure~\ref{fig:detection-performance-chiller} for the chiller dataset, and in Figure~\ref{fig:detection-performance-wine} for the wine dataset. The plots show the trends in the detection rate for each data subset for ensembles of different sizes. For each ensemble size, we again plot the $95\%$ \ac{CI} of the performance indices of the top $K$ models. Below are the findings and our analyses.

\subsubsection{Improved detection performance from \textsc{cfar}}

From Figures~\ref{fig:detection-performance-chiller}\,\&\,\ref{fig:detection-performance-wine}, we can see that \textsc{cfar} gives improved fault detection performance in almost all scenarios for bagging ensembles (\ac{RF} \& \ac{NN}), with little impact on the \ac{FPR} on normal data. For \ac{RF} models, we can see that the detection rates on SL4 faults converge to $100\%$ under \textsc{cfar} than under \textsc{argmax}. In other words, with the \textsc{cfar} rule smaller-sized \acp{RF} can achieve better detection performance than with \textsc{argmax} on in-distribution fault data. Similar phenomena do not appear for \ac{NN} ensembles because single-learner \ac{NN} models already achieve almost $100\%$ detection rate under either \textsc{argmax} or \textsc{cfar}, probably due to the fact that \acp{NN} are much stronger learners than individual \acp{DT}.
On \ac{OOD} (SL1--3) faults, we can see that \textsc{cfar} also gives improved performance compared to \textsc{argmax}, especially in the \ac{RF} case. The performance gap between \textsc{cfar} and \textsc{argmax}  becomes increasingly evident with large ensembles: about 15\% on the chiller data and about 35\% on the wine data for \ac{RF} models. The gaps widen fast at the beginning and then saturate at ensemble sizes of $32$ and $64$.

\begin{figure}[tb]
    \centering
    \includegraphics[width=0.9\linewidth]{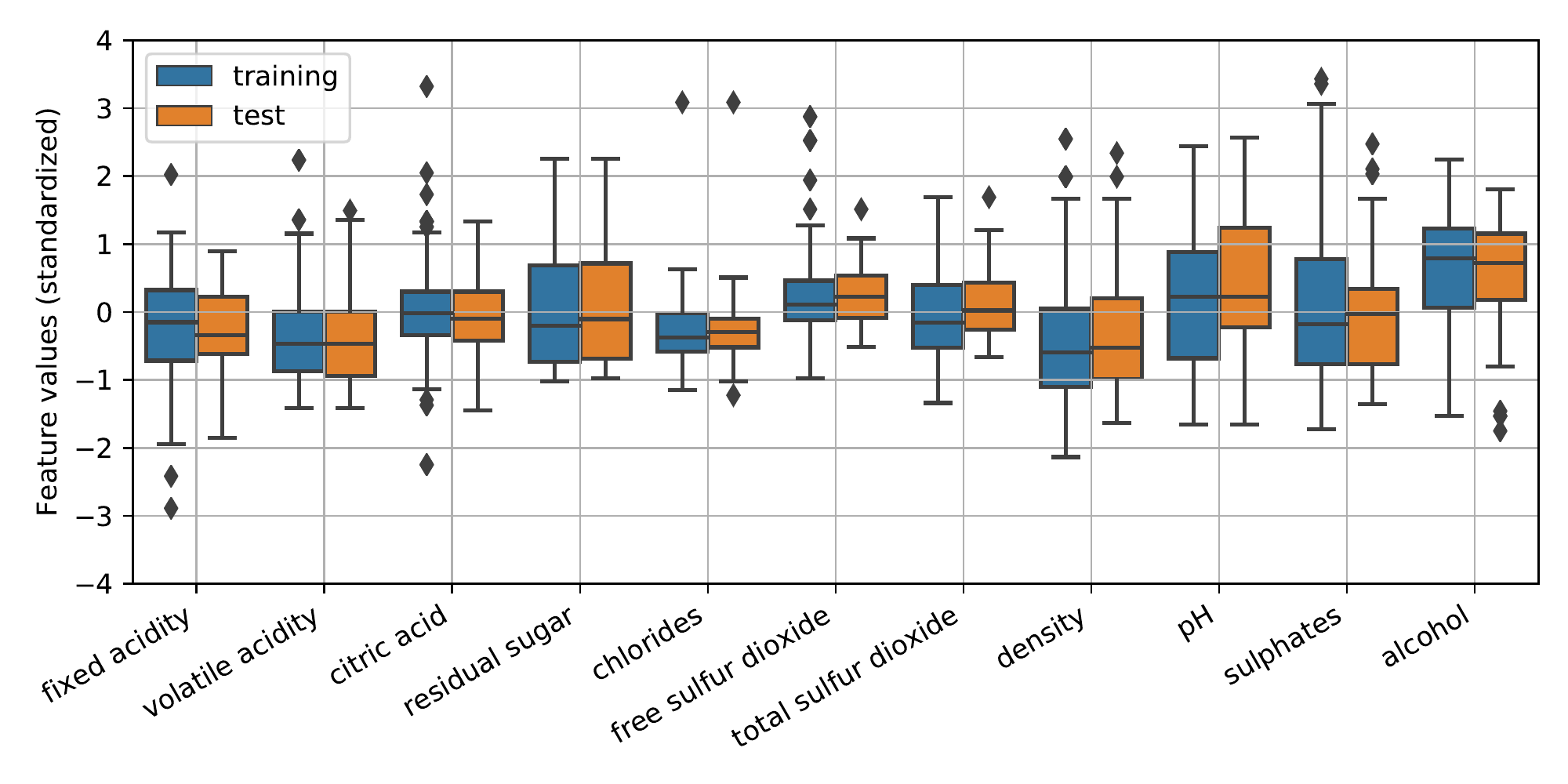}
    \vspace{-3mm}
    \caption{Box plot for visualizing the distribution of each feature in the SL0 wine data.}
    \label{fig:wine-SL0}
\end{figure}

It is also interesting that bagging ensembles lead to slightly increased detection rates on the test set SL0 data from the wine dataset; see Figure~\ref{fig:detection-performance-wine} for details. We speculate the ``false positives''  are because the training and the test SL0 data may follow different distributions. One possible cause is the small sample size (only 141 SL0 instances) of the training data, which may not be enough to represent the true distribution of SL0 data. A boxplot is shown in Figure~\ref{fig:wine-SL0} to show the distribution of each feature in the SL0 wine data. As can be seen from the plot, the training and the test SL0 data distributions do not fully overlap, indicating some potential distribution shift between the two.

\subsubsection{Degraded detection performance from \textsc{argmax} with ensembles}

\begin{figure*}[tb]
    \centering
    \begin{subfigure}[t]{0.32\linewidth}
        \centering
        \includegraphics[height=3.5cm]{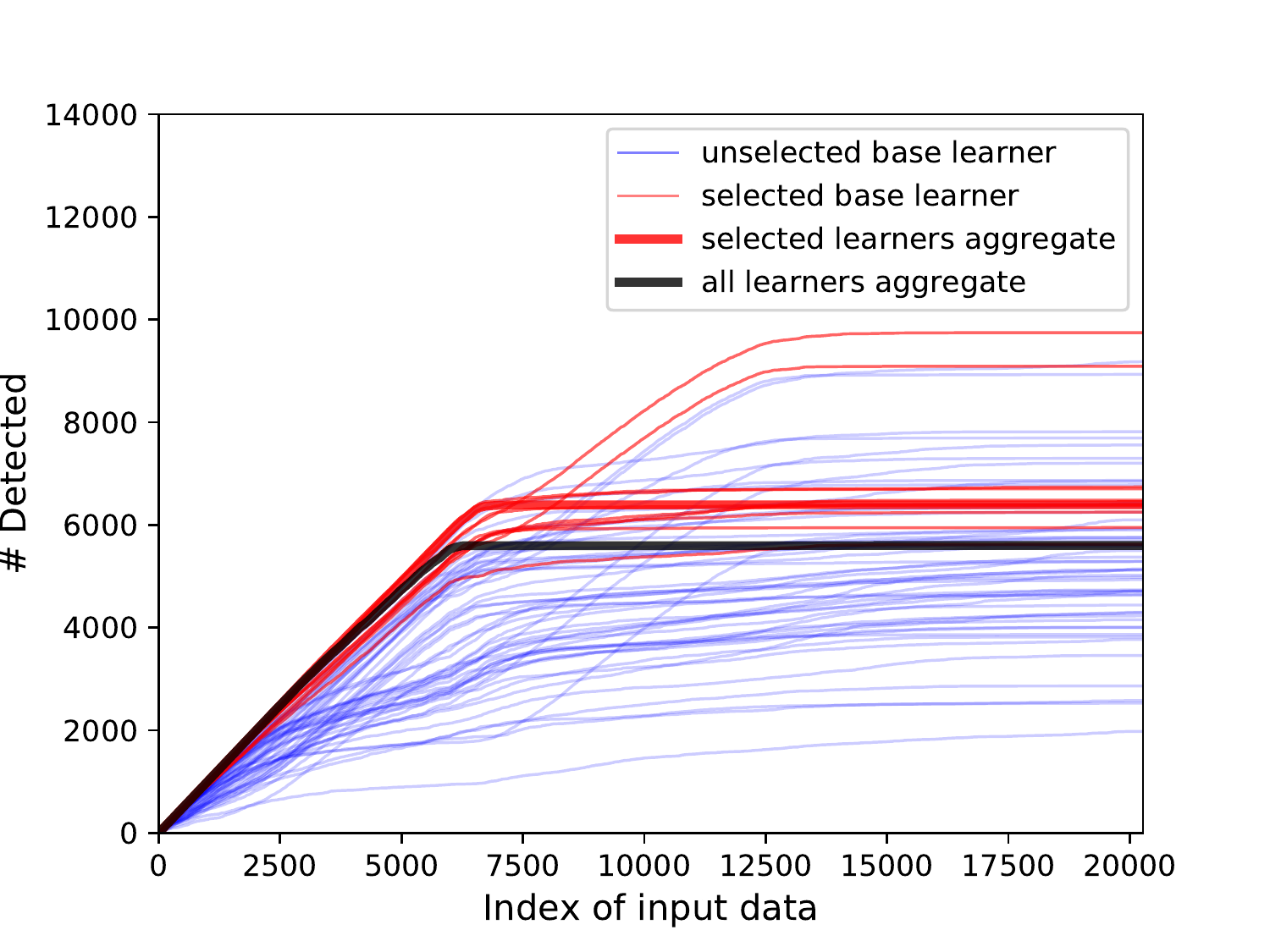}
        \caption{\ac{RF}}
        \label{fig:zoomin-RF}
    \end{subfigure}
    \begin{subfigure}[t]{0.32\linewidth}
        \centering
        \includegraphics[height=3.5cm]{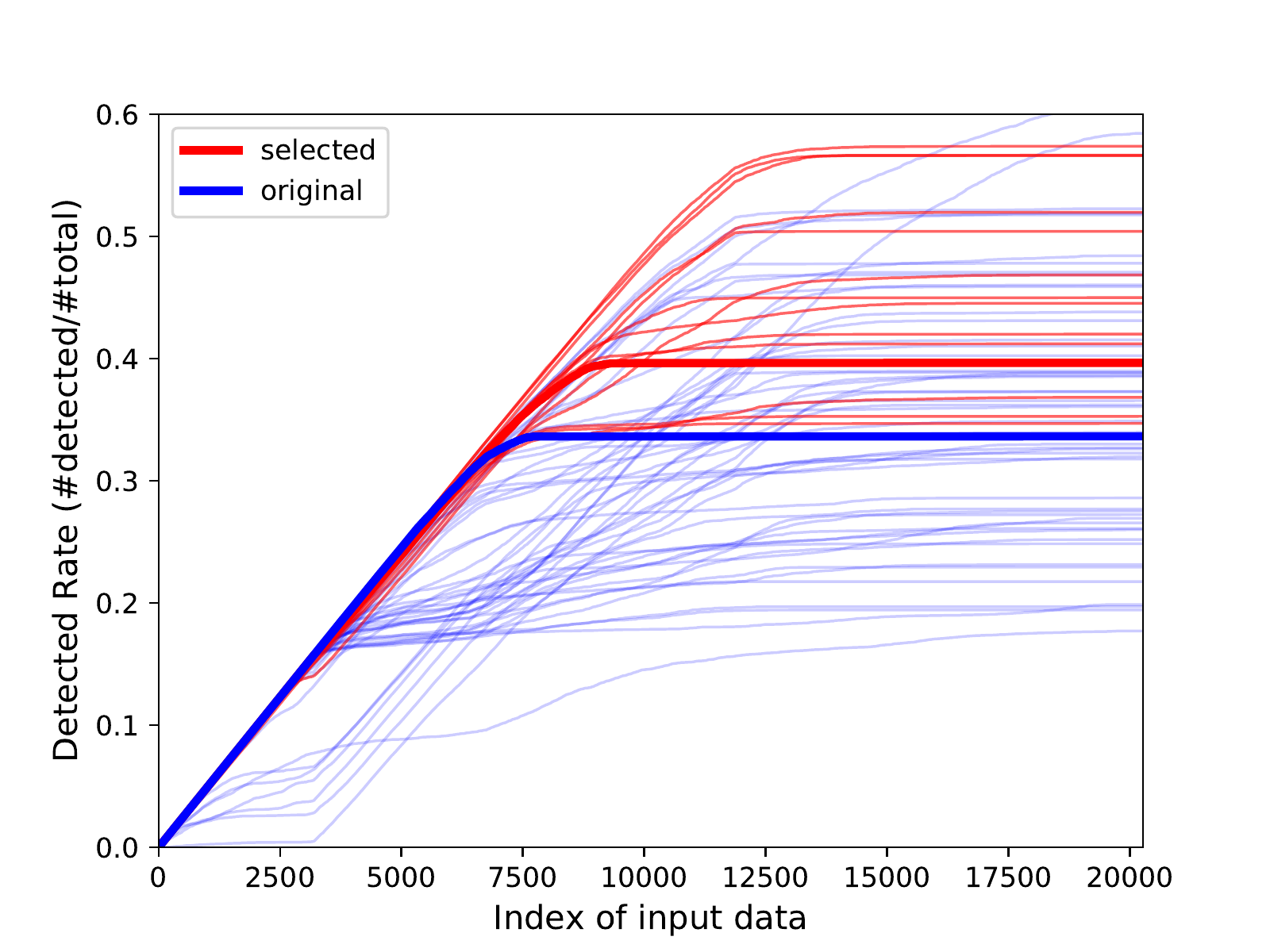}
        \caption{\ac{NN}}
        \label{fig:zoomin-NN}
    \end{subfigure}
    \begin{subfigure}[t]{0.32\linewidth}
        \centering
        \includegraphics[height=3.5cm]{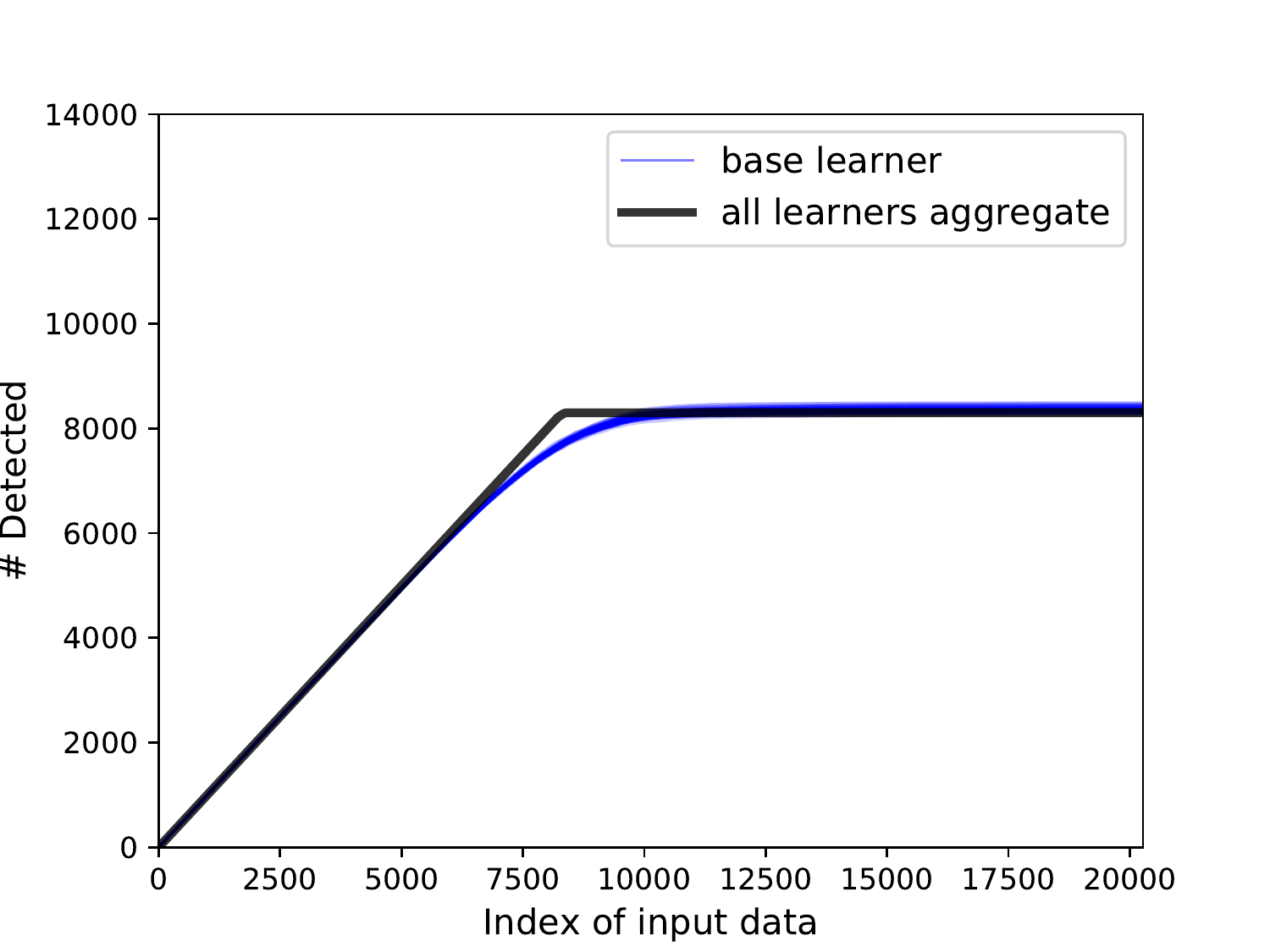}
        \caption{Dropout \ac{NN}}
        \label{fig:zoomin-DPNN}
    \end{subfigure}
\vspace{-3mm}
\caption{Cumulative number of detected SL1 faults. }
\end{figure*}

One surprising phenomenon is that under the \textsc{argmax} detection rule, the detection performance on low-severity faults even assumes a declining trend with increasing ensemble sizes. In comparison, with the \textsc{cfar} rule we can see some slight increase in the detection performance on SL1 faults. To understand the causes of declining performance on SL1 faults under \textsc{argmax}, we pick two examples, an \ac{RF} model and an \ac{NN} model, to take a deeper look. Both models consist of 64 base learners and are trained on the chiller data. $90\%$ of the 16 original features (i.e. 14 features) and $80\%$ of the original 20272 training examples are randomly chosen for training each base learner in the bagging ensembles. To visualize the performances of each individual base learner in detecting SL1 faults, we plot Figures~\ref{fig:zoomin-RF}\,\&\,\ref{fig:zoomin-NN} where the horizontal axes show the indices of SL1 examples and the vertical axes show the ``cumulative'' number of detected faults. A point $(i,j)$ on the plot means that a total of $j$ data points among $\bm{x}_0,\bm{x}_1,\ldots,\bm{x}_i$ will be identified as faults by a model. The SL1 fault examples $\bm{x}_i$'s are sorted in ascending order by their $\mu_{i,0}$, such that data points that are more likely to be recognized as normal are sorted towards the right. In other words, the faults on the right are presumably more difficult to detect. 

In Figures~\ref{fig:zoomin-RF}\,\&\,\ref{fig:zoomin-NN}, the thin lines (red and blue) show the detection performance of individual base learners, where red lines signify base learners that achieve below $2\%$\footnote{The $2\%$ cutoff threshold for individual learners is chosen to be less tight than the $1\%$ threshold we use for selecting ensemble models, because in ensemble learning we can allow individual learners to be weaker and still get satisfactory performance when the individual learners are aggregated.} \ac{FPR} and \ac{FNR} on the training distribution. As can be seen, there is large variation in their performances.
The dark thick line shows the performance of an ensemble model made up of all 64 base learners (aggregated using \textsc{mean}), and the red thick line corresponds to an ensemble consisting of only the learners in red. As can be seen, all red lines achieve a better performance than the dark one, meaning that a better performance on \ac{IS} faults could have been achieved by ensemble models had cross-validation been applied to exclude poorly-performing base models.

\subsubsection{Small performance differences between \textsc{mean-std} and \textsc{mean}}
By comparing Figures~\ref{fig:detection-performance-chiller}\,\&\,\ref{fig:detection-performance-wine}, it is interesting to see that the differences between \textsc{mean-std} and \textsc{mean} are rather minimal for detection tasks.
One possible explanation is that the disagreement among ensemble members has already been captured by the \textsc{mean} statistic, resulting in minimal gain from applying \textsc{mean-std} for the detection tasks. However, as we are going to show shortly, the choice between \textsc{mean-std} and \textsc{mean} does yield a difference in diagnosis performance.

\subsubsection{Diagnosis performance}

In Figure~\ref{fig:detection-performance-chiller}, we visualize both the top-1 and the top-2 accuracy results from the \ac{RF} and ensemble \ac{NN} models. As expected, the top-2 accuracy values are much higher than their top-1 counterparts. The same models as we have selected to evaluate the detection performances in Figure~\ref{fig:detection-performance-chiller} are used to evaluate the diagnosis performances. Again, we plot the $95\%$ \ac{CI} associated with each performance index. 
For bagging ensembles (\acp{RF} and \acp{NN}), we can see obvious improvement in both the top-1 and the top-2 accuracy by using \textsc{mean-std} (colored lines) compared to \textsc{mean} (gray lines), especially with large ensembles. The gain is more evident with \ac{RF} models than with \ac{NN} models. For \ac{RF}, the top-2 diagnosis accuracy on SL1 faults reaches almost $60\%$ with size-64 ensembles.

\subsection{Dropout Ensembles}

Similar to our experiments on regular \ac{NN} models, we use grid search to evaluate MC-dropout models under different hyperparameter settings (e.g.,~dropout rates and batch sizes) and network structures. Additional details about our experimental setup can be found in Section~\ref{sec:dropout-ensembles-setup}.

Similar to what we discovered with bagging ensembles, with dropout ensembles we can see some small performance gains from using \textsc{mean-std} instead of \textsc{mean}. In addition, we can see that the detection performances of dropout ensembles are not sensitive to the ensemble size, unlike bagging ensembles.
To understand the reasons behind that, we pick an MC-dropout model with dropout rate of $0.15$ and plot the individual detection performance as we did with bagging ensembles, as shown in Figure~\ref{fig:zoomin-DPNN}. It can be seen that there is very little variation in the prediction performances from each individual member (shown as thin blue lines), indicating a lack of diversity among the ensemble members. The performance of the aggregate model (shown as dark line) is close to those of ensemble members.

In terms of diagnosis performance, the differences between \textsc{mean} and \textsc{mean-std} on the chiller dataset for dropout \acp{NN} are rather small; see Figures~\ref{fig:DPNN-top1-diagnosis}\,\&\,\ref{fig:DPNN-top2-diagnosis} for a comparison. The gains in top-1 and top-2 accuracy from using larger ensembles are on the level of 5 to 10 percentage points.

To summarize, dropout ensembles deliver slightly inferior detection and diagnosis performances to \ac{RF} and \ac{NN} ensembles; however, since they require less effort to train, they can serve as cheaper alternatives to bagging ensembles.

\subsection{Comparison with One-Class Models}

\begin{figure}[tb]
  \centering
  
  \begin{subfigure}[t]{0.49\linewidth}
    \centering
    \includegraphics[height=3.3cm]{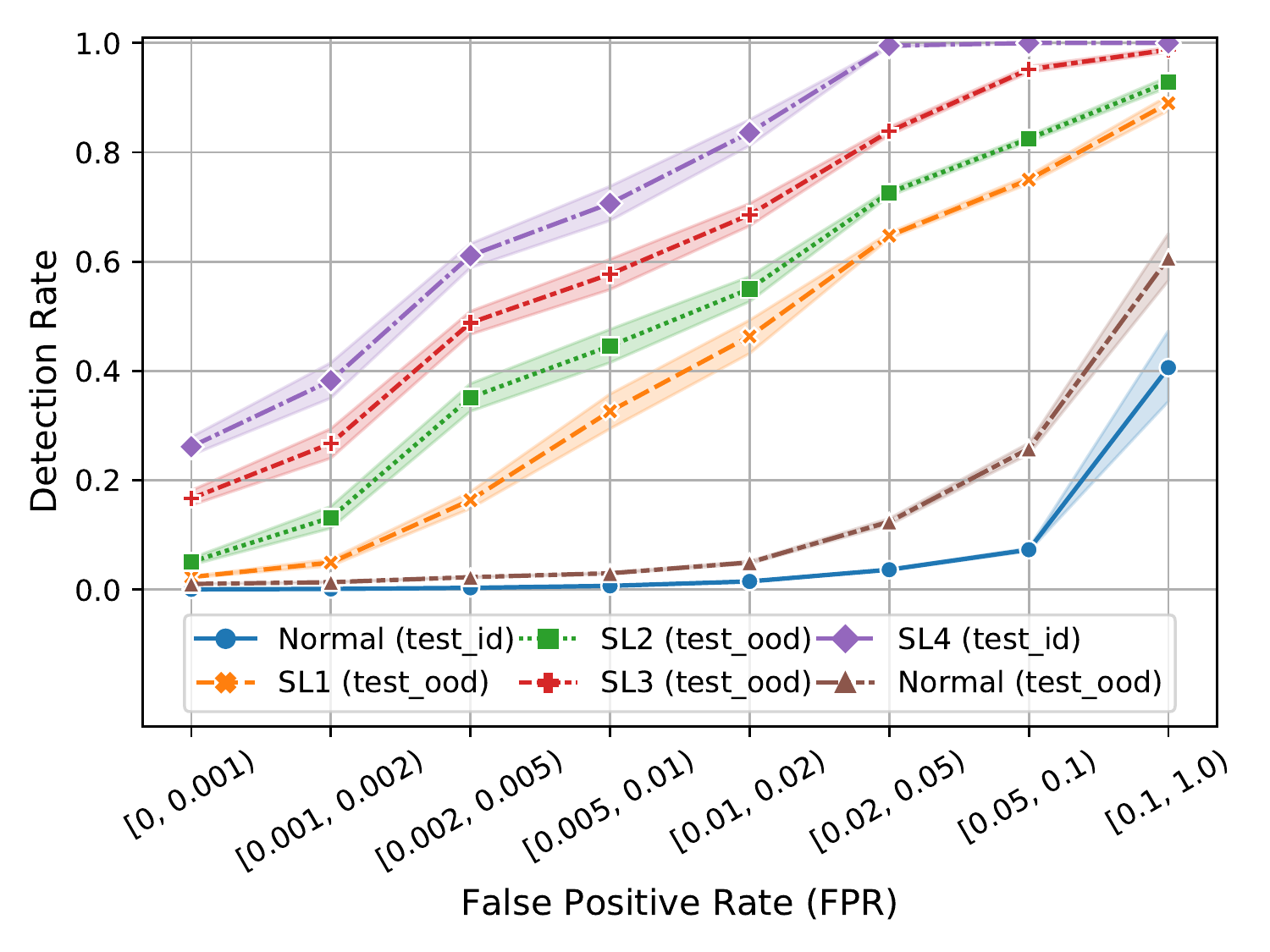}
    \caption{{OC-SVM}: chiller}
    \label{fig:OCSVM-chiller-fpr-sweep}
  \end{subfigure}
  \begin{subfigure}[t]{0.49\linewidth}
    \centering
    \includegraphics[height=3.3cm]{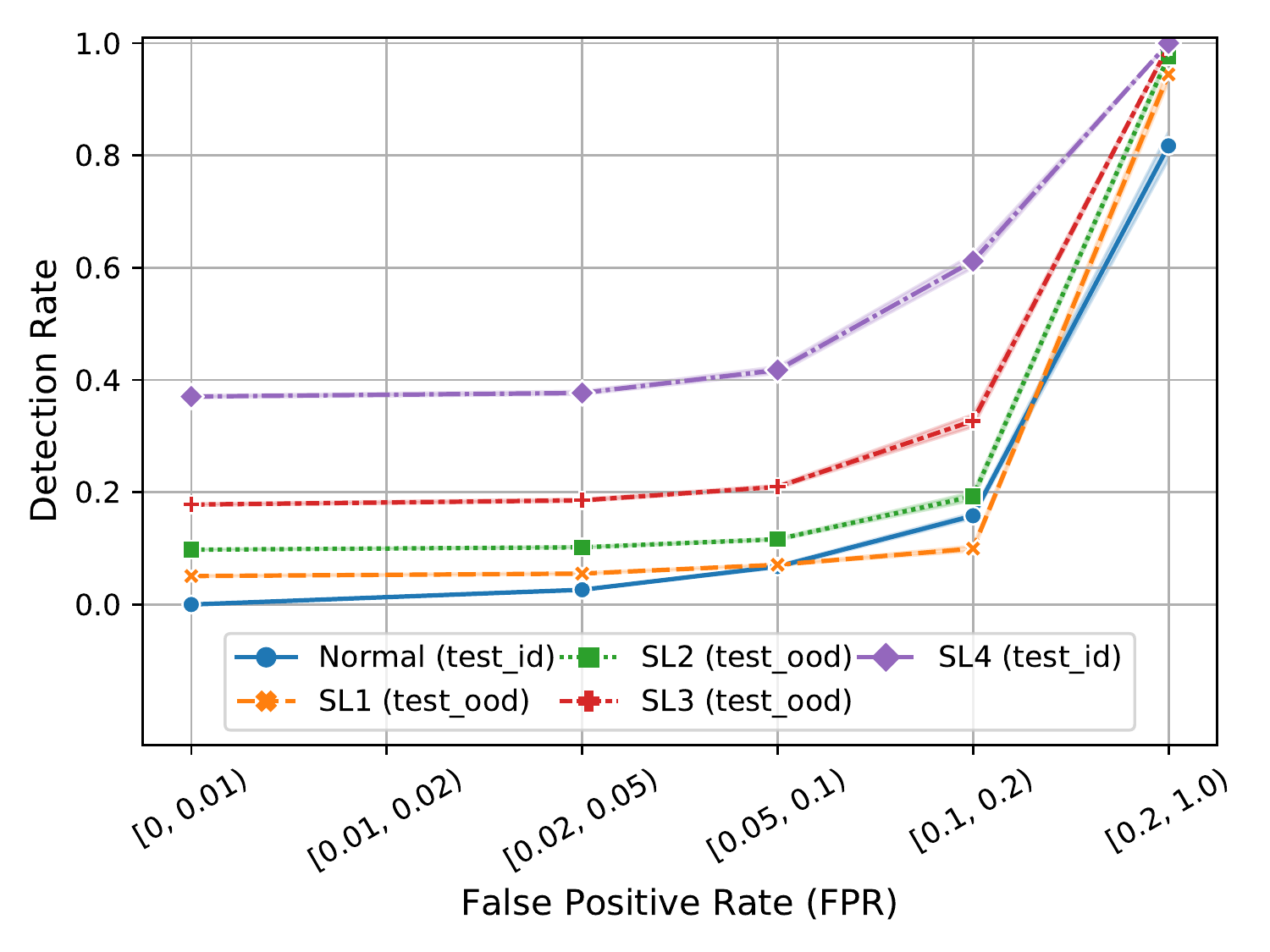}
    \caption{{OC-SVM}: wine}
    \label{fig:OCSVM-wine-fpr-sweep}
  \end{subfigure}
 
  \begin{subfigure}[t]{0.49\linewidth}
    \centering
    \includegraphics[height=3.3cm]{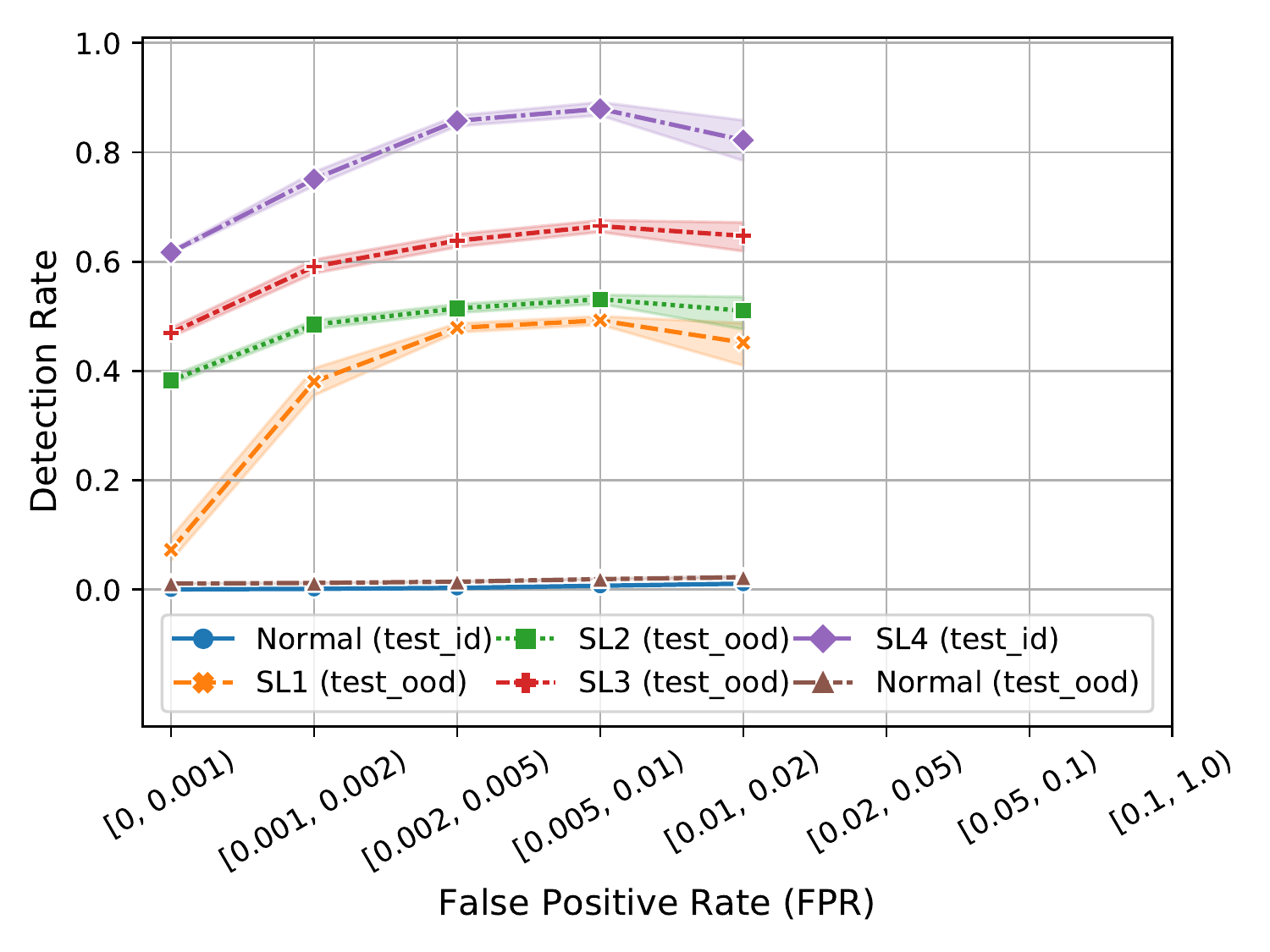}
    \caption{Autoencoder: chiller}
    \label{fig:AE-chiller-fpr-sweep}
  \end{subfigure}
  \begin{subfigure}[t]{0.49\linewidth}
    \centering
    \includegraphics[height=3.3cm]{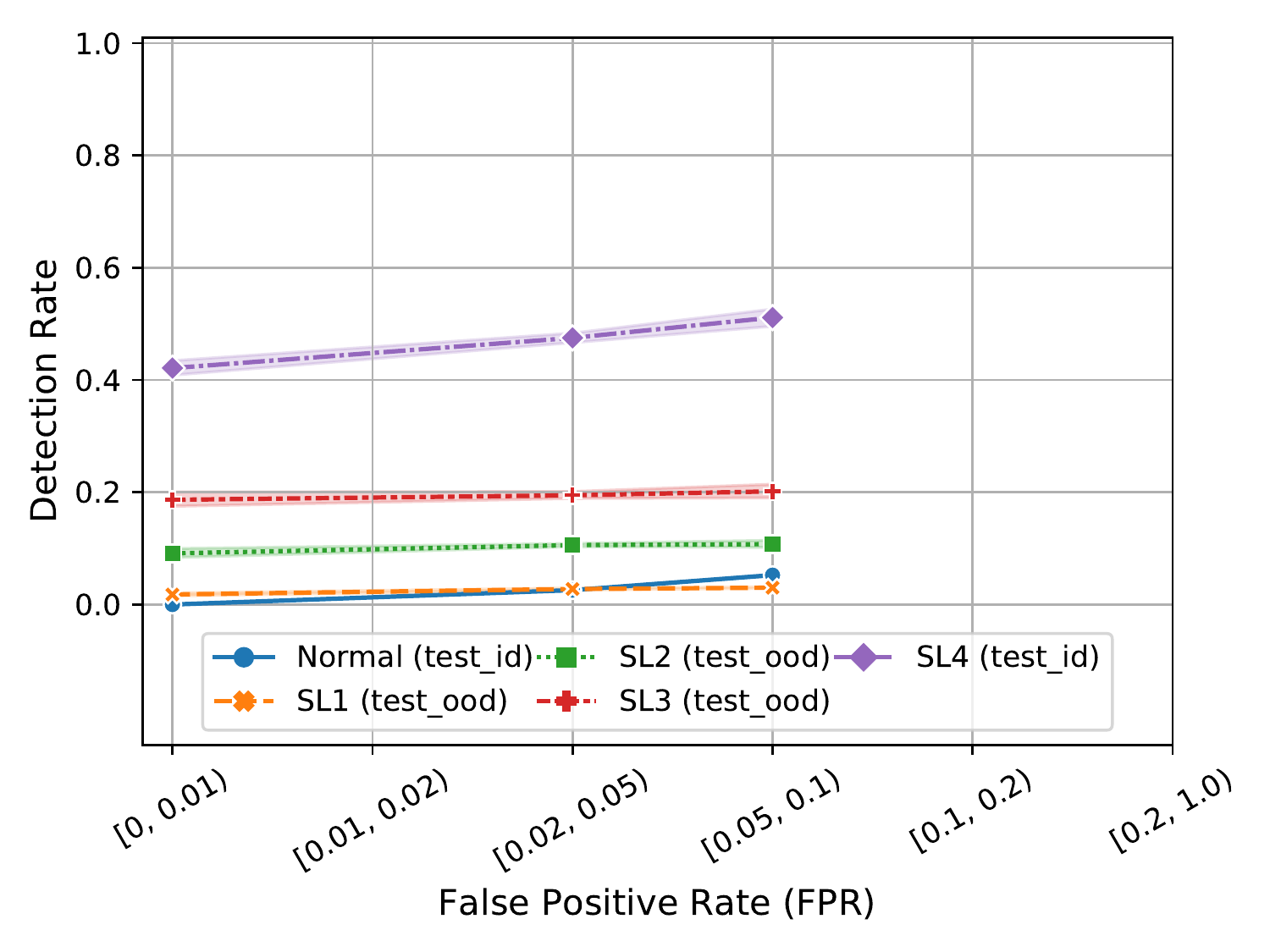}
    \caption{Autoencoder: wine}
    \label{fig:AE-wine-fpr-sweep}
  \end{subfigure}
  
  \caption{Detection performances of one-class classifiers.}
  \label{fig:FPR-sweep-OC}
\end{figure}

One-class models such as \acp{OC-SVM}~\cite{scholkopf2001estimating} and autoencoders~\cite{sakurada2014anomaly} are classical models for semi-supervised anomaly detection and can be used to detect \ac{IS} faults. They only use the negative class (normal) data to learn a decision boundary that encompasses the negative class data.

It will not be fair to directly compare the performances of one-class models and supervised models, because the SL4 data are not utilized in training one-class models. To evaluate the two families of models on a fair basis, we use SL4 data for selecting one-class models after training, which will help us find good hyperparameter settings for one-class models that achieve high sensitivity and specificity.

The performances of one-class models are plotted in Figure~\ref{fig:FPR-sweep-OC} in a similar fashion as in Figure~\ref{fig:FPR-sweep}. Again, only the performance indices of the top $K$ models in each \ac{FPR} interval are shown on the plot. 
By comparing the plots in Figures~\ref{fig:FPR-sweep}\,\&\,\ref{fig:FPR-sweep-OC}, we can immediately see that all three supervised methods have superior performance to semi-supervised models; the selected models give low performance even on SL4 data. Considering the fact that top models are chosen based on their detection rates on SL4 in the model selection process, this is a surprising result.

Now let us compare the detection performances on \ac{IS} faults. For the chiller dataset, we use performance indices under the $[0.005, 0.01)$ \ac{FPR} bin for comparing all model classes, because the one-class models with lower \ac{FPR} give much lower detection rates. By comparing Figures~\ref{fig:OCSVM-chiller-fpr-sweep}\,\&\,\ref{fig:AE-chiller-fpr-sweep}, we can see that autoencoders have slightly better performance than \acp{OC-SVM}. Comparing to the supervised models, autoencoder gives a similar performance (around $50\%$) in detecting SL1 faults; however, its detection performance on SL2--SL4 faults is much behind supervised models on the chiller dataset. For the wine dataset, we can see from Figures~\ref{fig:OCSVM-wine-fpr-sweep}\,\&\,\ref{fig:AE-wine-fpr-sweep} that both one-class models give similar performances. In addition, the detection performances on SL1--SL4 faults do not vary much for models with below $0.1$ \ac{FPR}. The performances given by one-class models are much worse than those from the supervised models; the gaps in detection rates are at least $0.2$ across all \acp{SL}. The results show that supervised information from severe (SL4) faults is very helpful for detecting \ac{IS} faults.




\section{Related Work}\label{sec:related-works}

In recent years, several papers have been published related to the detection of \ac{OOD} data, especially in the deep learning community. While it is obviously impossible for us to give a complete account of related research in this paper, we  give a brief overview of the most relevant papers and highlight their connections to our work.

\paragraph{Model uncertainty} 

There have been a strong and growing interest in estimating the \textit{prediction uncertainty}~\cite{lakshminarayanan2017simple,gal2016uncertainty} of supervised models, especially in the deep learning community. Two primary main motivations of this area of research are 1) to understand when to \textit{trust} a model's predictions, and 2) to design defense mechanisms against adversarial examples. 
One application of uncertainty estimation is to detect \ac{OOD} examples with supervised deep learning models. Deep ensembles~\cite{lakshminarayanan2017simple} and MC-dropout~\cite{gal2016uncertainty,jin2019detecting} are two popular approaches in academia, and interestingly they both explicitly or implicitly employ the ensemble idea.
Although promising results from these deep ensemble approaches have been demonstrated on certain types of \ac{OOD} data such as dataset shift and unseen/unknown classes~\cite{lakshminarayanan2017simple}, it is difficult to evaluate their effectiveness in general, because the \ac{OOD} part of the world is obviously much ``larger'' than its in-distribution counterpart and is presumably much harder to analyze. In contrast, our work still embraces a closed-world assumption, and restricts the focus to \ac{IS} faults---a special subset of \ac{OOD} data that has a close connection to the in-distribution data. We speculate that some knowledge necessary for detecting \ac{IS} faults is already entailed in the training data, thus making the detection of \ac{IS} faults possible with supervised methods. 

\paragraph{Model calibration}

Another line of work aims to produce good probability outputs using \textit{model calibration} techniques~\cite{niculescu2005predicting, guo2017calibration}, e.g. temperature scaling~\cite{guo2017calibration}. The calibration techniques are typically applied in a post-processing manner; in other words, a calibration method learns a transformation that is applied to a model's uncalibrated output probabilities, without affecting the parameters (weights) of the original model itself. Although good confidence measures are important in many other fields, we are skeptical about the role of model calibration in the context of anomaly detection, either with the more commonly used \textsc{argmax} method or our proposed \textsc{cfar} method. By design, calibration methods should only adjust probabilities, but not class predictions. Therefore, under \textsc{argmax} the predicted class of input $x_i$, as well as the anomaly detection decisions, will not change. Under \textsc{cfar}, as previously mentioned, it is the \textit{ranking} of a test data point's score among those of the normal training data points that determines the detection decision. If the calibration method results in an isotonic transformation (as are popular methods), the rankings will not change. Nor will the detection decisions.

\section{Conclusions and Future Work}\label{sec:conclusion}

\acf{IS} examples can pose critical risks to \acf{FDD} systems built upon \acf{ML} techniques, especially under situations where these examples are lacking in the training data. The resulting classifier trained without \ac{IS} examples can easily mistake them for normal ones, which can lead to costly consequences because the best time for treatment is missed.

To tackle this problem, we study how to use ensemble methods to improve the detection and diagnosis performance on \acf{OOD} \ac{IS} anomaly instances. In fact, ensemble methods have been considered popular candidates for detecting \ac{OOD} examples~\cite{lakshminarayanan2017simple}; however, there are caveats when applying them to \ac{FDD}. In this paper, we focused on a special type of \ac{OOD} data---\ac{IS} faults, and find that ensemble methods are effective in detecting and diagnosing \ac{IS} faults only when proper aggregation functions and detection rules are used for interpreting the probability values generated by ensemble classification models. 

In the future, we plan to extend the proposed ensemble-based \ac{FDD} approach to high-dimensional data (e.g.,~thermal infrared temperature measurement data of 3D printing process~\cite{jin2019encoder}).

\begin{acks}
This work is supported in part by the National Research Foundation of Singapore through a grant to the Berkeley Education Alliance for Research in Singapore (BEARS) for the Singapore-Berkeley Building Efficiency and Sustainability in the Tropics (SinBerBEST) program, and by the National Science Foundation under Grant No.~1645964. BEARS has been established by the University of California, Berkeley as a center for intellectual excellence in research and education in Singapore. We appreciate Prof. Jiantao Jiao and Mr. Xiangyu Yue for their valuable suggestions.
\end{acks}

\bibliographystyle{ACM-Reference-Format}
\bibliography{all.bib}


\appendix
\clearpage
\section{Implementation Details}\label{sec:app-experimental-setup}

The code for our empirical studies was implemented in Python~3.6, and will be publicly released on GitHub upon paper acceptance. The code during the review phase can be downloaded via \url{https://tinyurl.com/kdd20-ensemble-fdd}. \texttt{sklearn}~\cite{scikit-learn} and \texttt{Keras}~\cite{chollet2015keras} packages were used for implementing the \ac{ML} models used in our experiments. The hyperparameters we used in the grid search are summarized in Table~\ref{tab:hyperparameters}, the ``---'' symbol is used to indicate inapplicable entries for a certain model class. With some abuse of notation, we used the \textsc{MATLAB} notation to represent linearly spaced vectors in a closed interval, e.g., {0:0.2:1} means $\{0,0.2,0.4,0.6,0.8,1.0\}$.  
In our grid search, all combinations of the hyperparameters are experimented, resulting in a large number of trained models. At the bottom of Table~\ref{tab:hyperparameters}, we also report the total number of models generated for each model class. Models with different statistical power (e.g., controlled by the \textsc{cfar} cutoff parameter $\alpha$) are counted as separate models, even if the parameters (weights) of the models are identical. As a result, the actual number of training runs during our grid search is much smaller.

\begin{table*}[tb]
\centering
\resizebox{\textwidth}{!}{%
\begin{tabular}{ccccccccccc}
\hline
Dataset & \multicolumn{5}{c}{Chiller Dataset} & \multicolumn{5}{c}{Wine Dataset} \\ \hline
Model Class & RF & Regular NN & Dropout NN & Autoencoder & OC-SVM & RF & Regular NN & Dropout NN & Autoencoder & OC-SVM \\
$\alpha$ (\textsc{cfar}) & \multicolumn{4}{c}{$10^{\{-2:0.2:0\}}$} & --- & \multicolumn{4}{c}{$10^{\{-2:0.2:0\}}$} & --- \\
Max Features \% (Bagging) & \multicolumn{3}{c}{\{50, 60, 70, 80, 90, 100\}} & --- & --- & \multicolumn{3}{c}{\{50, 60, 70, 80, 90, 100\}} & --- & --- \\
Max Samples\% (Bagging) & \multicolumn{3}{c}{\{20, 40, 60, 80, 100\}} & --- & --- & \multicolumn{3}{c}{\{20, 40, 60, 80, 100\}} & --- & --- \\
\# Layers (NN) & --- & \{2, 4, 6, 8\} & \{2, 4, 6, 8\} & \{4, 6, 8, 10\} & --- & --- & \{2, 4, 6, 8\} & \{2, 4, 6, 8\} & \{4, 6, 8, 10\} & --- \\
Max Depth (RF) & \{8, 10, 12, 15\} & --- & --- & --- & --- & \{4, 6, 8, 10, 12, 15\} & --- & --- & --- & --- \\
Optimizer (NN) & --- & \multicolumn{3}{c}{\texttt{adam}, \texttt{adagrad}} & --- & --- & \multicolumn{3}{c}{\texttt{adam}, \texttt{adagrad}} & --- \\
Batch Size (NN) & --- & \multicolumn{3}{c}{\{32, 64, 128\}} & --- & --- & \multicolumn{3}{c}{\{32, 64, 128\}} & --- \\
Dropout Rate & --- & --- & \{5, 10, 15, 20, 25\} & --- & --- & --- & --- & \{5, 10, 15, 20, 25\} & --- & --- \\
$\nu$ (OC-SVM) & --- & --- & --- & --- & $10^{\{-2:0.1:2\}}$ & --- & --- & --- & --- & $10^{\{-2:0.1:2\}}$ \\
$\gamma$ (OC-SVM) & --- & --- & --- & --- & $10^{\{-3:0.1:3\}}$ & --- & --- & --- & --- & $10^{\{-3:0.1:3\}}$ \\
$\beta$ (Autoencoder) & --- & --- & --- & \{1, 2\} & --- & --- & --- & --- & \{1, 2\} & --- \\ \hline
\# Models & 21060 & 120960 & 36288 & 704 & 2501 & 21060 & 120960 & 36288 & 704 & 2501 \\ \hline
\end{tabular}%
}
\caption{Hyperparameters used in our experiments.}
\label{tab:hyperparameters}
\end{table*}

\subsection{Bagging Ensembles}
The bagging ensembles in this study are implemented using the \texttt{Bagging} module from the \texttt{sklearn} package~\cite{scikit-learn}, which allows us to create bagging models with different types of base learners. The sizes of random feature subsets and sample subsets for training each base learner can be configured through the \texttt{max\_features} and \texttt{max\_samples} options in the \texttt{BaggingClassifier} class. In our experiments, we sweep \texttt{max\_features} from 50\% to 100\%, and \texttt{max\_samples} from 10\% to 100\%. 

\subsubsection{\acf{RF}}\label{sec:RF-ensembles-setup}

The base learners in \acp{RF} are implemented as instances of the \texttt{DecisionTreeClassifier} class from \texttt{sklearn}, which are then aggregated into \acp{RF} using the \texttt{Bagging} module. To increase the randomness among base learners, we set the \texttt{splitter} option in \texttt{DecisionTreeClassifier} to ``random'' so that the best random split will be chosen at each node. The depths of \acp{DT} range from 8 to 15 for the chiller dataset, and from 4 to 15 for the wine dataset. Only trees with the same \texttt{max\_depth} settings are aggregated into an \ac{RF}. In addition, we experiment \acp{RF} of six different sizes (1, 4, 8, 16, 32, 64). We leave the settings as default for everything else.

\subsubsection{\acf{NN} ensembles}\label{sec:NN-ensembles-setup}

The neural networks are implemented in \texttt{Keras} and then interfaced with \texttt{sklearn} \texttt{Bagging} module through a wrapper class in \texttt{Keras}. Since the design space of \acp{NN} is huge, it is impossible for us to enumerate all possible network topologies; as a result, in our experiments we only explore \acp{NN} that have a uniform width (same number of nodes across all layers except for the last one) but with varying depths; see Table~\ref{tab:hyperparameters} for further details. In addition, different optimizer options (\texttt{adam}~\cite{kingma2014adam} and \texttt{adagrad}~\cite{duchi2011adagrad}) and batch size settings (32, 64, 128) are also explored. Each \ac{NN} is trained for 100 epochs minimizing the cross-entropy loss. As with \acp{RF}, we test six different ensemble sizes (1, 4, 8, 16, 32, 64).  

\subsection{Dropout Ensembles}\label{sec:dropout-ensembles-setup}

The MC-dropout networks are also implemented in Keras using the same topologies as the regular \acp{NN} described above. The only difference is the additional dropout layers inserted between every adjacent fully connected layers. For each dropout \ac{NN} instance, the same dropout rate is used across all dropout layers. Each dropout network is trained for 200 epochs minimizing the cross-entropy loss. In the experiments, we sweep over five different dropout rate settings (0.05, 0.10, 0.15, 0.20, 0.25); see Table~\ref{tab:hyperparameters} for further details.

\subsection{Autoencoders}

\begin{figure}[tb]
    \centering
    \includegraphics[width=0.7\linewidth]{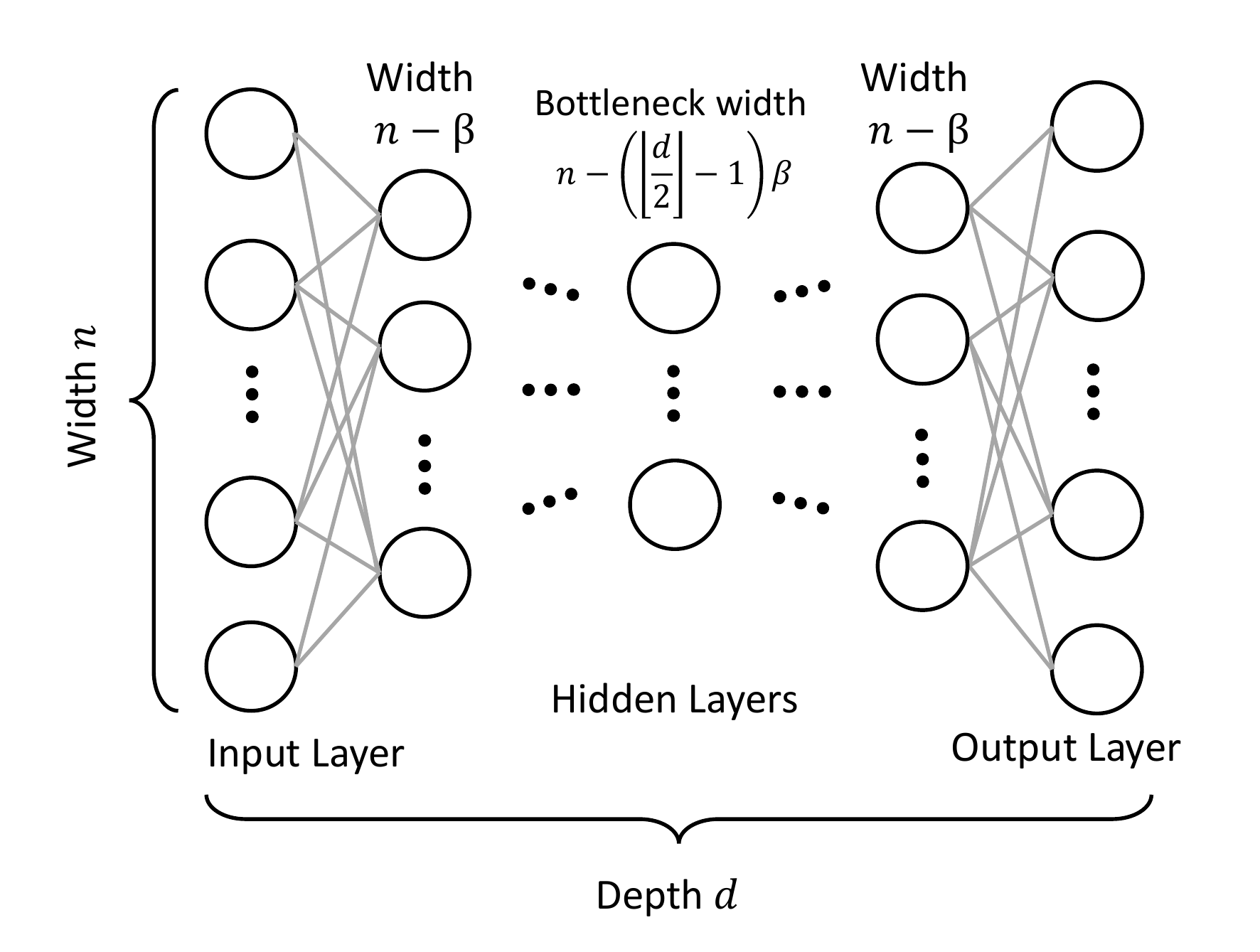}
    \caption{Parameters for autoencoder specifications}
    \vspace{-7mm}
    \label{fig:autoencoder}
\end{figure}

Autoencoders are \ac{NN} models with the same number of input and output nodes and a narrow bottleneck in the middle. In our experiments, we explore a restricted subset of the autoencoder design space, where the network topology can be parameterized by two variables, the network depth $d$ and the ``width decrease rate'' $\beta$. Illustrated in Figure~\ref{fig:autoencoder}, the network has $d$ layers (including the input and the output layers); the number of nodes reduces by $\beta$ by each layer towards the middle. The depth parameter $d$ takes value in $\{4, 6, 8, 10\}$, and $\beta$ takes value in $\{1,2\}$. Suppose the width of the input layer (number of input features) of our model is $n$. 
For a parameter combination to be feasible, we require the width of the bottleneck to be positive, i.e. $n-\left( \left \lfloor\frac{d}{2}\right\rfloor -1 \right) \beta > 0$.
Each autoencoder network is trained for 200 epochs minimizing the $\ell_2$ reconstruction error.

\begin{figure}[tb]
  \centering
  \begin{subfigure}[t]{0.49\linewidth}
    \centering
    \includegraphics[height=3.2cm]{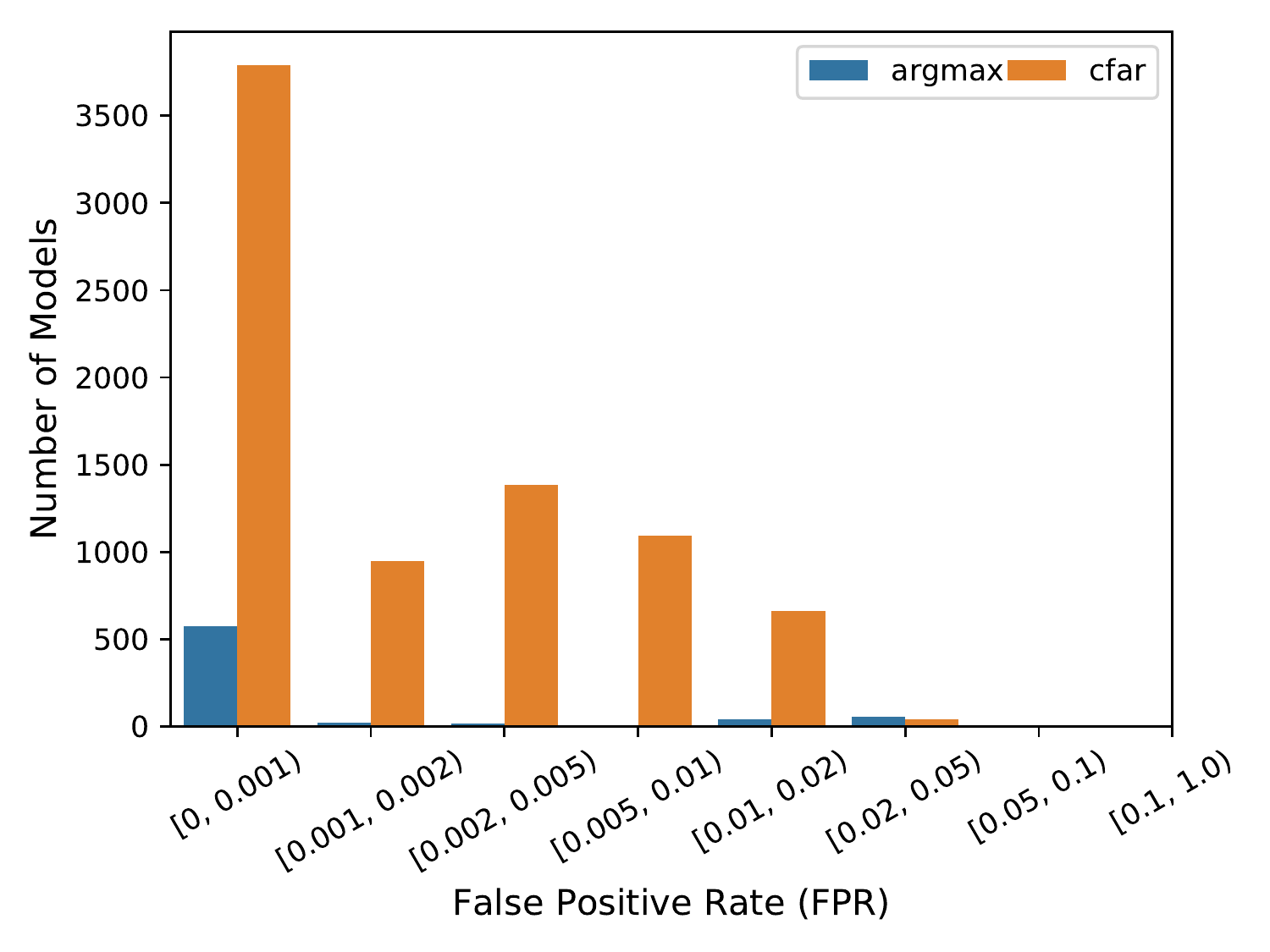}
    \caption{\ac{RF}: chiller}
  \end{subfigure}
  \begin{subfigure}[t]{0.49\linewidth}
    \centering
    \includegraphics[height=3.2cm]{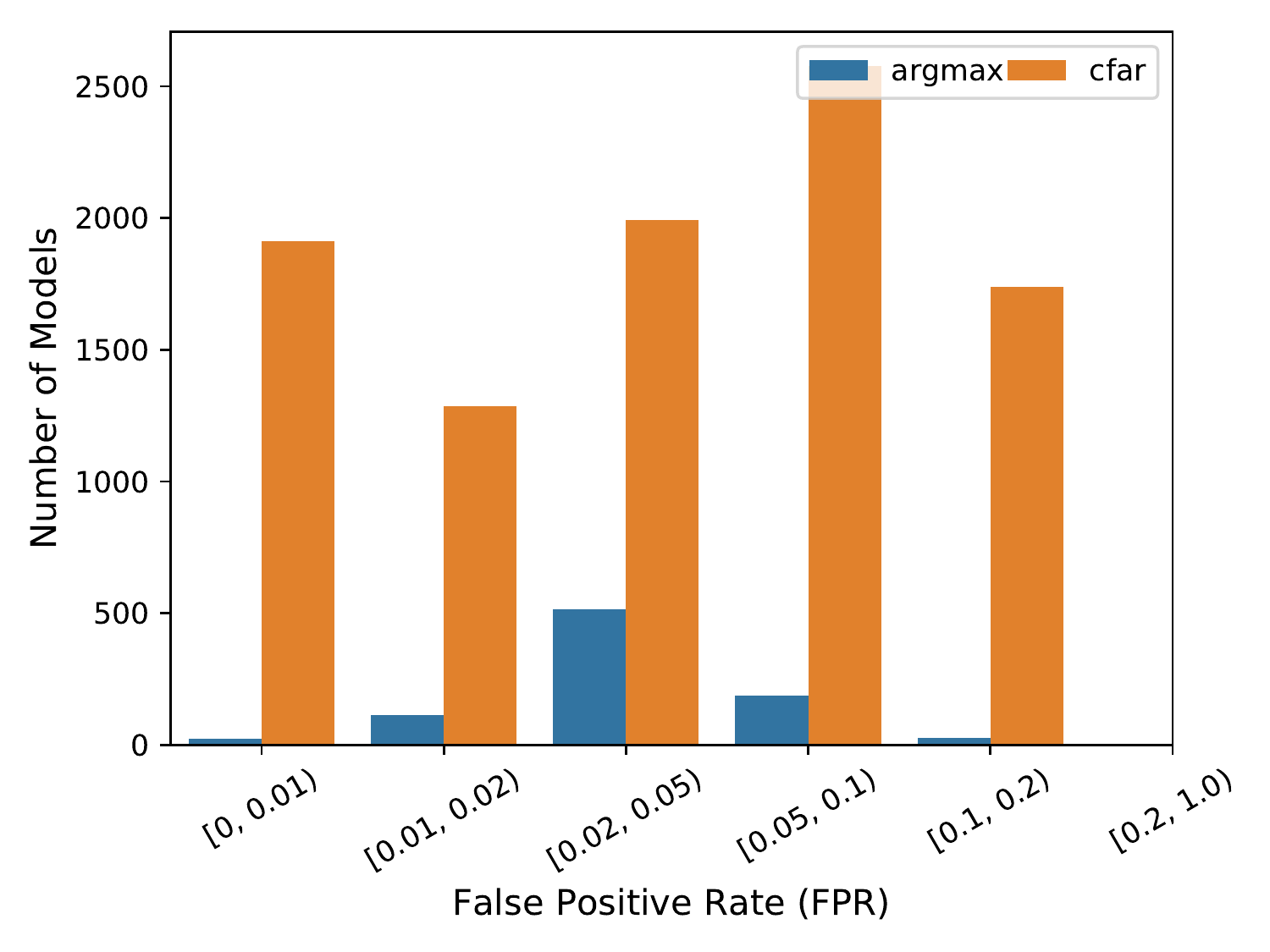}
    \caption{\ac{RF}: wine}
  \end{subfigure}
  
  \begin{subfigure}[t]{0.49\linewidth}
    \centering
    \includegraphics[height=3.2cm]{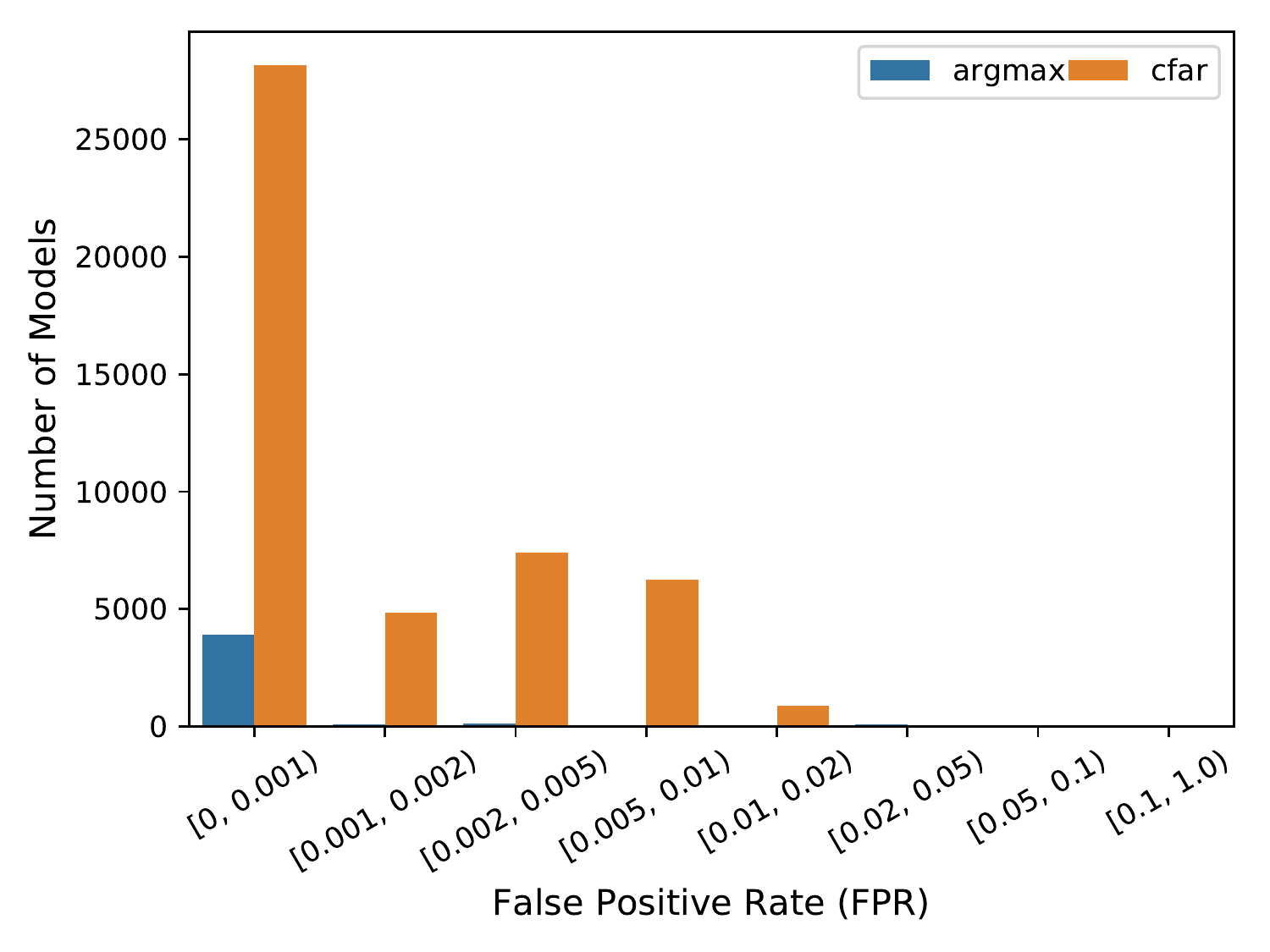}
    \caption{\ac{NN}: chiller}
  \end{subfigure}
  \begin{subfigure}[t]{0.49\linewidth}
    \centering
    \includegraphics[height=3.2cm]{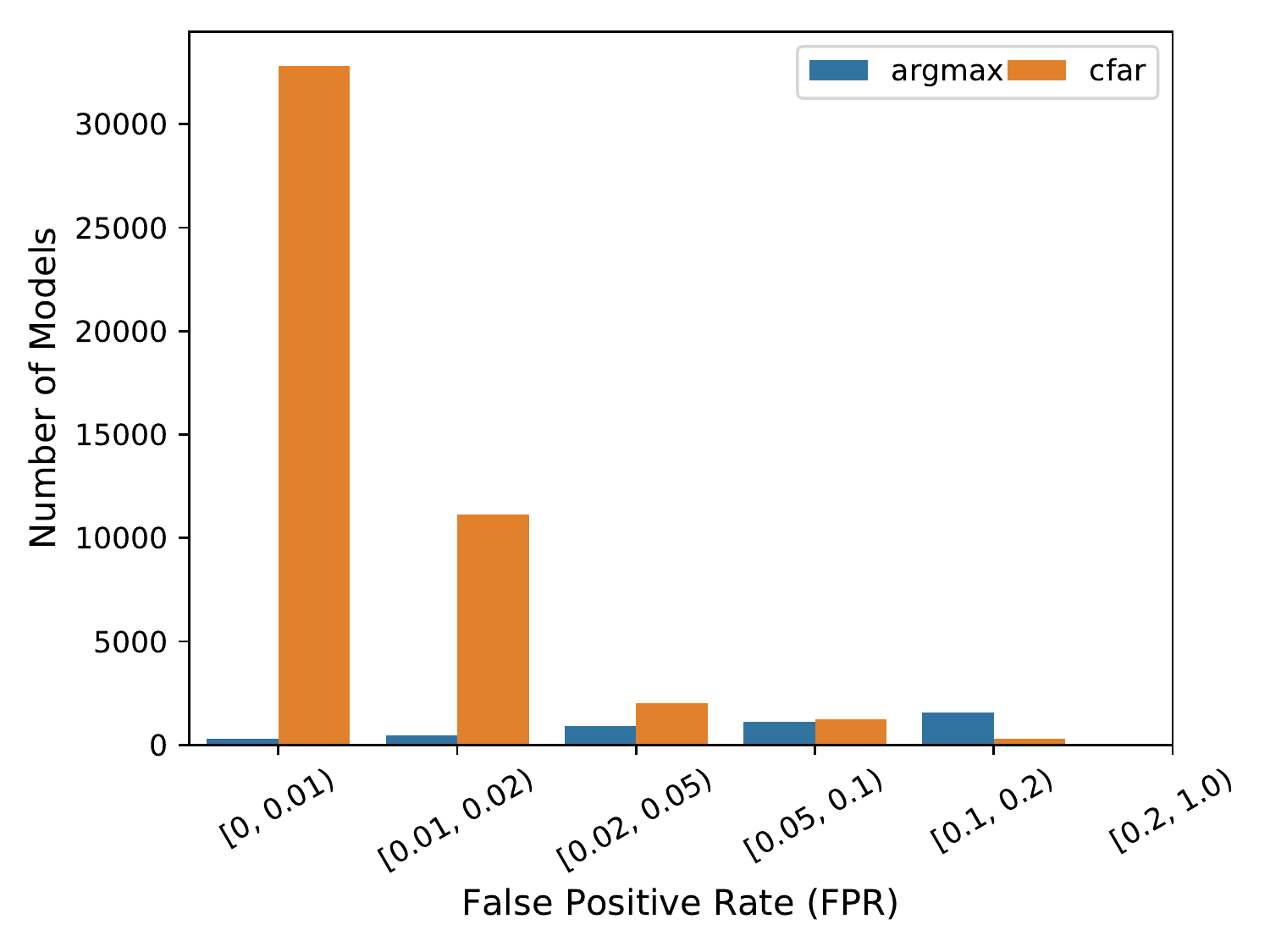}
    \caption{\ac{NN}: wine}
  \end{subfigure}
  
  \begin{subfigure}[t]{0.49\linewidth}
    \centering
    \includegraphics[height=3.2cm]{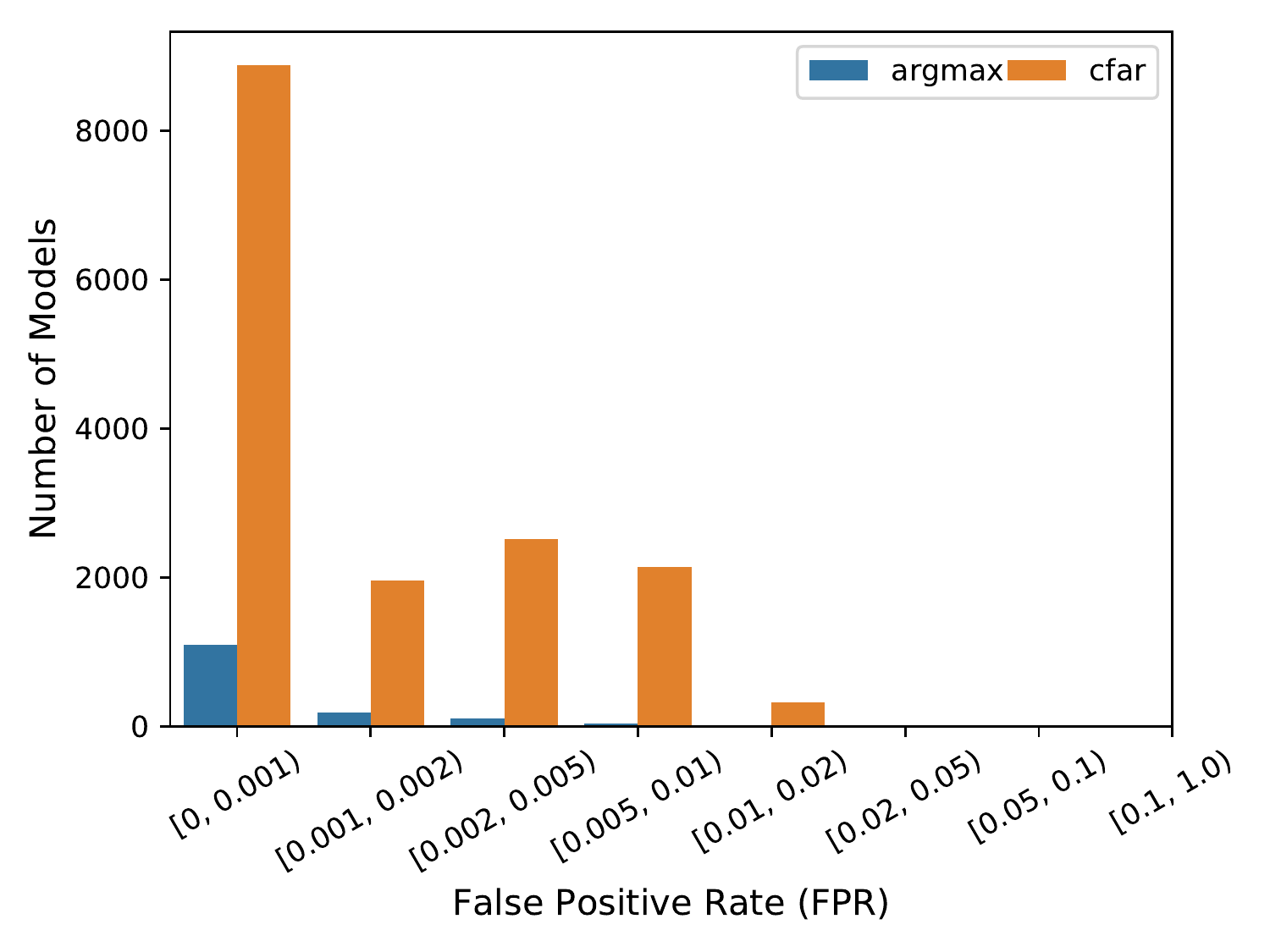}
    \caption{Dropout \ac{NN}: chiller}
  \end{subfigure}
  \begin{subfigure}[t]{0.49\linewidth}
    \centering
    \includegraphics[height=3.2cm]{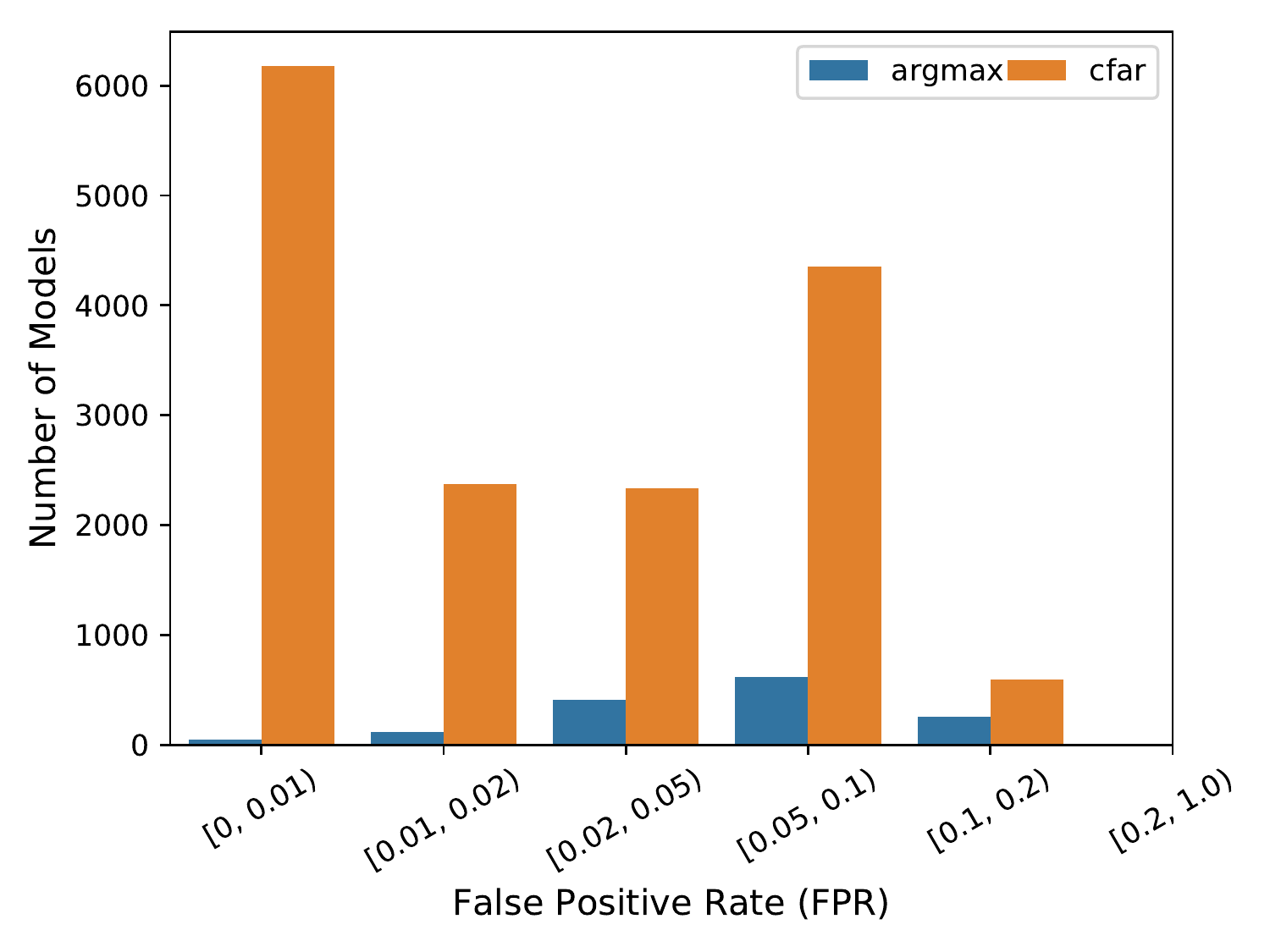}
    \caption{Dropout \ac{NN}: wine}
  \end{subfigure}
  
  
  
  \caption{Histograms showing the number of supervised classifiers falling under different \ac{FPR} intervals. The top $K$ classifiers among each \ac{FPR} bin are used to produce Figure~\ref{fig:FPR-sweep}.}
  \label{fig:FPR-hist}
\end{figure}

\subsection{\ac{OC-SVM}}
The \ac{OC-SVM} models in our experiments are implemented using the \texttt{OneClassSVM} class in \texttt{sklearn}. \ac{RBF} is used as the kernel type due to its flexibility in modeling complex distributions. $\nu$ and $\gamma$ are the two configurable hyperparameters with the \ac{RBF} kernel. The $\nu$ parameter takes a value between 0 and 1, which upper bounds the fraction of training errors and lower bounds the fraction of support vectors~\cite{jin2019one}. The $\gamma$ parameter controls the width of the \ac{RBF} kernel used for training. The larger the $\gamma$ parameter is, the smaller the width of the kernel. If $\gamma$ is too large, the model may overfit the data. As with other models, we conduct a grid search over different hyperparameter settings; see Table~\ref{tab:hyperparameters} for details.

\section{RP-1043 Chiller Dataset}\label{sec:chiller-dataset-appendix}
The RP-1043 chiller dataset is not publicly available but available for purchase from ASHRAE. The same sixteen features and six types of faults as used in previous work~\cite{jin2019encoder} (see details therein) are used to train our models in our empirical study. A few bad data points are removed first. For models other than \acp{RF}, the data are standardized before they are used for training. The layout of the generated training and test sets is illustrated in Figure~\ref{fig:dataset-layout}. During training, $20\%$ of the training data is held out as the validation set.

\begin{figure}[tb]
    \centering
    \begin{subfigure}[t]{0.48\linewidth}
    \centering
    \includegraphics[width=\linewidth]{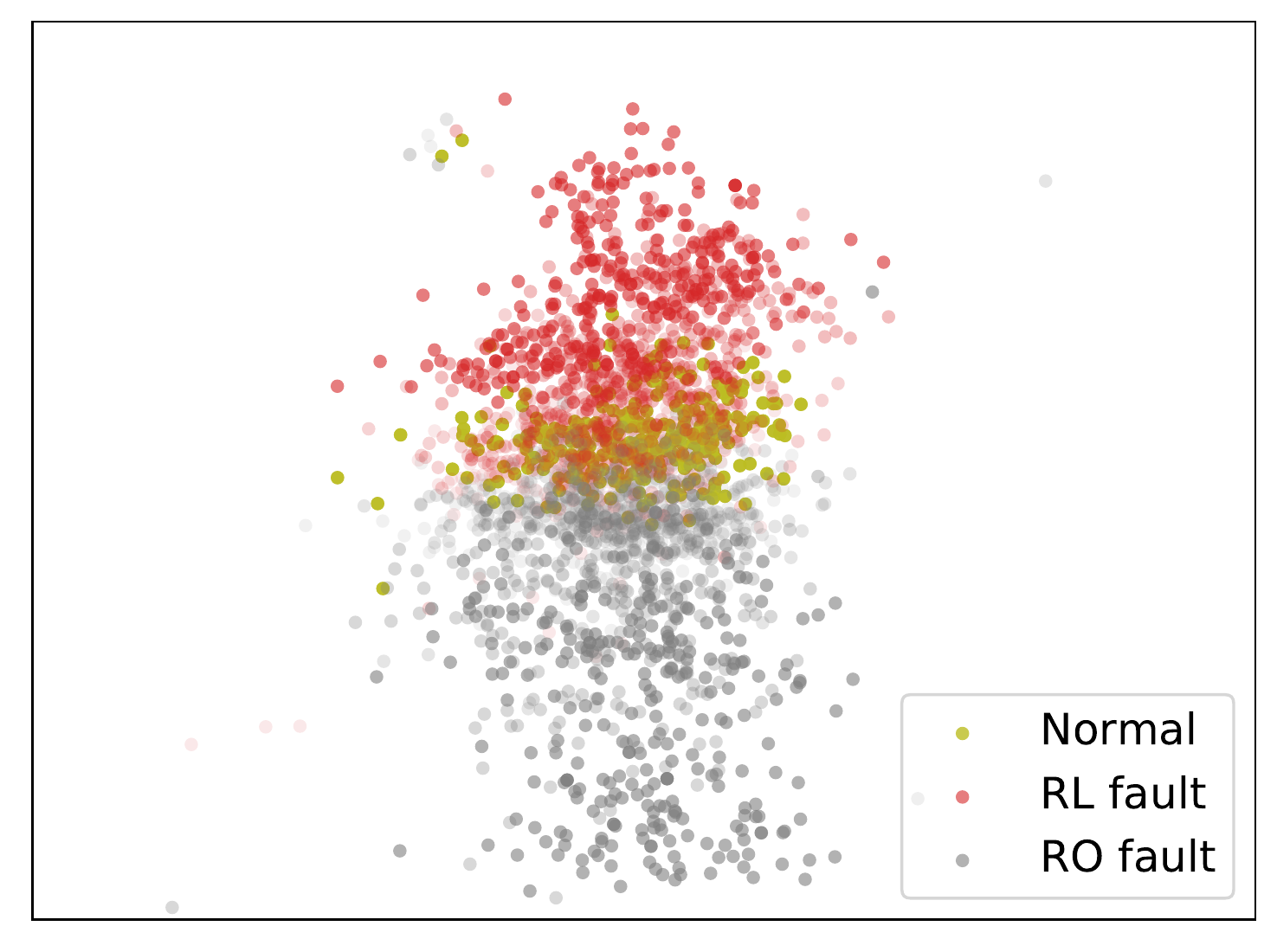}
    \caption{Normal \& RL faults \& RO faults}
    \end{subfigure}
    \begin{subfigure}[t]{0.48\linewidth}
    \centering
    \includegraphics[width=\linewidth]{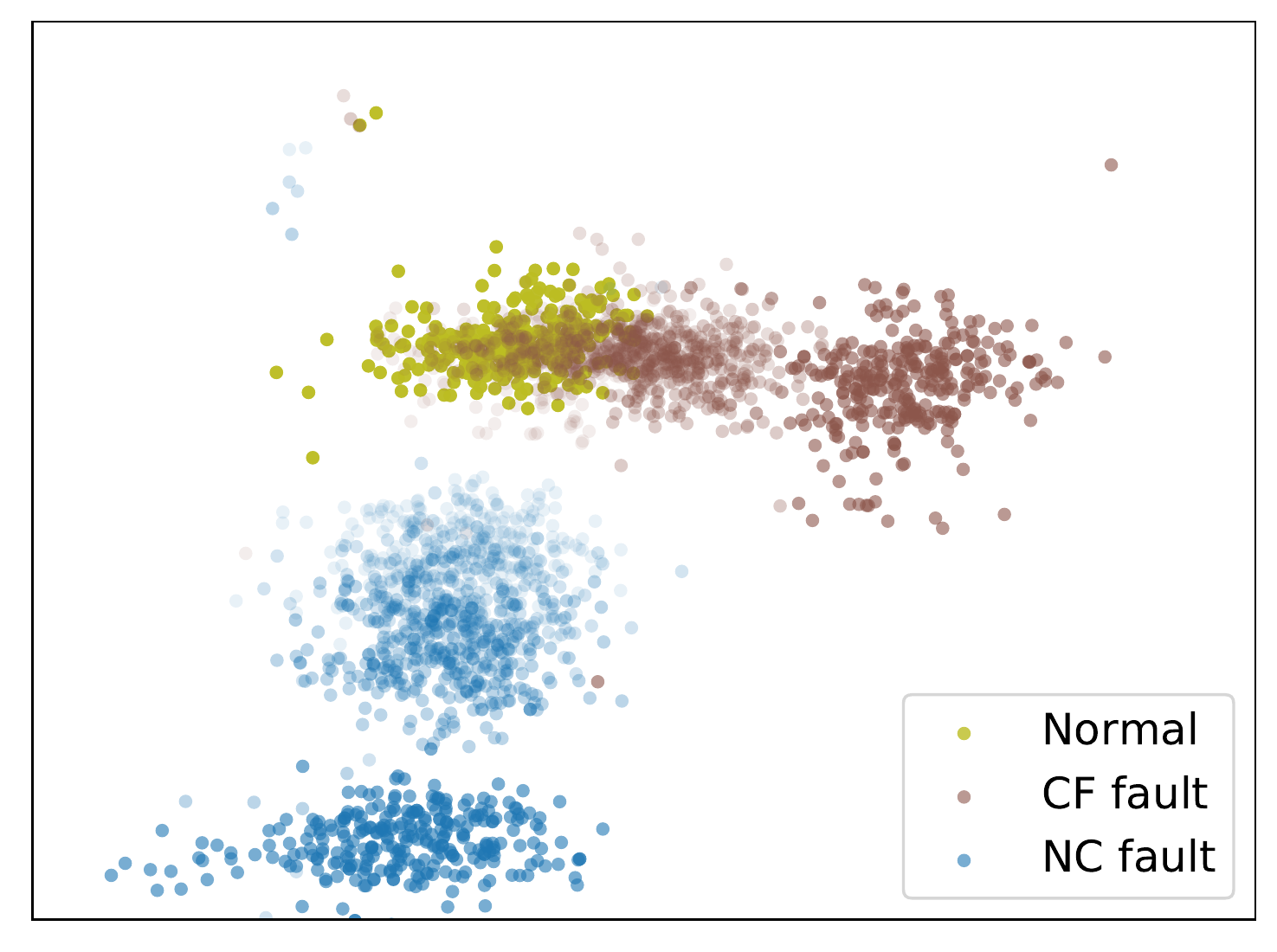}
    \caption{Normal \& CF faults \& NC faults}
    \end{subfigure}
    \caption{Additional visualization of the chiller dataset. Faults with lower \acp{SL} are plotted with lighter color intensities.}
    \label{fig:rp-1043-detailed}
\end{figure}

In Figure~\ref{fig:rp-1043-detailed}, we provide additional illustrations of the spatial relationship between the normal data and the fault data (RL, RO, CF and NC faults) that are not depicted in Figure~\ref{fig:visualization-chiller}. By comparing Figure~\ref{fig:rp-1043-detailed} and Figure~\ref{fig:visualization-chiller}, we can see that the RL, RO, CF and NC faults are much closer to the cluster of the normal data than FWE and FWC faults. Some \ac{IS} faults even overlap with the normal data in the reduced-dimension space, making them presumably much harder to detect.

\section{Wine Tasting Dataset}\label{sec:wine-dataset-appendix}
The wine tasting dataset~\cite{cortez2009modeling} is available for download from the UCI machine learning repository. All 11 original features in the wine data are used for training our models in our studies. For models other than \acp{RF}, the data are standardized before they are used for training. The layout of the generated training and test sets can also be found in Figure~\ref{fig:dataset-layout}. During training, $20\%$ of the training data is held out as the validation set.

\end{document}